\documentclass[twoside,11pt]{article}
\usepackage{jmlr2e}

\usepackage{morefloats}
\usepackage{caption}
\usepackage{graphicx}
\usepackage{epstopdf}
\usepackage{float}
\usepackage{latexsym}
\usepackage{multicol}
\usepackage{algorithmic}
\usepackage{pbox}
\usepackage{multirow}
\usepackage{hhline}
\usepackage{color}
\usepackage[table]{xcolor}
\definecolor{fluorescentpink}{rgb}{1.0, 0.08, 0.58}

\usepackage{amssymb}
\usepackage{amsmath}

\newcommand{\nop}[1]{}

\newcommand{\QED}{\mbox{\rule[0pt]{1.0ex}{1.0ex}}}
\def\boxend{\hspace*{\fill} $\QED$}

\makeatletter
\newcommand*{\indep}{%
  \mathbin{%
    \mathpalette{\@indep}{}%
  }%
}
\newcommand*{\nindep}{%
  \mathbin{
    \mathpalette{\@indep}{\not}
  }%
}
\newcommand*{\@indep}[2]{%
  \sbox0{$#1\perp\m@th$}
  \sbox2{$#1=$}
  \sbox4{$#1\vcenter{}$}
  \rlap{\copy0}
  \dimen@=\dimexpr\ht2-\ht4-.2pt\relax
  \kern\dimen@
  {#2}%
  \kern\dimen@
  \copy0 
}

\makeatother

\begin{document}

\title{A Unified View of Causal and Non-causal Feature Selection}
\author{\name Kui Yu \email ykui713@gmail.com \\
         \name Lin Liu \email Lin.Liu@unisa.edu.au \\
          \name Jiuyong Li \email Jiuyong.Li@unisa.edu.au \\
       \addr School of Information Technology and Mathematical Sciences\\
      University of South Australia\\
       Adelaide, 5095, SA, Australia
        }

\editor{}

\maketitle

\begin{abstract}

In this paper, we aim to develop a unified view of causal and non-causal feature selection methods. The unified view will fill in the gap in the research of the relation between the two types of methods. Based on the Bayesian network framework and information theory, we first show that causal and non-causal feature selection methods share the same objective. That is to find the Markov blanket of a class attribute, the theoretically optimal feature set for classification. 
We then examine the assumptions made by causal and non-causal feature selection methods when searching for the optimal feature set, and unify the assumptions by mapping them to the restrictions on the structure of the Bayesian network model of the studied problem. We further analyze in detail how the structural assumptions lead to the different levels of approximations employed by the methods in their search, which then result in the approximations in the feature sets found by the methods with respect to the optimal feature set.
With the unified view, we are able to interpret the output of non-causal  methods from a causal perspective and derive the error bounds of both types of methods. 
Finally, we present practical understanding of the relation between causal and non-causal methods using extensive experiments with synthetic data and various types of real-word data.
\end{abstract}


\section{Introduction}\label{sec1}

Feature selection is to identify a subset of features (predictor variables)
from the original features for model building or data understanding~\citep{guyon2003introduction,liu2005toward}. In the big data era, feature selection is more pressing than ever, since high-dimensional datasets have become ubiquitous in various applications~\citep{zhai2014emerging}. For example, in cancer genomics, a gene expression dataset can contain tens of thousands of features (genes). For another example, the Webb Spam Corpus 2011 has a collection of approximately 16 million features for web spam detection~\citep{wang2012evolutionary}. The high dimensionality not only incurs high computational cost and memory usage, but also deteriorates the generalization ability of prediction models~\citep{brown2012conditional}.
Therefore, many feature selection methods have been proposed, and they fall into three main categories, filter, wrapper, and embedded methods~\citep{li2016feature}. While filter feature selection methods are classifier or prediction model agnostic, the other two types of methods  are classifier dependent. With the rapid increase of high dimensional data, filter feature selection methods are attracting more attentions than ever, because of their fast processing speed, independence of prediction models, and robustness against overfitting (i.e. no bias on specific prediction models).
In this paper, we focus on filter methods,  and in the rest of this paper, feature selection refers to filter feature selection, unless otherwise mentioned. 

In the last two decades, feature selection has been well studied and has achieved great successes in building high quality classification models.
In classical feature selection, an input feature is considered as a strongly relevant feature, a weakly relevant feature, or an irrelevant feature with respect to a class attribute~\citep{kohavi1997wrappers},
and the methods aim to find the strongly relevant features of the class attribute. To achieve this goal, typically, a classical feature selection method will rank the features according to their relevance to the class attribute, and then iteratively selects for inclusion the top $\psi$ most relevant features~\citep{guyon2003introduction}.

An emerging feature selection approach is to identify a Markov blanket (MB) of the class attribute~\citep{koller1995toward,guyon2007causal,aliferis2010local1,aliferis2010local2}. The notion of MB was invented by Pearl~\citep{pearl1988probabilistic,pearl2014probabilistic} under the framework of causal Bayesian network (CBN). The MB of a variable in a CBN consists of its parents (direct causes), children (direct effects), and spouses (other parents of this variable's children) (For an exemplar MB, please see Figure~\ref{fig2-1} in Section~\ref{sec3}).
By tying feature predictive power and causality together, the MB discovery approach to feature selection can achieve more parsimonious feature subset than classical feature selection methods, thus lead to more interpretable and robust prediction models~\citep{guyon2007causal}.
Since the MB discovery approach explicitly induces local causal relations between a class attribute and the features while  classical feature selection methods do not, in this paper, we call the MB discovery approach causal feature selection while the classical (filter) feature selection approach non-causal feature selection~\citep{guyon2007causal,aliferis2010local1}.

A series of causal feature selection algorithms, such as IAMB~\citep{tsamardinos2003towards}, MMMB~\citep{tsamardinos2003time}, HITON-MB~\citep{aliferis2003hiton}, PCMB~\citep{pena2007towards}, and STMB~\citep{gao2017efficient} have been developed. 
Tsamardinos et al.~\citep{tsamardinos2003towards} were the first to build the connection between local causal discovery and feature selection, which opened the way to study the relation of causal and non-causal feature selection methods. 
Guyon et al.~\citep{guyon2007causal} conducted a comparison of the motivations and pros/cons of causal and non-causal feature selection approaches. However, the analysis was at conceptual and general discussion level.
Brown et al.~\citep{brown2012conditional} unified information theoretic feature selection methods. 
These pioneer work provides a basis of studying causal and non-causal feature selection methods. However, to the relations between the two major approaches to feature selection,  the following fundamental questions are yet to be investigated:

\begin{itemize}
\item Firstly, what is the relation between the objectives of causal feature selection and non-causal feature selection, i.e. what is the relation between the set of all features strongly relevant to the class attribute and finding the MB of the class attribute?



\item Secondly, driven by their respective objectives, how are the search strategies employed by the two types of feature selection methods different? 

\item Thirdly, what are the underlying assumptions leading to the different search strategies?

\end{itemize}

To answer these questions, in this paper, we develop a unified view of causal and non-causal feature selection by systematically studying the relation between the two approaches from the perspectives of their objective functions,  assumptions, search strategies, and the error bounds by employing the Bayesian network framework and information theory. Specifically,  we have made the following contributions in this paper:
\begin{itemize}
\item We derive a mutual information based representation of the optimal feature set for classification. Based on the representation, we develop a unified representation of the objective function of causal and non-causal feature selection by showing that both types of methods share the same objective.

\item We analyze the assumptions made by the major causal and non-causal feature selection methods in their search for the feature set specified by the objective function. Our findings show that these assumptions can be unified under the Bayesian network framework, and the assumptions can be represented as different levels of restrictions on the structure of the Bayesian network model of the problem under consideration.


\item We analyze the search strategies of the causal and non-causal feature selection methods, and discover that as a result of the different levels of assumptions, different search strategies have been taken by the methods, which then result in different levels of approximations of the optimal feature set.

\item We analyze the output of non-causal feature selection methods from a causal perspective and derive the error bounds of the two major approaches to feature selection.
\item We conduct extensive experiments using synthetic and real-world datasets to validate the relationship between the assumptions and approximations made by causal and non-causal feature selection methods, the causal interpretations of non-causal feature selection, and the derived error bounds of both types of feature selection methods.
\end{itemize}

In summary, we propose a unified view to bridge the gap in current understanding of the relation between causal and non-causal feature selection methods. With the unified view, we are able to understand the mechanisms of two major feature selection approaches, and thus to connect causality  to predictive feature selection and interpret  the output of non-causal methods using a causal framework. Moreover, by filling in the gap, we hope to leverage the cross-pollination between causal and non-causal feature selection to develop new  methodologies promising to deliver more robust data analysis than each field could individually do.

The rest of the paper is organized as follows. Section~\ref{sec2} discusses the related work, and Section~\ref{sec3} presents the key notations and definitions. Section~\ref{sec4} analyzes the objective functions and the rationale of causal and non-causal feature selection methods. Section~\ref{sec5} identifies and examines the assumptions made by causal and non-causal feature selection methods and their corresponding search strategies. Section~\ref{sec6} discusses the error bounds of causal and non-causal feature selection methods. Section~\ref{sec8} presents the experiments and demonstrates how the developed unified view provides practical understanding the relations between causal and non-causal feature selection methods, and Section~\ref{sec9} concludes the paper.

\section{Related work}\label{sec2}

In this section, we will review causal and non-causal (filter) feature selection methods. Excellent reviews of  non-causal feature selection (i.e. filter, embedded, wrapper) algorithms can be found in~\citep{guyon2003introduction,liu2007computational,brown2012conditional,li2016feature} and the reference therein.

\subsection{Non-causal feature selection}

A general filter feature selection method consists of two elements: a search strategy for feature subset generation and an evaluation criterion for measuring relevance of the features. This evaluation criterion is to estimate how useful a feature or a feature subset may be when used in a learning algorithm (e.g. a classifier). As the feature selection by a filter method is carried out separately from the process of learning a model, an effective evaluation criterion plays a key role in filter methods. In the past decades, different evaluation criteria have been proposed, such as those based on distance~\citep{kira1992practical,robnik2003theoretical}, mutual information~\citep{bontempi2010causal,nguyen2014effective,shishkin2016efficient}, dependency~\citep{song2012feature}, and consistency~\citep{dash2003consistency}. 
Since mutual information is a general measure of feature relevance with several unique properties~\citep{cover2012elements}, there has been a significant amount of work on mutual information-based feature selection methods developed in the past two decades (see~\citep{brown2012conditional,vergara2014review} for an exhaustive list). 

In this paper, we use mutual information as a basic tool to develop the unified view, so in this section, we focus on non-causal feature selection methods which are based on mutual information. 
Many advances in the field have been reported since the pioneer work of Lewis~\citep{lewis1992feature} and Battiti~\citep{battiti1994using}. Lewis proposed the MIM (Mutual Information Maximisation) criterion. MIM simply ranks the features in order of their MIM scores (i.e. the value of mutual information between a feature and the class attribute) and selects the top $\psi$ most relevant features from the original feature set. However, MIM only considers feature relevance. Then Battiti proposed the MIFS (Mutual information Feature Selection) criterion which not only considers feature relevance, but also adds a penalty for feature redundancy. MIFS uses a greedy search to select features sequentially (i.e. a single feature at a time), and iteratively constructs the final feature subset, as an alternative to the evaluation of the combinatorial explosion of all subsets of features. 

Based on the MIFS criterion, many variants have been proposed. The representative algorithms include mRMR~\citep{peng2005feature}, CIFE~\citep{lin2006conditional},  FCBF~\citep{yu2004efficient}, mIMR~\citep{bontempi2010causal}, and MRI~\citep{wang2017feature}.
Yang and Moody proposed the JMI (Joint Mutual Information) criterion~\citep{yang2000data}.  Compared to the MIFS criterion, the JMI criterion considers complementary information between features by evaluating class-conditional relevance, to see if a feature would provide more predictive information when it is used jointly with other features in the prediction compared with the case when the feature is used alone.
The IF~\citep{vidal2003object}, DISR~\citep{meyer2008information}, CMIM~\citep{fleuret2004fast}, and RelaxMRMR~\citep{vinh2016can} methods can be considered as the variants of the JMI criterion. Brown et al.~\citep{brown2012conditional}  unified almost two decades of research on commonly used heuristics of mutual information based feature selection methods into the framework of conditional likelihood maximisation.

Owing to the difficulty of estimating mutual information with high dimensional data, most existing mutual information-based methods use various low-order approximations for estimating mutual information. While those approximations have been successful in certain applications, they are heuristic in nature and lack theoretical guarantees.  Thus, the main problems with the majority of mutual information-based methods are that in most cases it is unknown what consists an optimal feature selection solution independent of the type of models fitted, and under which conditions a filter method will output an optimal feature set for classification~\citep{guyon2007causal,aliferis2010local1}.

\subsection{Causal feature selection}

 As an emerging type of filter methods, causal feature selection has attracted much attention in recent years. By bringing causality into play, causal feature selection naturally provides causal interpretation about the relationships between features and the class attribute, enabling a better understanding of the mechanisms behind data. Compared to non-causal feature selection, causal feature selection has been shown to be theoretically optimal~\citep{tsamardinos2003towards}, and thus answers the questions of what consists an optimal feature selection solution and under which conditions a filter method will output an optimal feature for classification.
 
Causal feature selection is to find the MB of the class attribute in a causal Bayesian network (CBN), where an edge $X\rightarrow Y$ indicates that $X$ is a direct cause (parent) of $Y$, and Y is a direct effect (child) of X. Then the MB of a variable of interest, such as the class attribute, consists of direct causes, direct effects, and direct causes of the direct effects of the class attribute.
Therefore, the MB of the class attribute provides a complete picture of the local causal structure around it and the MB is a minimal set of features which renders the class attribute statistically independent from all the remaining features conditioned on the MB~\citep{pearl2014probabilistic}. Theoretically, the MB of the class attribute is the optimal feature subset for classification~\citep{koller1995toward,tsamardinos2003towards}.
Accordingly, the discovery of the MB of a class attribute is actually a procedure of feature selection~\citep{aliferis2010local1}.

Koller and Sahami~\citep{koller1995toward} were the first to introduce MBs to feature selection and proposed the Koller-Sahami (KS) algorithm. However, the KS algorithm is not guaranteed to find the actual MB.
Margaritis and Thrun~\citep{margaritis2000bayesian} invented the first sound MB discovery algorithm, GS (Growing-Shrinking) for Bayesian network structure learning. 



Tsamardinos and Aliferis~\citep{tsamardinos2003towards} improved the GS algorithm and proposed a series of MB discovery algorithms for optimal feature selection, which led to the IAMB (Incremental Association-based MB) family of algorithms, such as IAMB, inter-IAMB, IAMBnPC~\citep{tsamardinos2003algorithms}, and Fast-IAMB~\citep{yaramakala2005speculative}.

However, given a variable of interest, the IAMB and its variants discover the parents and children  (PC) and spouses simultaneously and do not distinguish PC from spouse during MB discovery. And these algorithms require a  large number of data samples exponential to the size of the MB of the variable, thus they would not be effective for MB discovery when a dataset has thousands of variables with a small-sized data samples. 

Then a divide-and-conquer approach was proposed to mitigate the problem.
The representative algorithms include HITION-MB~\citep{aliferis2003hiton,aliferis2010local1},  MMMB~\citep{tsamardinos2003time}, PCMB~\citep{pena2007towards}, IPC-MB~\citep{fu2008fast}, and STMB~\citep{gao2017efficient}. The ideas behind these algorithms are as follows. They firstly find the parents and children (PC) of a variable of interest. Then, they discover the variable's spouses. Thus, these methods can return both the PC and MB sets of the variable. How to efficiently and effectively find the PC set of a variable is the key to this type of approach. The PC-simple~\citep{buhlmann2010variable}, MMPC~\citep{tsamardinos2006max}, HITON-PC~\citep{aliferis2003hiton}, and semi-HITON-PC~\citep{aliferis2010local1} algorithms are for PC discovery.

\section{Bayesian network, Markov blanket, and feature selection}\label{sec3}

Let $C$ be the class attribute of interest, and $C$ has $\varphi$ distinct values (class labels), denoted as $c=\{c_1,c_2,\cdots,c_\varphi\}$ and $F=\{F_1,F_2,\cdots,F_n\}$ be the set of all $n$ distinct features. Assuming that a training dataset $D$ is defined by $D=\{(d_i,c_i), 1\leq i \leq m,\ c_i\in c\}$, where $m$ is the number of data instances, $d_i$ is the $i$th data instance which is a $n$-dimensional vector defined on $F$, and $c_i$ is a class label associated with $d_i$. 
For the convenience of presentation, we use $V$ to represent the set of all variables under consideration, i.e. $V=F\cup\{C\}=\{V_1, V_2, \cdots, V_{n+1}\}$, where $V_i=F_i\ (1\le i \le n)$, and $V_{n+1}=C$.
For $\forall V_i\in V$, let $V\setminus V_i$ indicate the set $V\setminus\{V_i\}$, that is, all features excluding $V_i$.
We use $V_i\indep V_j|S$, where $i\neq j$ and $S\subseteq V\setminus\{V_i, V_j\}$,  to denote that $V_i$ is conditionally independent of $V_j$ given  $S$, and $V_i\nindep V_j|S$  to represent that $V_i$ is conditionally dependent on $V_j$ given $S$. The definition of conditional independence (and dependence) is given as follows.
\begin{definition}[Conditional independence] Given two distinct variables $V_i, V_j\in V$  are said to be conditionally independent given a subset of variables $S\subseteq V\setminus\{V_i, V_j\}$ (i.e. $V_i\indep V_j|S$), if and only if $P(V_i,V_j|S)=P(V_i|S)P(V_j|S)$. Otherwise, $V_i$ and $V_j$ are conditionally dependent given $S$, i.e. $V_i\nindep V_j|S$.
\label{def2-1}
\end{definition}


\subsection{Bayesian network and Markov blanket}\label{sec3-1}

In this section, we introduce the background knowledge related to causal feature selection, including the basics of Bayesian network, Markov blanket, and the aim of causal feature selection.
Let $P(V)$ be the joint probability distribution over the set of all variables $V$, and $G=(V, E)$ represent a directed acyclic graph (DAG) with nodes $V$ and edges $E$, where an edge represents the direct dependence relationship between two variables. In a DAG, $V_i\rightarrow V_j$ denotes that $V_i$ is a parent of $V_j$ and $V_j$ is a child of $V_i$.
\begin{definition} [Bayesian network]~\citep{pearl2014probabilistic} 
The triplet $\langle V, G, P(V)\rangle$ is called a Bayesian network if the Markov condition as defined in Definition~\ref{def2-2} holds.
\label{def2-01}
\end{definition}
\begin{definition}
[Markov condition]~\citep{pearl2014probabilistic} For a DAG $G$, the Markov condition holds in $G$ if and only if every node of $G$ is independent of any subset of its non-descendants conditioned on its parents.
\label{def2-2}
\end{definition}

A Bayesian network encodes the joint probability over a set of variables $V$ and decomposes $P(V)$ into the product of the conditional probability distributions of the variables given their parents in $G$. Let $Pa(V_i)$ be the set of parents of $V_i$ in $G$.  Then, $P(V)$ can be written as
\begin{equation}\label{eq1}
P(V_1, V_2,\cdots,V_{n+1}) = \prod^{n+1}_{i=1}{P(V_i|Pa(V_i))}
\end{equation}

In this paper, we consider a causal Bayesian network, a Bayesian network in which an
edge $V_i\rightarrow V_j$ indicates that $V_i$ is a direct cause of $V_j$~\citep{pearl2014probabilistic,spirtes2000causation}. For simple presentation, however, we use the term Bayesian network instead of causal Bayesian network. In the following, we introduce the key concepts and assumptions related to Bayesian networks and Markov blankets.

\begin{definition}
[Faithfulness]~\citep{pearl2014probabilistic} Given a Bayesian network $<V, G, P(V)>$,  $G$ is faithful to $P(V)$ if and only if every conditional independence present in $P$ is entailed by $G$ and the Markov condition. $P(V)$ is faithful if and only if $G$ is faithful to $P(V)$.
\label{def2-4}
\end{definition}

\begin{definition}
[Causal sufficiency]~\citep{pearl2014probabilistic} Causal sufficiency assumes that any common cause of two or more variables in $V$ is also in $V$.
\label{def2-yk1}
\end{definition}

\begin{definition}[d-separation]~\citep{pearl2014probabilistic} In a DAG $G$, a path $\pi$ is said to be d-separated by a set of nodes $S\subset V$ if and only if (1)
$\pi$ contains a chain $V_i\rightarrow V_\omega\rightarrow V_j$ ($V_i\leftarrow V_\omega\leftarrow V_j$) or a fork $V_i\leftarrow V_\omega\rightarrow V_j$ such that the middle node $V_\omega$ is in $S$, or
(2) $\pi$ contains a v-structure $V_i\rightarrow V_\omega\leftarrow V_j$ such that $V_\omega\notin S$ holds and no descendants of $V_\omega$ are in $S$.
A set $S$ is said to d-separate $V_i$ from $V_j$ if and only if $S$ blocks every path from  $V_i$ to $V_j$.
\label{def2-5}
\end{definition}

\begin{theorem}~\citep{pearl2014probabilistic,spirtes2000causation} Given a Bayesian network $<V, G, P(V)>$, under the faithfulness assumption,  d-separation captures all conditional independence relations that are encoded in $G$, which implies that $V_i$ and $V_j$ in $G$ are d-separated by $S\subset V\backslash\{V_i, V_j\}$, if and only if $V_i$ and $V_j$ are conditionally independent given $S$ in $P(V)$.
\label{the2-1}
\end{theorem}

Theorem~\ref{the2-1} concludes that under the assumption of faithfulness, conditional independence in a data distribution and  d-separation in the corresponding DAG are equivalent.

\begin{definition}[Markov blanket, MB]~\citep{pearl2014probabilistic}
Under the faithfulness assumption, the MB of a variable in a Bayesian network is unique and consists of its parents (direct causes), children (direct effects), and spouses (other parents of the variable's children). 
\label{def2-6}
\end{definition}

\begin{figure}
\centering
\includegraphics[height=1.6in,width=3.2in]{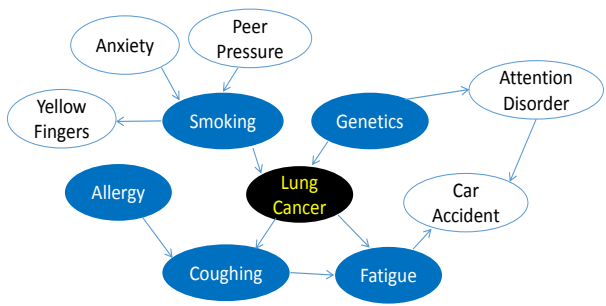}
\caption{An example of an MB in a lung-cancer Bayesian network}
\label{fig2-1}
\end{figure}

Figure~\ref{fig2-1} gives an example of a MB in the Bayesian network of lung cancer~\citep{guyon2007causal}. The MB of the variable \emph{lung cancer} comprises:  \emph{Smoking} and \emph{Gentics} (parents), \emph{Coughing} and \emph{Fatigue} (children), and \emph{Allergy} (spouse). 

Given a dataset $D$ defined on $F\cup C$, causal feature selection aims to find the MB of the class attribute $C$ (denoted as $MB(C)$) from $D$~\citep{aliferis2010local1}.
In the following, Proposition~\ref{pro2-1} illustrates the relations between parents and children in a Bayesian network, and Proposition~\ref{pro2-2} presents the idea of how to discover spouses.


\begin{proposition}\citep{spirtes2000causation}\label{pro2-1}
In a Bayesian network, there is an edge between the pair of nodes $V_i$ and $V_j$, if and only if  $V_i\nindep V_j|S$, for all $S\subseteq V\setminus \{V_i, V_j\}$.
\label{pro2-1}
\end{proposition}

\begin{proposition}\citep{spirtes2000causation}\label{pro2-2}
In a Bayesian network, assuming that $V_i$ is adjacent to $V_j$, $V_j$ is adjacent to $V_\omega$, and $V_i$ is not adjacent to $V_\omega$ (e.g. $V_i\rightarrow V_j\leftarrow V_\omega$), if
$\forall S\subseteq V\setminus\{V_i, V_j, V_\omega\}$, $V_i\indep V_\omega|S$ and $V_i\nindep V_\omega|S\cup \{V_j\}$ hold, then $V_i$ is a spouse of $V_\omega$.
\label{pro2-2}
\end{proposition}


\subsection{ Feature relevancy and non-causal feature selection}

Non-causal feature selection categorizes a feature as strongly relevant, weakly relevant, or irrelevant to $C$~\citep{kohavi1997wrappers} based on the following definitions in terms of conditional independence.
\begin{definition} [Strongly relevant feature]~\citep{kohavi1997wrappers} $F_i\in F$ is strongly relevant to $C$, if and only if there exists an assignment $F= f=(f_1, \cdots, f_{i-1}, f_i, f_{i+1},\cdots, f_n)$ 
and  $C=c_i,\ c_i\in c$, such that $P(F=f)>0$ and 
$P(C=c_i|F=f)\neq P(C=c_i|F\setminus F_i=(f_1, \cdots, f_{i-1}, f_{i+1}, \cdots, f_n))$. 
\end{definition}

\begin{definition} [Weakly relevant feature]~\citep{kohavi1997wrappers} $F_i\in F$ is weakly relevant to $C$, if and only if $F_i$ is not a strongly relevant feature and there exist $S\subset F\setminus F_i$, and
an assignment $F_i=f_i$, $C=c_i$ and $S=s$ such that $P(S=s, F_i=f_i)>0$ and $P(C=c_i|S=s,F_i=f_{i})\neq P(C=c_i|S=s)$.
\end{definition}
\begin{definition} [Irrelevant feature]~\citep{kohavi1997wrappers} $F_i\in F$ is irrelevant to $C$, if and only if for any $S\subseteq F\setminus F_i$, for any assignment of $F_i$, $S$ and $C$, denoted as $f_i$, $s$, and $c_i$, such that $P(C=c_i|S=s,F_i=f_{i})=P(C=c_i|S=s)$.
\end{definition}

A strongly relevant feature affects the conditional class distribution, and provides unique information about $C$, i.e. it cannot be replaced by other features. A weakly relevant feature is informative but redundant  since it can be replaced by other features without losing information about $C$. An irrelevant feature does not bring any information about $C$ and should be discarded. 
Thus, given a dataset $D$ defined on $F\cup C$, non-causal (filter) feature selection aims to select all features that are strongly relevant to $C$~\citep{tsamardinos2003towards}.

In addition to the above conditional probability based definitions, recently, an explanation of feature relevance based on mutual information  was proposed~\citep{brown2012conditional,bell2000formalism,vergara2014review}. Before discussing the explanation, we first introduce the concepts of mutual information below. Given variable $X$, the entropy of $X$ is defined as~\citep{cover2012elements}.
\begin{equation}\label{eq_2}
H(X)=-\Sigma_{x}P(x)\log{P(x)}
\end{equation}

The entropy of $X$ after observing values of another variable $Y$ is defined as
\begin{equation}\label{eq_3}
H(X|Y)=-\Sigma_{y}P(y)\Sigma_{x}P(x|y)\log{P(x|y)}.
\end{equation}

In Eq.(\ref{eq_2}) and Eq.(\ref{eq_3}), $P(x)$ is the prior probability of $X=x$ (i.e. the value $x$ that $X$ takes), and $P(x|y)$ is the posterior probability of $X=x$ given  $Y=y$. According to Eq.(\ref{eq_2}) and Eq.(\ref{eq_3}), the mutual information between $X$ and $Y$, denoted as $I(X,Y)$, is defined as
\begin{equation}\label{eq_4}
\begin{array}{rcl}
I(X; Y)&=&H(X)-H(X|Y)\\
&=&\Sigma_{x, y}P(x,y)\log\frac{P(x,y)}{P(x)P(y)}.
\end{array}
\end{equation}

From Eq.~(\ref{eq_4}), the conditional mutual information between $X$ and $Y$ give another feature $Z$ is defined as:
\begin{equation}\label{eq_5}
\begin{array}{rcl}
 I(X;Y|Z)& =& H(X|Z)-H(X|YZ) \\
 &=&\Sigma_{z\in Z}P(z)\Sigma_{x\in X, y\in Y}P(x,y|z)\log\frac{P(x,y|z)}{P(x|z)P(y|z)}
\end{array}
\end{equation}

Based on the above definitions mutual information, we have the following propositions.

\begin{proposition}~\citep{brown2012conditional}
$F_i$ is strongly relevant to $C$ if and only if $I(F_i;C|F\setminus F_i)>0$.
\label{def2-9}
\end{proposition}

\begin{proposition}~\citep{brown2012conditional}
$F_i$ is weakly relevant to $C$ if and only if $I(F_i;C|F\setminus F_i)=0$ and  $\exists S\subset F\setminus F_i$ such that $I(F_i;C|S)>0$.
\label{def2-10}
\end{proposition}

\begin{proposition}~\citep{brown2012conditional}
$F_i$ is irrelevant to $C$,  if and only if $\forall S\subseteq F\setminus F_i$, $I(F_i;C|S) = 0$.
\label{def2-11}
\end{proposition}




\section{Causal and non-causal feature selection have the same objective}\label{sec4}

To develop a unified view of causal feature selection and non-causal feature selection, in this section, we will show that the two types of feature selection, although originating from different fields, share the same  objective.
In order to derive this conclusion (in Section~\ref{sec4-2}), firstly in Section~\ref{sec4-1}, inspired by the work in (Brown et al. 2012),  
we  propose a mutual information based description of the optimal feature set for classification (i.e.~Eq.(\ref{eq3_2})), and then link the description to Bayes error rate of classification.

\subsection{A mutual information based representation of the objective function of optimal feature selection}\label{sec4-1}

Given a dataset $D$ containing $C$ and $F$, (filter) feature selection can be  formulated as the problem of finding a subset $S^*\subset F$ such that
\begin{equation}\label{eq1_0}
S^*  =  \mathop{\arg\max}_{S\subset F}P(C|S)
\end{equation}
i.e. finding a subset $S^*$ given which the conditional probability of $C$ is maximized~\citep{guyon2006introduction,brown2012conditional}.

Let $F=\{S\cup \overline{S}\}$ where $S$ denotes the selected feature set and $\overline{S}$ represents  the remaining features, i.e. $F\setminus S$.  Given a dataset $D$ of $m$ instances,  let $p(C|S)$ denote the true class distribution and $q(C|S)$ represent the predicted class distribution given $S$, then the conditional likelihood of $C$ is $L(C|S,D)=\prod_{i=1}^{m}{q(c_i|s_i)}$, where $c_i\in c\ (c=\{c_1,c_2,\cdots,c_\varphi\})$ represents the value of $C$ in the $i$-th data instance and $s_i$ denotes the value of feature set $S$ in the $i$-th data instance. The (scaled) conditional log-likelihood of $L(C|S,D)$ is calculated by
\begin{equation}\label{eq_10}
\ell(C|S,D)=\frac{1}{m}\sum_{i=1}^{m}{\log{q(c_i|s_i)}}.
\end{equation}

Eq.(\ref{eq_10}) can be re-written as Eq.(\ref{eq_11}) below~\citep{brown2012conditional}\footnote{Please refer to Section 3.1 of~\cite{brown2012conditional} for the details on how to obtain Eq.(\ref{eq_10}) and Eq.(\ref{eq_11}).}.
\begin{equation}\label{eq_11}
\ell(C|S,D)=\frac{1}{m}\sum_{i=1}^{m}{\log{\frac{q(c_i|s_i)}{p(c_i|s_i)}}}+\frac{1}{m}\sum_{i=1}^{m}{\log{\frac{p(c_i|s_i)}{p(c_i|f)}}}+\frac{1}{m}\sum_{i=1}^{m}{\log{p(c_i|f)}}
\end{equation}

By negating Eq.(\ref{eq_11}) and using $E$ to represent statistical expectation, we have:
\begin{equation}\label{eq_12}
-\ell(C|S,D)=E\bigg\{{\log{\frac{p(c|s)}{q(c|s)}}}\bigg\}+E\bigg\{{\log{\frac{p(c|f)}{p(c|s)}}}\bigg\}-E\bigg\{{\log{p(c|f)}}\bigg\}
\end{equation}

On the right hand side of Eq.(\ref{eq_12}), the first term is the likelihood ratio between the true and predicted class distributions given $S$, averaged over the input data space. The second term equals to $I(C;\overline{S}|S)$, that is, the conditional mutual information between $C$ and $\overline{S}$ given $S$~\citep{brown2012conditional}. The final term is $H(C|F)$ by Eq.(\ref{eq_3}), the conditional entropy of $C$ given all features, and is an irreducible constant. 
\begin{definition}[Kullback Leibler divergence]\citep{kullback1951information} The Kullback Leibler divergence between two probability distributions $P(X)$ and $Q(X)$ is defined as
 $KL(P(X)||Q(X))=\Sigma_{x}P(x)\log{\frac{P(x)}{Q(x)}}=E_{x}\log{\{\frac{P(X)}{Q(X)}\}}.$
 \label{def3-1}
\end{definition}

By Definition~\ref{def3-1} and Eq.(\ref{eq_12}), we have
\begin{equation}\label{eq_14}
\lim_{m\to\infty}{-\ell(C|S,D)}=KL(p(C|S)||q(C|S))+I(C;\overline{S}|S)+H(C|F).
\end{equation}

Since in Eq.(\ref{eq_14}), $KL(p(C|S)||q(C|S))$ will approach zero with a large $m$. Based on Eq.(\ref{eq_14}), we see that for large $m$ minimizing $I(C;\overline{S}|S)$ maximizes $L(C|S,D)$. 
By the chain rule of mutual information, Eq.(\ref{eq_16}) below holds.
\begin{equation}\label{eq_16}
\begin{array}{lll}
I(C;F)&=&I(C;\{S,\overline{S}\})\\
            &=&I(C;S)+I(C;\overline{S}|S)
\end{array}
\end{equation}

Given the feature set $F$ and the class attribute $C$, if $I(C;F)$ is fixed, then in Eq.(\ref{eq_16}), minimizing $I(C;\overline{S}|S)$ is equivalent to maximizing $I(C;S)$. If $I(C;\overline{S}|S)=0$ holds, $I(C;S)$ is maximized.  Accordingly, by Eq.(\ref{eq_14}) and Eq.(\ref{eq_16}), 
maximizing $I(S;C)$ is equivalent to maximizing the conditional likelihood of $C$ (i.e.  equivalent to maximizing $P(C|S)$).
Thus, using mutual information, the objective function of feature selection of Eq.(\ref{eq1_0}) can be re-formulated as  Eq.(\ref{eq3_2}) below.

\begin{equation}\label{eq3_2}
 S^*  =  \mathop{\arg\max}_{S\subset F}I(C;S)
\end{equation}



In the following, we will show that the feature set $S^*$ defined in Eq.(\ref{eq3_2}) is the set of features that leads to the minimal Bayes error rate.
For a given classification problem, the minimum achievable classification error by any classifier is called its Bayes error rate~\citep{fukunaga2013introduction}.  We choose the Bayes error rate for justifying Eq.(\ref{eq3_2}) since it is the tightest possible classifier-independent lower-bound by depending on predictor features and the class attribute alone. Fano and Hellman et. al.~\citep{Fano1961,tebbe1968uncertainty,hellman1970probability} proposed the lower and upper bounds on the Bayes error rate, which connect the Shannon conditional entropy~\citep{shannon2001mathematical} to the Bayes error rate.

Let $P_{err}$ represent the Bayes error rate, and the entropy $H(P_{err})$ is defined as
\begin{equation}\label{eq3_3}
H(P_{err}) = -P_{err}\log{P_{err}}-(1-P_{err})\log{(1-P_{err})}.
\end{equation}

Then given $C$ and $S$, Fano's lower bound of the Bayes error rate~\citep{Fano1961} is defined as Eq.(\ref{eq3_4}) below.
\begin{equation}\label{eq3_4}
H(C|S)\leq H(P_{err})+P_{err}\log{(K-1)}
\end{equation}

Let $H(P_{err})^{-1}$ be the inverse of $H(P_{err})$, the upper bound of the Bayes error rate for a binary classification problem (K=2) is given as Eq.(\ref{eq3_5}) below~\citep{tebbe1968uncertainty,hellman1970probability}.
\begin{equation}\label{eq3_5}
H(P_{err})^{-1}\leq P_{err}\leq 1/2H(C|S).
\end{equation}

Meanwhile, considering $H(C|F)=H(C)-I(C; F)$ and $ I(C;\overline{S}|S)=I(C; F)-I(C;S)$, Eq.(\ref{eq_14}) is re-written as Eq.(\ref{eq3_6}) below.

\begin{equation}\label{eq3_6}
\lim_{m\to\infty}{-\ell(C|S,D)}=KL(p(C|S)||q(C|S))+H(C|S)
\end{equation}

In Eq.(\ref{eq3_6}), with a large $m$, $KL(p(C|S)||q(C|S))$ will approach zero. Thus,  we conclude that minimizing $H(C|S)$, that is, the conditional entropy of the class attribute $C$ given the predictor feature set $S$, is equivalent to maximizing the conditional likelihood of $C$ or minimizing the Bayes error rate (from Eq.(\ref{eq3_5})). 
Since $H(C|S)=H(C)-I(C;S)$, maximizing $I(C;S)$ in Eq.(\ref{eq3_2}) equals to minimizing the upper bound of $H(C|S)$, i.e. the upper bound of $P_{err}$. This thus justifies that the feature set selected by Eq.(\ref{eq3_2}) for classification will best facilitate minimizing the Bayes error rate. Eq.(\ref{eq3_7}) illustrates the relationships among $I(C; S)$, $P_{err}$, and $L(C|S,D)$ where both ``$<=>$"  denote "equivalent to", respectively.
\begin{equation}\label{eq3_7}
 \arg\min_{S\subset F}{P_{err}(S)}\ <=>\arg\max_{S\subset F}{I(C; S)}\ <=>\arg\max_{S\subset F}{L(C|S,D)}
\end{equation}

\subsection{The objectives of causal and non-causal feature selection are the same}\label{sec4-2}

In this section, we will demonstrate that the Markov blanket of $C$ ($MB(C)$) is the feature set that maximizes Eq.(\ref{eq3_2}), and the set of strongly relevant features aimed by non-causal feature selection.
\begin{lemma}~\citep{pearl2014probabilistic}
 $\forall S\subset F\setminus MB(C),\ P(C|MB(C),S)=P(C|MB(C))$.

\label{lem3-1}
\end{lemma}

\begin{lemma} $I(X;Y)\geq 0$ with equality if and only if $P(X,Y)=P(X)P(Y)$.
\label{lem3-2}
\end{lemma}
\begin{lemma} $I(X;Y|Z)\geq 0$ with equality if and only if $P(X,Y|Z)=P(X|Z)P(Y|Z)$.
\label{lem3-3}
\end{lemma}
Clearly, by Eq.(\ref{eq_4}) and Eq.(\ref{eq_5}), Lemmas~\ref{lem3-2} and~\ref{lem3-3} hold. Then according to Lemmas~\ref{lem3-1} to~\ref{lem3-3}, Theorem~\ref{the3-1} below illustrates that $MB(C)$  is the solution to Eq.(\ref{eq3_2}).
\begin{theorem}
$\forall S\subset F$, $I(C;MB(C))\geq I(C;S)$ with equality if and only if $MB(C)=S$.

\textit{Proof:}
in the proof, we use $MB$ to represent $MB(C)$.

Case 1: $\forall S\subseteq F\setminus MB$,
by Eq.(\ref{eq_5}), we have:
\begin{equation}
\nonumber
\begin{array}{ll}
I(C;S|MB)=E_{\{C,S, MB\}}\log\frac{P(C,S|MB)}{P(C|MB)P(S|MB)}.
\end{array}
\end{equation}
As $P(C, S|MB)=P(C|MB)P(S|MB)$, $I(C;S|MB)=0$.
By the chain rule, $I((S,MB);C)=I(C;MB)+I(C;S|MB)=I(C;S)+I(C;MB|S)$.
Since $I(C;S|MB)=0$, $I(C;MB)=I(C;S)+I(C;MB|S)$. By Lemmas~\ref{lem3-2} and~\ref{lem3-3}, we get that $\forall S\subseteq F\setminus MB$, $I(C;MB)>I(C;S)$.

Case 2: $\forall S\subseteq MB$ and let $S'=MB\setminus S$, by $I(C;MB)-I(C;S)=I(C;S\cup S')-I(C;S)=I(C;S)+I(C;S'|S)-I(C;S)=I(C;S'|S)$, then $I(C;MB)\geq I(C;S)$ holds with equality if $S$ equals to $MB$.

Case 3: Let $S'\subset MB$ and $S''\subset F\setminus MB$, and $S=S'\cup S''$, 
by Eq.(\ref{eq3-4}) below, $I(C;S|MB)=0$. Then by $I(C;MB)+I(C;S|MB)=I(C;S)+I(C;MB|S)$, in the case, $I(C;MB)>I(C;S)$.
\begin{equation}
\small
\begin{array}{lll}
\frac{P(C,S|MB)}{P(C|MB)P(S|MB)}
=\frac{P(C, S'', MB)}{P(C|MB)P(S'',MB)}
=\frac{P(C|S'',MB)P(S'',MB)}{P(C|MB)P(S'',MB))}
=1.
\end{array}
\label{eq3-4}
\end{equation}
By Cases 1 to 3, $I(C;MB)\geq I(C;S)$ with equality holds if $S$ equals to $MB$.
\boxend
\label{the3-1}
\end{theorem}

\begin{corollary}\label{cor3-2}
Under the faithfulness assumption, $\forall F_i\in F$, $F_i$ belongs to $MB(C)$, if and only if $F_i$ is a strongly relevant feature.

\textit{Proof:}
In the proof, we use $MB$  to represent $MB(C)$. $PC(C)$ denotes parents and children of $C$ and $SP(C)$ represents spouses of $C$. 

We firstly prove that if $F_i\in MB$, $F_i$ is a strongly relevant feature. Since $MB=PC(C)\cup SP(C)$ and $PC(C)\cap SP(C)=\emptyset$, then (1) $\forall F_i\in PC(C)$ and  $\forall S\subseteq F\setminus F_i$, by Proposition~\ref{pro2-1}, $I(F_i;C|S)>0$, and thus, $I(F_i;C|F\setminus F_i)>0$ holds; (2) $\forall F_i\in SP(C)$ via child $F_\omega\in PC(C)$, by Proposition~\ref{pro2-2}, there exists a $S\subset F\setminus F_i$ such that $I(F_i;C|S)=0$ but $I(F_i;C|S\cup\{F_\omega\})>0$. Then, $\forall F_j\in F\setminus\{F_\omega,F_i\}$, $I(F_i;C|F\setminus F_j)=I(F_i;C|\{S\cup F_\omega, F\setminus \{S\cup F_\omega\cup F_i\}\})$. So if $F_i\in SP(C)$, $I(F_i;C|F\setminus F_j)>0$ holds. By Proposition~\ref{def2-9}, $F_i$ is a strongly relevant feature.

We now prove that a strongly relevant feature of $C$ must be in $MB$.  If $F_i$ is a strongly relevant feature, by Proposition~\ref{def2-9}, $I(F_i;C|F\setminus F_i)>0$.  Assume $F_i\notin MB$,  $S'=F\setminus\{F_i\}\cup MB$, and $S=F\setminus F_i=MB\cup S'$, we have:
\begin{equation}
\begin{array}{lll}
I(F_i;C|F\setminus F_i)&=&I(F_i;C|S)\\
           &=&E_{\{C,S, F_i\}}\log\frac{P(C,F_i|S)}{P(C|S)P(F_i|S)}\\
               &=&E_{\{C,S, F_i\}}\log\frac{P(C,F_i, S)}{P(C|S)P(F_i|S)P(S)}\\
              &=&E_{\{C,S, F_i\}}\log\frac{P(C|F_i, S)P(F_i|S)}{P(C|S)P(F_i|S)}\\
              &=&E_{\{C,S, F_i\}}\log\frac{P(C|F_i, S)}{P(C|S)}\\
              &=&E_{\{C,S', MB, F_i\}}\log\frac{P(C|F_i, S',MB)}{P(C|S',MB)}\\
             &=&0.
\end{array}
\label{eq3-1}
\end{equation}
This makes a contrary, and thus $F_i\in MB(C)$.
\boxend
\end{corollary}

Accordingly, given a dataset $D$ defined on $F\cup C$,  by the analysis above, we show that $MB(C)$ maximizes the objective function in Eq.(\ref{eq3_2}) and it is the same as the set of strongly relevant features.



\section{Causal and non-causal feature selection: assumptions and approximations}\label{sec5}

For both causal and non-causal feature selection methods, finding a subset $S$ that maximizes $I(S;C)$ (i.e. solving the objective function in Eq.(\ref{eq3_2}))  is a challenging combinatorial optimization problem. An exhaustive search will be of $O(2^n)$ time complexity. Although restricting the maximum size of $S$ to $\varsigma$ ($\varsigma<n$) will reduce the time complexity to $O(\varsigma^n)$ where $\varsigma^n$ is the number of all subsets of $F$ containing $\varsigma$ or less features, the computational cost will still be high.
Therefore, both causal and non-causal feature selection methods have adopted a greedy strategy by considering features one by one to optimize Eq.(\ref{eq3_2})~\citep{aliferis2010local1,balagani2010feature,brown2012conditional}. That is, at each iteration, given the set $S$ currently selected, choose $X^*\in F\setminus S$ such that 
\begin{equation}
\begin{array}{lll}
X^*&=&\mathop{\arg\max}_{X\in F\setminus S}I(S\cup X;C)\\
&=&\mathop{\arg\max}_{X\in F\setminus S}\{I(S;C)+I(X;C|S)\}
\end{array}
\label{eq5-01}
\end{equation}

As for all $X\in F\setminus S$, the first item in Eq.(\ref{eq5-01}) is the same, finding $X^*$ becomes solving the following optimization problem:
\begin{equation}
X^*=\mathop{\arg\max}_{X\in F\setminus S}I(X;C|S)\\
\label{eq5-012}
\end{equation}

However, in Eq.(\ref{eq5-012}), when the size of $S$ increases, computing the multidimensional mutual information becomes impractical because it demands a large number of training samples, exponential in the number of features in $S$.
To tackle this challenge, different feature selection methods make different assumptions on the interactions (or dependency) between features in the underlying data distributions for the calculation of $I(X;C|S)$.


As described previously, a Bayesian network provides a representation of the probabilistic dependence among a set of variables under consideration. This provides us the opportunity to unify the dependence assumptions made by the feature selection methods under the  Bayesian network framework.
In this paper, we propose a structure assumption approach to understanding the assumptions made by causal and non-causal feature selection methods and how these different levels of structural assumptions lead to the different approximations in their search for the solutions to Eq.(\ref{eq5-012}).

In the following, firstly Section~\ref{sec5-0} provides a summary of our findings on the structural assumptions and how they are related to the approximations, then in Sections~\ref{sec5-1} and~\ref{sec5-2} we discuss the findings in detail by analyzing the assumptions and approximations made by the commonly used non-causal and causal feature selection methods.

\subsection{Summary of findings}\label{sec5-0}

\subsubsection{Structural assumptions and search strategies}

\begin{figure}[t]
\centering
\includegraphics[height=1in,width=4.5in]{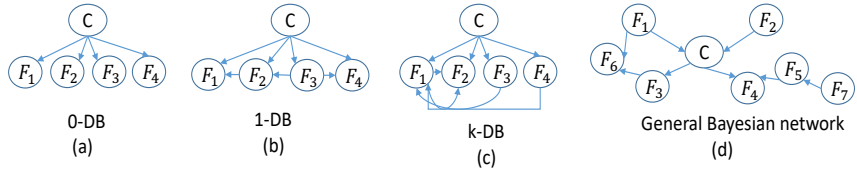}
\caption{An illustration of the Bayesian network structures corresponding to the structural assumptions made by the non-causal and causal feature selection methods}
\label{fig6-2}
\end{figure}

As illustrated in Figure~\ref{fig6-2}, we have found that the dependence/independence relationships among features assumed by both causal and non-causal feature selection methods can be represented as different restrictions to the structure of the Bayesian network model of the set of variables under study.
Based on the assumed Bayesian network structures, causal and non-causal  methods select the subset of features, $S\subset F$, with the conditional likelihood of the class attribute $C$ given $S$, $P(C|S)$ as close to $P(C|F)$ as possible. 
\begin{figure}
\centering
\includegraphics[height=8.3in]{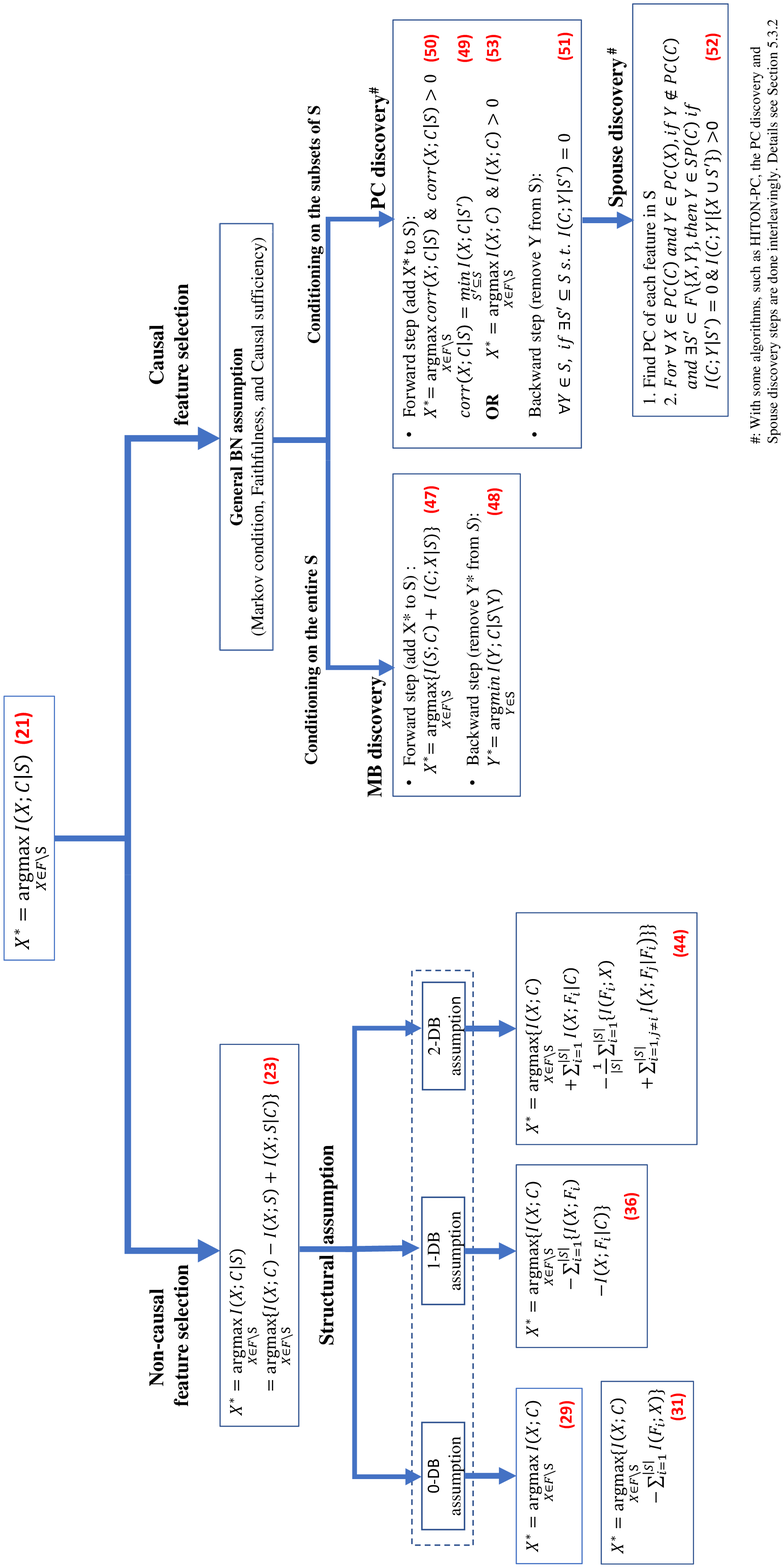}
\caption{A road map of how causal and non-causal feature selection searches for  $X^*$ in  Eq.(\ref{eq5-012})}
\label{fig4-001}
\end{figure}

Figure~\ref{fig4-001} summarizes the Bayesian network structure assumptions and search strategies used by causal and non-causal feature selection methods for the calculation of $I(X;C|S)$. 
The number after each equation in Figure~\ref{fig4-001} are the same as the equation numbers given in Sections~\ref{sec5-1} and~~\ref{sec5-2}.
From Figure~\ref{fig4-001}, we see that a non-causal feature selection method firstly decomposes the multidimensional mutual information $I(X;C|S)$ into three terms $\{I(X;C)-I(S;X)+I(S;X|C)\}$ (See Eq.(\ref{eq_28_2})), then calculates the multidimensional mutual information $\{-I(S;X)+I(S;X|C)\}$ using linear combination of low-order mutual information terms based on the respective naive Bayesian network assumption made on the dependence/independence between features. We call the assumptions made by non-causal feature selection methods the series of naive Bayesian network assumptions, because the assumptions can be represented by the family of Bayesian networks with the restricted structures as illustrated in Figures~\ref{fig6-2} (a), (b) and (c). For these naive Bayesian network structures, the class attribute has no parents while all the features each can only have a fixed number of parents, denoted as $k$-dependency (or $k$-DB) assumptions, where each feature can have at most other $k$ features as its parents (details in Section~\ref{sec5-1}).

Causal feature selection methods assume that one can learn from the given dataset a (general) Bayesian network without structural restrictions (as the example in Figure~\ref{fig6-2} (d)), and in the learnt Bayesian network,  $X^*$ in Eq.(\ref{eq5-012}) is a feature in the MB of the class attribute. Therefore, as shown in  Figure~\ref{fig4-001}, causal feature selection does not decompose $I(X;C|S)$ for the use of any structural assumptions, and the  assumptions made by causal feature selection are only those for a general Bayesian network and its learning, i.e. the Markov condition (Definition~\ref{def2-2}), the faithfulness (Definition~\ref{def2-4}), and causal sufficiency (Definition~\ref{def2-yk1}) assumptions. Unlike the non-causal feature selection methods, these assumptions do not pose any structural restrictions on a Bayesian network learnt from data (thus called the general Bayesian network assumptions in this paper).

\subsubsection{Linking the assumptions with approximations}

\begin{figure}[t]
\centering
\includegraphics[height=3in,width=6in]{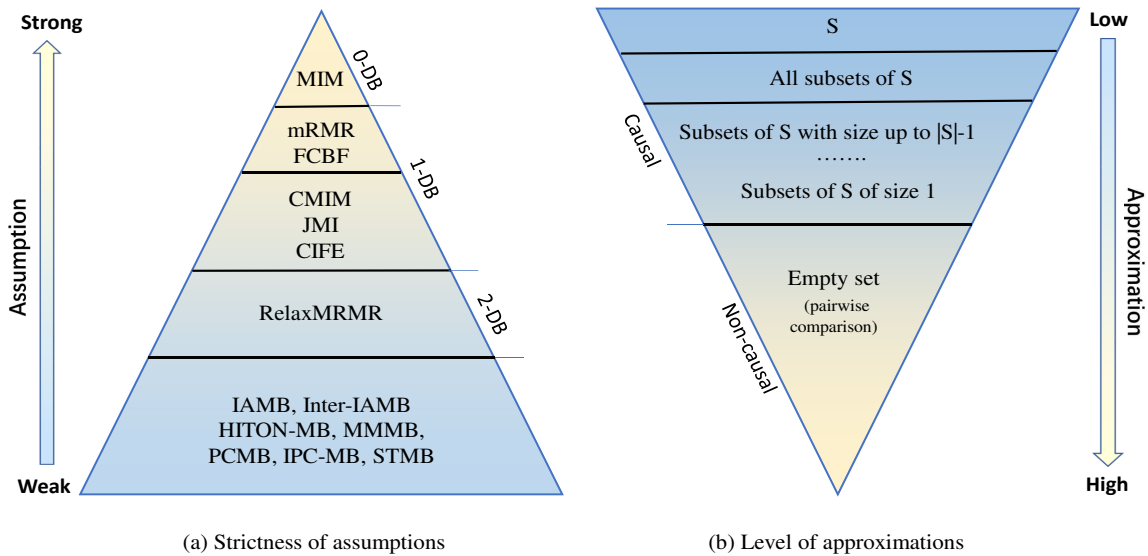}
\caption{Strictness of structural assumptions and the corresponding level of approximations taken by causal and non-causal feature selection methods when calculating $I(X;C|S)$ (a) the strictness of structural assumptions in terms of maximum number of parents a feature can have (excluding the class attribute). Names of typical methods are shown. 
(b) the level of approximations in terms of the size of conditioning set used in the calculation.}
\label{fig4-0}
\end{figure}

We use the pyramid in Figure~\ref{fig4-0} (a) to visualize the difference in the strictness of the structural assumptions made by the different feature selection methods. 
We see that causal feature selection methods make the weakest assumptions (no restrictions on the structures of the Bayesian network), while the non-causal feature selection methods make assumptions with different levels of strictness in terms of the maximum number of parents that a feature can have in addition to the class attribute (the value of $k$ in Figure~\ref{fig4-0} (a)).

As a result of the differences in the strictness of the structural assumptions, the degree of the corresponding approximations taken by the feature selection methods in their calculation of the multidimensional mutual information ($I(X;C|S)$) are different, and they can be visualized using an upside down pyramid (Figure~\ref{fig4-0} (b)).  Causal feature selection methods, since having had no structural restrictions, take less approximations by calculating higher order mutual information between $X$ and $C$ conditioning on all or a subset of the already selected features $S$ (details of the conditioning sets are to be discussed in Section~\ref{sec5-2}).  Referring back to Figure~\ref{fig4-001}, the non-causal feature selection methods eventually only look at the pairwise mutual information between $X$ and $C$ without conditioning on other features.  

Therefore, in theory, the feature set obtained by a causal feature selection methods is closer to the optimal feature set, i.e. the MB of the class attribute. However, as we will see in later sections, in practice, causal feature selection does not always outperform non-causal feature selection, because the number of samples required by causal feature selection can be exponential in the number of features in $S$.

\subsubsection{Causal interpretation and non-causal feature selection}

By representing the dependency between features and the class attribute using Bayesian network structures, we present a causal interpretation of the features selected by non-causal methods.

We have found that the non-causal feature selection methods prefer features within $MB(C)$ to the features not in $MB(C)$, which confirms that strongly relevant features belong to $MB(C)$ (i.e. Corollary~\ref{cor3-2}). This finding provides a causal interpretation of the output of the non-causal feature selection methods and explains why non-causal feature selection also can achieve excellent classification results. This also provides a novel perspective to understand the relations between the two types of feature selection methods, and may motivate researchers to use the cross-pollination between causal and non-causal feature selection methods to develop novel methodologies promising to scalable local-to-global causal structure learning and feature selection with theoretical guarantees.



\subsection{Non-causal feature selection:  assumptions and approximations}\label{sec5-1}
In this section, we will explore in detail the assumptions made by non-causal feature selection under the naive Bayesian network framework, and under the assumptions how the major existing non-causal feature selection algorithms produce the same result as Eq.({\ref{eq5-012}). 

By $I(X;S;C)=I(X;S)-I(X;S|C)=I(X;C)-I(X;C|S)$, we have:
\begin{equation}\label{eq_28_2}
I(X;C|S)=I(X;C)-I(X;S)+I(X;S|C).
\end{equation}

The three terms on the right side of Eq.(\ref{eq_28_2}) have the following interpretation:
\begin{itemize}
\item $I(X;C)$ corresponds to the relevancy of $X$ to $C$.
\item $I(X;S)$ represents the redundancy of $X$ with respect to $S$.
\item 
$I(X;S|C)$ indicates the class-conditional relevance, which considers the situation where a feature provides more predictive information by jointly with another feature than by itself with respect to $C$. Since $I((S,X);C)=I(S;C)+I(X;C|S)$, $I(X;C|S)=I((S,X);C)-I(S;C)$. In Eq.(\ref{eq_28_2}), when $I(X;S)>I(X;S|C)$,  $I(X;C|S)<I(X;C)$ holds, and thus $I((S,X);C)<I(S;C)+I(X;C)$. This means that $X$ contains redundant information about $C$ when we add $X$ to $S$. When $I(X;S)<I(X;S|C)$, $I(X;C|S)>I(X;C)$ holds, and thus $I((S,X);C)>I(S;C)+I(X;C)$. This indicates that $X$ and $S$ have a positive interaction and $I((S,X);C)$ provides more information than $I(S;C)+I(C; X)$.
\end{itemize}

By Eq.(\ref{eq_28_2}),  Eq.({\ref{eq5-012}) can be re-written as
\begin{equation}\label{eq_29}
X^*=\mathop{\arg\max}_{X\in F\setminus S}\{I(X;C)-I(X;S)+I(X;S|C)\}
\end{equation}

To reduce computational costs in the search for $X^*$ in Eq.(\ref{eq_29}), different non-causal feature selection methods make different assumptions, and thus adopt different level of approximations when calculating $I(X;S)$ and $I(X;S|C)$ by using a linear combination of low-order mutual information terms.

In the following, we will explore these assumptions and approximations of Eq.(\ref{eq_29}).
Using a general Bayesian all features and the class attribute, we have
\begin{equation}\label{eq6-0}
P(C|F)\propto P(C|Pa(C))\prod_{i=1}^{n}P(F_i |Pa(F_i)).
\end{equation}


A naive Bayesian network is a restricted Bayesian network, which considers  the class attribute $C$ as a special variable that has no parents and each of the remaining variables in the network only has  the class attribute $C$ and a fixed number of other features as its parents. Let $k$ represent the maximum number of parents (excluding the class attribute) a feature can have, we call the naive Bayesian network a $k$-dependency ($k$-DB) naive Bayesian network.
A $0$-DB network (as illustrated in Figure~\ref{fig6-2} (a)) is the commonly know naive Bayes (NB) network ~\citep{maron1960relevance,minsky1961steps}.
A NB network assumes that each variable only has one parent, i.e. $C$, and all features are conditionally independent given $C$. 
A $1$-DB network (as illustrated in Figure~\ref{fig6-2} (b)) is known as a Tree-Augmented Naive (TAN) Bayes network, which allows each variable to have at most one other feature in addition to $C$ as its parent.
A $2$-DB network ((see an example in Figure~\ref{fig6-2} (c)) relaxes NB's and TAN's independence assumptions by allowing each feature  to have a maximum of  two other features as parents to generalize to higher degrees of variable interactions. 

Let $ncl\_pa(F_i)$ denote the set of  parents of $F_i$ excluding the class attribute $C$, in a $k$-DB naive Bayesian network, Eq.(\ref{eq6-0}) becomes

\begin{equation}
P(C|F)\propto P(C)\prod_{i=1}^{n}P(F_i |C, ncl\_pa(F_i)),\ | ncl\_pa(F_i)|=k\ \&\ |pa(C)|=0.
\label{eq6-1}
\end{equation}
	
\subsubsection{Approximations under $0$-DB(NB) structural assumptions}\label{sec5-01}

The following NB network assumption ($k=0$) is often made by non-causal feature selection methods.

\textbf{Assumption 1.} In a NB network, $\forall F_i, F_j\in F$ and $i\neq j$, $F_i$ and $F_j$ are assumed to be conditionally independent given the class attribute $C$, that is, $P(F_i,F_j|C) = P(F_i|C)P(F_j|C)$. 

By Assumption 1, Eq.(\ref{eq6-1}) is transformed  into
\begin{equation}
 P(C|F)\propto P(C)\prod_{i=1}^{n}P(F_i |C), \ |ncl\_pa(F_i)|=0\ \&\ |pa(C)|=0
 \label{eq6-6}
\end{equation}

By Assumption 1 and Eq.(\ref{eq6-6}), in Eq.(\ref{eq_29}), the class-conditional relevancy $I(X;S|C)$ is calculated as Eq.(\ref{eq6-7}) as follows.
\begin{equation}
\begin{array}{lll}
 I(X;S|C)&=&E_{x,s,c}\log{\frac{P(X,S|C)}{P(S|C)P(X|C)}}\\
&=&E_{x,s,c}\log{\frac{P(S|C)P(X|C)}{P(S|C)P(X|C)}}\\
&=&0
\end{array}
\label{eq6-7}
\end{equation}

Then under Assumption 1 and Eq.(\ref{eq6-7}), Eq.(\ref{eq_29}) becomes
\begin{equation}
\mathop{\arg\max}_{X\in F\setminus S}\{I(X;C)-I(X;S)+I(X;S|C)\}=\mathop{\arg\max}_{X\in F\setminus S}\{I(X;C)-I(X;S)\}.
\label{eq6-8}
\end{equation}

Since the redundancy term $I(X;S)=H(S)-H(S|X)$, and by the chain rule of entropy, we have
$H(S|X) =\sum_{F_i\in S}H(F_i|F_{i-1},\cdots,F_1,X)$. If we further employ  Assumption 2 below to restrict the interactions between a feature in $S$ and a feature in $F\setminus S$, in Eq.(\ref{eq_29}), $I(X;S)=0$ holds.

\textbf{Assumption 2.} For $\forall F_i\in S$ and $\forall F_j\in F\setminus S$, $P(F_i,F_j) =P(F_i)P(F_j)$.

By Assumptions 1 and 2, the objective function in Eq.(\ref{eq_29}) is simplified to the following, which is only based on the mutual information between a feature and the class attribute: 
\begin{equation}
X^*=\mathop{\arg\max}_{X\in F\setminus S}I(X;C).
\label{eq5-02}
\end{equation}

The objective in Eq.(\ref{eq5-02}) is the mutual information maximization (MIM) criterion initially presented in~\citep{lewis1992feature}.

Assumption 2 is a strong assumption that  the features in $S$ and the features in $F\setminus S$ are pairwise independent. To deal with the redundancy between features, we discuss Assumption 3 below, which is a less restrictive than Assumption 2.

\textbf{Assumption 3.} The selected features in $S$ are conditionally independent given an unselected feature $X\in F\setminus S$ , that is,  $P(S|X) = \prod_{i=1}^{|S|}P(F_i|X)$ ($F_i\in S$).

Since $I(X;S)=H(S)-H(S|X)$, by the chain rule and Assumption 3, we have
\begin{equation}
\begin{array}{lll}
 I(X;S)&=&H(S)-\sum_{i=1}^{|S|}H(F_i|X)\\
                  &=&H(S)-\sum_{i=1}^{|S|}H(F_i)+\sum_{i=1}^{|S|}I(F_i;X).
\end{array}
\label{eq5-03}
\end{equation}

Since at each time, $\forall X\in F\setminus S$, the first two terms in Eq.(\ref{eq5-03}) are the same, then  $I(X;S)$ is decomposed into a sum of  pairwise mutual information terms. Further based on Assumption 1, $I(X;S|C)=0$, then the objective function in Eq.(\ref{eq_29}) becomes:
\begin{equation}
X^*=\mathop{\arg\max}_{X\in F\setminus S}\{I(X;C)-\sum_{i=1}^{|S|}I(F_i;X)\}.
\label{eq5-04}
\end{equation}

Eq.(\ref{eq5-04}) is the criterion of ``max-relevance and min-redundancy''~\citep{peng2005feature}.  Based on Eq.(\ref{eq5-04}), Battiti~\citep{battiti1994using}
presents the following Mutual Information Feature Selection (MIFS) criterion:
\begin{equation}
X^*=\mathop{\arg\max}_{X\in F\setminus S}\{I(X;C)-\beta\sum_{i=1}^{|S|}I(F_i;X)\}
\label{eq5-05}
\end{equation}

$\beta\in [0,1]$ in the MIFS criterion is a penalty for balancing the relevance and redundancy terms. When $\beta=0$, Eq.(\ref{eq5-05}) becomes Eq.(\ref{eq5-02}), that is, the MIM criterion. As  $\beta=1$,  Eq.(\ref{eq5-05}) is reduced to Eq.(\ref{eq5-04}). If $\beta=1/|S|$, Eq.(\ref{eq5-05}) becomes
\begin{equation}
X^*=\mathop{\arg\max}_{X\in F\setminus S}\{I(X;C)-\frac{1}{|S|}\sum_{i=1}^{|S|}I(F_i;X)\}
\label{eq5-06}
\end{equation}


Eq.(\ref{eq5-06}) is the mRMR (max-Relevance and Min-Redundancy) criterion presented in~\citep{peng2005feature}. 
Meanwhile, from Eq.(\ref{eq5-06}), we can see that as the size of $S$ increases, Eq.(\ref{eq5-06}) will tend asymptotically towards  Eq.(\ref{eq5-02}).

There are other feature selection methods based on the idea of max-relevance and min-redundancy shown in Eq.(\ref{eq5-04}), such as a representative algorithm, Fast Correlation Based Filter (FCBF)~\citep{yu2004efficient}.  FCBF divides the``max-relevance and min-redundancy''  criterion into two steps,  that is, the forward step (max-relevance) and backward step (min-redundancy).
\begin{itemize}
\item Forward step: FCBF selects a subset of features $S$ that $\forall X\in S$, $I(C;X)>0$, then sorts the features in $S$ by their mutual information with $C$ in descending order.

\item Backward step: beginning with the first feature $X\in S$, if $\exists Y\in S\setminus X$ such that  $I(X;Y)>I(X;C)$, then it removes $Y$ from $S$ as a redundant feature to $X$. The FCBF algorithm is terminated until the last feature in $S$ is checked.
\end{itemize}

At the forward step, FCBF only selects features that are relevant to $C$, and this implies Assumption 1. The backward step  implies Assumption 3. At the  backward step, for $X$, $Y\in S$, if  $I(X;C)>I(Y;C)$  and $I(X;Y)>I(X;C)$, then $Y$ can be removed from $S$. FCBF does not need to specify the number of selected features in advance.
Instead, FCBF uses a threshold $\delta\ (\delta>0)$ at the forward step and keeps features satisfying $I(C;X)\geq\delta$.

\subsubsection{Approximations with $1$-DB(TAN) structural assumptions}\label{sec5-02}

Under Assumption 1, in Eq.(\ref{eq_29}), $I(X;S|C)=0$ holds.  A TAN Bayesian network relaxes Assumption 1 to allow each feature to be dependent on one other feature in addition to $C$ and makes the following assumption. Assumption 4 states that the features within $S$ are class-conditionally independent given an unselected feature $X\in F\setminus S$ and $C$.

\textbf{Assumption 4.} $\forall F_i, F_j\in S$ and $i\neq j$, $F_i$ and $F_j$  are assumed to be conditionally independent given an unselected feature $X\in F\setminus S$ and $C$, that is, $P(F_i,F_j|X,C) = P(F_i|C,X)P(F_j|C,X)$. 

Thus for a TAN Bayesian network, Eq.(\ref{eq6-1}) becomes:
\begin{equation}
P(C|F)\propto P(C)\prod_{F_j\in F,\ F_i\in F\setminus\ F_j}P(F_i |C, F_j), \ |ncl\_pa(F_i)|=1\ \&\ |pa(C)|=0.
\label{eq5-07}
\end{equation}

Then by the chain rule,  we get $H(S|X,C)=\sum_{F_i\in S}H(F_i|X,C)$.
By Eq.(\ref{eq6-7}), $I(X;S|C)=0$ only and if only Assumption 1 holds, and thus by Assumption 4, $I(X;S|C)$ can be decomposed as follows.
\begin{equation}
\begin{array}{lll}
I(X;S|C)&=&H(S|C)-H(S|X,C)\\
&=&H(S|C)-\sum_{F_i\in S}H(F_i|X,C)\\
&=&H(S|C)-\sum_{F_i\in S}\{H(F_i|C)-I(F_i;X|C)\}
\end{array}
\label{eq5-08}
\end{equation}

Since $H(S|C)-\sum_{F_i\in S}H(F_i|C)$ in Eq.(\ref{eq5-08}) is the same for $\forall F_i\in S$. Meanwhile, assuming that Assumption 3 holds for feature interactions between the selected features in $S$ and the unselected feature in $F\setminus S$, then by Eq.(\ref{eq5-03}) (under Assumption 3) and Eq.(\ref{eq5-08}) (under Assumption 4),  Eq.(\ref{eq_29}) becomes: 
\begin{equation}
X^*=\mathop{\arg\max}_{X\in F\setminus S}\{I(X;C)-\Sigma_{F_i\in S}I(X;F_i)+\Sigma_{F_i\in S}I(X;F_i|C)\}
\label{eq5-09}
\end{equation}

Brown et al.~\citep{brown2012conditional} have proposed that many mutual information-based non-causal feature selection methods can fit within the following parameterized criterion. $\beta$ and $\gamma$ play the role of balancing factors (in general $\beta\in [0,1]$ and $\gamma\in [0,1]$).
\begin{equation}
X^*=\mathop{\arg\max}_{X\in\{F\setminus S\}}\{I(X;C)-\beta\sum_{F_i\in S}I(X;F_i)+\gamma \sum_{F_i\in S}I(X;F_i|C)\}
\label{eq5-10}
\end{equation}

If $\beta=1/|S|$ and $\gamma=1/|S|$, then we have:
\begin{equation}
X^*=\mathop{\arg\max}_{X\in F\setminus S}\{I(X;C)-\frac{1}{|S|}\Sigma_{F_i\in S}I(X;F_i)+\frac{1}{|S|}\Sigma_{F_i\in S}I(X;F_i|C)\}
\label{eq5-092}
\end{equation}

Using Eq.(\ref{eq5-092}) for feature selection, the representative algorithm is the JMI algorithm~\citep{yang2000data}.
If $\beta=1$ and $\gamma=1$, Eq.(\ref{eq5-10}) is reduced to Eq.(\ref{eq5-09}) used by the CIFE algorithm~\citep{lin2006conditional}.
The CMIM method~\citep{fleuret2004fast} adopts an objective function as follows.
\begin{equation}
X^*=\mathop{\arg\max}_{X\in F\setminus S}\{I(X;C)-\mathop{\max}_{F_i\in S}\{I(X;F_i)-I(X;F_i|C)\}\}
\label{eq5-11}
\end{equation}

\subsubsection{Approximations with $2$-DB structural assumptions}\label{sec5-03}

To deal with a higher-order dependency between features, the recent work in~\citep{vinh2016can} calculates $I(X;S)$ in Eq.(\ref{eq_29}) by exploring the $2$-DB structure assumptions.

The $2$-DB structure relaxes NB's and TAN's independence assumptions by allowing each feature to have at most two features as parents, i.e., $ |ncl\_pa(F_i)|=2$, in addition to $C$, and makes the following assumptions.

\textbf{Assumption 5a.}  $\forall F_i\in S$ and $\forall F_j\in S\ (i\neq j)$ are assumed to be conditionally independent given an unselected feature $X\in F\setminus S$ and any feature $Y\in F\setminus\{F_i\cup F_j\}$, that is, $P(F_i,F_j|X,Y) = P(F_i|X,Y)P(F_j|X,Y)$. 

\textbf{Assumption 5b.} For $\exists F_j\in S$ and $\forall F_i\in F\setminus F_j$ are conditionally independent given an unselected feature $X\in F\setminus S$, that is, $P(F_j,F_i|X) = P(F_i|X)P(F_j|X)$.

 Thus, with a $2$-DB structure, Eq.(\ref{eq6-1}) is transformed  into Eq.(\ref{eq5-13}).
\begin{equation}
P(C|F)\propto P(C)\prod_{i=1 (F_i\in F\setminus\{F_j\cup F_\omega\})}^{n}P(F_i |C, F_j,F_\omega)),\ |ncl\_pa(F_i)|=2\ \&\ |pa(C)|=0
\label{eq5-13}
\end{equation}

With the structure assumptions, the redundancy term $I(X;S)$ in Eq.(\ref{eq_29}) is computed as follows. 
Since $I(X;S)=H(S)-H(S|X)$，
under Assumptions 5a and  5b,  $H(S|X)$ is calculated as follows.
\begin{equation}
\begin{array}{lll}
H(S|X) &=&-\sum_{i=1}^{|S|}\sum_{F_1,\cdots, F_i,X}P(F_1,\cdots, F_i,X)\log P(F_i|F_{i-1},\cdots,F_1,X)\\
                           &=&P(F_1,\cdots, F_j,X)\log P(F_j|F_{j-1},\cdots,F_1, X)\\
                           && +\sum_{i=1 (i\neq j)}^{|S|-1}P(F_{i-2},\cdots, F_1,F_j,X)\log P(F_i|F_{i-2},\cdots,F_1, F_j, X)\\
                           &=&H(F_j|X)+\sum_{i=1,i\neq j}^{|S|-1}H(F_i|F_j,X)
\end{array}
\label{eq5-14}
\end{equation}

By Eq.(\ref{eq5-14}),  $I(X; S)$ is decomposed as Eq.(\ref{eq5-15}) as follows. 
\begin{equation}
\begin{array}{lll}
I(X; S)&=&H(S)-H(S|X)\\
                  &=&H(S)-\{H(F_j|X)+\sum_{i=1,i\neq j}^{|S|-1}H(F_i|F_j,X)\}\\
                   &=&H(S)-H(F_j)+I(F_j;X)-\sum_{i=1,i\neq j}^{|S|-1}\{H(F_i|F_j)-I((F_i,X|F_j)\}
\end{array}
\label{eq5-15}
\end{equation}

In Eq.(\ref{eq5-15}), at each iteration, for $\forall X\in F\setminus S$, $H(S)-H(F_j)-\sum_{i=1,i\neq j}^{|S|-1}H(F_i|F_j)$ is the same. Meanwhile, to avoid the need of checking which feature in $S$ satisfying Assumption 5b, by averaging over all features in  $S$, we have
\begin{equation}\label{eq5-16}
\begin{array}{lll}
X^*&=&\mathop{\arg\max}_{X\in F\setminus S}\{I(X;C)+H(S|C)-H(S|C,X)\\
&&-\frac{1}{|S|}\Sigma_{F_i\in S}\{I(X;F_i)+\Sigma_{F_j\in S, i\neq j}I(X;F_j|F_i)\}
\end{array}
\end{equation}

If we employ Assumption 4 for $I(X;S|C)$ in Eq.(\ref{eq5-16}), we get the following objective function in Eq.(\ref{eq5-12}) used by the RelaxMRMR algorithm proposed by~\citep{vinh2016can}.
\begin{equation}\label{eq5-12}
\begin{array}{lll}
X^*&=&\mathop{\arg\max}_{X\in F\setminus S}\{I(X;C)+\Sigma_{F_i\in S}I(X;F_i|C)\\
&&-\frac{1}{|S|}\Sigma_{F_i\in S}\{I(X;F_i)+\Sigma_{F_j\in S, i\neq j}I(X;F_j|F_i)\}\}
\end{array}
\end{equation}

\subsubsection{Time complexity and sample requirement of non-causal feature selection}\label{sec5-05}

In this section, we will analyze the time complexity and sample requirement of non-causal feature selection methods. Under the $k$-DB structural assumption, the most common family of non-causal feature selection methods decompose Eq.(\ref{eq5-012}) into different objective functions, such as Eq.(\ref{eq5-02}), Eq.(\ref{eq5-04}), Eq.(\ref{eq5-09}), or Eq.(\ref{eq5-12}), in a linear combination of low-order mutual information terms. By these objective functions, non-causal feature selection methods greedily select the $\psi$ features with the highest mutual information scores~\citep{guyon2003introduction}. The time complexity of non-causal feature selection methods depends on $\psi$. 
Solving Eq.(\ref{eq5-12}) requires $O(\psi^3n)$ mutual information computations.  Eq.(\ref{eq5-04}) and Eq.(\ref{eq5-09}) need $O(\psi^2n)$ pairwise comparisons, while Eq.(\ref{eq5-02}) (the MIM criterion) only requires $O(n)$ pairwise comparisons. However, how to determine a good value of the user-defined parameter $\psi$ for optimal feature selection is not an easy problem.

The sample requirement of a non-causal feature selection method depends on the number of samples needed to assure reliable computation of mutual information or independence tests.  With discrete data, $\chi^2$ (chi-square) test and $G^2$  test (a variant of chi-square test) are commonly used to determine the independence of two variables. For a reliable independence test between $X$ and $C$ given the current conditioning set $S$, the minimum number of data samples $N$ is:
\begin{equation}
 N\geq\xi\times r_X\times r_C\times  r_{S}
\label{eq7-01}
\end{equation}
where $r_X$ and $r_C$ represent the numbers of possibles values (i.e. levels) of $X$ and $C$ respectively, and $r_S=\prod_{i=1}^{|S|}r_{F_i}$, $F_i\in S$,  i.e. the multiplication of the numbers of possible values of all features in $S$. $\xi$ is often set to 5 as suggested by Agresti~\citep{agresti2011categorical}. 
As $\xi$ is a constant, the lower bound of the required data samples $N$ is only determined by $r_X$, $r_C$, and $r_{S}$ where $r_{S}$ plays the key role in (\ref{eq7-01}).

In the paper, since we formulate  feature selection using mutual information, Eq.(\ref{eq7-02}) below shows that the mutual information between two variables is proportional to the value of association of the two variables calculated by $G^2$ test ~\citep{yaramakala2004fast}, which guarantees the correctness of using  Eq.(\ref{eq7-01}) above to discuss the sample requirement of non-causal feature selection methods.

\begin{equation}
\frac{1}{2N}G^2(X;C ) = I(X;C)\ \&\ \frac{1}{2N}G^2(X;C|S) = I(X;C|S)
\label{eq7-02}
\end{equation}

To obtain the lower bounds of required samples of the non-causal feature selection methods, assume $X_{max}$, $Y_{max}$, and $W_{max}$ are the three features with the largest discrete values, then the minimum number of data samples required by Eq.(\ref{eq5-02}) (MIM), Eq.(\ref{eq5-04}) (MIFS, mRMR, and FCBF), Eq.(\ref{eq5-09}) (JMI and CMIM), and Eq.(\ref{eq5-12}) (RelaxMRMR) is bounded by $r_{X_{max}}\times r_C$, $r_{X_{max}}\times r_{Y_{max}}$, $r_{X_{max}}\times r_{Y_{max}}\times  r_C$, $r_{X_{max}}\times r_{Y_{max}}\times  r_{W_{max}}$, respectively. Since the existing major non-causal feature selection methods calculate $I(X;C|S)$ using linear combination of low-order mutual information terms (i.e. the size of $S$ in $r_{S}$ in (\ref{eq7-01}) is never bigger than 1), the sample requirement of non-causal feature selection is not high.

\subsubsection{Discussion}\label{sec5-04}

Let $X$ be the candidate feature under consideration, and $Y$ a previously selected feature. In Eq.(\ref{eq5-02}), Eq.(\ref{eq5-04}), Eq.(\ref{eq5-09}), and Eq.(\ref{eq5-12}), we can see that those methods only consider at most one of the selected features when evaluating $X$.
Therefore, in the following, by representing the interactions among the three variables $X$, $Y$, and $C$ (class attribute) using Bayesian network structures from Figures~\ref{fig4-1} (a) to~\ref{fig4-1} (g), firstly, we discuss some properties  between $X$, $Y$, and $C$, i.e. Properties~\ref{prop4-1} to~\ref{prop4-4} below. Secondly, with those properties, we will  investigate the causal interpretations of Eq.(\ref{eq5-02}), Eq.(\ref{eq5-04}), Eq.(\ref{eq5-09}), and Eq.(\ref{eq5-12}). Through the discussion, we will show that the major non-causal feature selection methods 
driven by the simplified objective functions shown in Eq.(\ref{eq5-02}), Eq.(\ref{eq5-04}), Eq.(\ref{eq5-09}), and Eq.(\ref{eq5-12})
 prefer direct causes, direct effects, and spouses of $C$ to the features which are not in $MB(C)$.
\begin{figure}
\centering
\includegraphics[height=1.5in,width=4in]{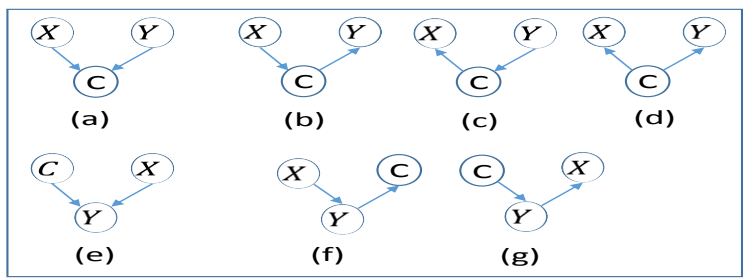}
\caption{The three-way causal interactions and non-causal feature selection}
\label{fig4-1}
\end{figure}

When $X$ and $Y$ are parents or children of $C$ as shown in Figures~\ref{fig4-1} (a) to (d), we have the following properties.

\begin{property}\label{prop4-1}
If $X$ and $Y$ are both direct causes (parents) of $C$, i.e. the class attribute $C$ is a common-effect of the two features, as shown in Figure~\ref{fig4-1} (a), then (1) $I(X;C)>I(X;Y)$, (2) $I(X;Y|C)\geq I(X;Y)$, and (3) $I(X;C)-I(X;Y)+I(X;Y|C)>0$.

\textit{Proof:} According to Proposition~\ref{pro2-2}, in the case shown in Figure~\ref{fig4-1} (a), $I(X;Y)=0$ holds. Clearly, $I(X;C)>I(X;Y)$ if $I(X;Y)=0$.  By $I(X;Y;C)=I(X;Y)-I(X;Y|C)=I(X;C)-I(X;C|Y)$, if $I(X;Y)=0$, $I(X;Y|C)\geq I(X;Y)$ and $I(X;C)-I(X;Y)+I(X;Y|C)>0$ hold.
\boxend
\end{property}

\begin{property}\label{prop4-2}

In the causal chain interaction cases in Figures~\ref{fig4-1} (b) to (c) or the common cause interaction case in Figure~\ref{fig4-1} (d), where $X$ or $Y$ is a direct cause of $C$, i.e. $X$, $Y$ and $C$ form a causal chain or a direct effect of $C$ (i.e. $X$ and $Y$ are the common effect of $C$),  (1) $I(X;C)>I(X;Y)$ and (2) $I(X;C)-I(X;Y)+I(X;Y|C)>0$.

\textit{Proof:} From Figures~\ref{fig4-1} (b) to (d),  according to the Markov condition in Definition~\ref{def2-2}, $I(X;Y|C)=0$. By $I(X;Y|C)-I(X;Y)=I(X;C|Y)-I(X;C)$, $I(X;C)>I(X;Y)$ and $I(X;C)-I(X;Y)+I(X;Y|C)>0$ hold.
\boxend
\end{property}

Since $I((Y,X);C)=I(Y;C)+I(X;C|Y)$ and $I(X;Y;C)=I(X;Y)-I(X;Y|C)=I(X;C)-I(X;C|Y)$, 
i.e. $I(X;C|Y)=I(X;C)-I(X;Y)+I(X;Y|C)$, we have $I((X,Y);C)=I(Y;C)+I(X;C)-I(X;Y)+I(X;Y|C)$, or $I((X,Y);C)-I(Y;C)=I(X;C)-I(X;Y)+I(X;Y|C)$. From  Properties~\ref{prop4-1} and~\ref{prop4-2} above, we know that if $X$ a direct cause or a direct effect of $C$, $I(X;C)-I(X;Y)+I(X;Y|C)>0$, therefore $I((X,Y);C)-I(Y;C)>0$, indicating that in the case when $X$ is a direct cause or direct effect of $C$, $X$ and $Y$ together provide more information about $C$ than $Y$ alone does.
When $X$ is a spouse of $C$ as shown in Figure~\ref{fig4-1} (e), we have the following property.

\begin{property}\label{prop4-3}
If $X$ is a spouse of $C$ through $Y$  i.e. $Y$ is a child of both $X$ and $C$, as shown in Figure~\ref{fig4-1} (e), $I(X;Y|C)>I(X;Y)$ and $I(X;C)-I(X;Y)+I(X;Y|C)>0$.

\textit{Proof:} By  Proposition~\ref{pro2-2}, $I(X;C)=0$ holds in Figure 1(e).  Since $I(X;Y|C)=I(X;Y)+I(X;C|Y)-I(X;C)$, $I(X;Y|C)> I(X;Y)$ and  $I(X;C)-I(X;Y)+I(X;Y|C)>0$ holds.
\boxend
\end{property}

Property~\ref{prop4-3} provides a causal interpretation for the class-conditional relevancy in  Eq.(\ref{eq5-09}). If $X$ is a spouse of $C$ and $Y$ is the common child of $X$ and $C$, $I(X;Y|C)-I(Y;X)>0$. Since $I((X,Y);C)=I(Y;C)+I(X;C)-I(X;Y)+I(X;Y|C)$, then even if $I(X;C)=0$, $I((X,Y);C)$ provides more information than $I(Y;C)$. This shows that although a spouse of $C$ is not a direct cause or a direct effect of $C$, from the viewpoint of class-conditional relevancy view, Property~\ref{prop4-3} confirms that the spouses of $C$ are strongly relevant features.

\begin{property}\label{prop4-4}
If $Y$ is a direct cause or a direct effect of $C$, and $X$ is an indirect cause or an indirect effect of $Y$, as shown in Figures~\ref{fig4-1} (f) to (g), then (1) $I(X;Y)>I(X;C)$, (2) $I(X;C)+I(X;Y|C)-I(X;Y)=0$, and (3) $I(Y;C)>I(X;C)$.

\textit{Proof:} By the Markov condition, in Figures~\ref{fig4-1} (f) to (g), $I(X;C|Y)=0$ holds. By $I(X;Y|C)-I(X;Y)=I(X;C|Y)-I(X;C)$, $I(X;Y)\geq I(X;C)$ and $I(X;C)+I(X;Y|C)-I(X;Y)=0$. Then by $I(Y;C|X)-I(Y;C)=I(X;C|Y)-I(X;C)$, $I(Y;C)-I(X;C)=I(Y;C|X)$. Since $I(Y;C|X)>0$, $I(Y;C)>I(X;C)$ holds.
\boxend
\end{property}

With Properties~\ref{prop4-1} to~\ref{prop4-4}, we analyze the causal interpretations of Eq.(\ref{eq5-02}), Eq.(\ref{eq5-04}), Eq.(\ref{eq5-09}), and Eq.(\ref{eq5-12}), and our observations are summarized in Table~\ref{tb-yu1}. These observations illustrate that the major non-causal feature selection methods prefer direct causes, direct effects, or spouses of $C$ to the features which are not in $MB(C)$ and further confirm that the strongly relevant features belong to $MB(C)$. Specifically, we get the following observations, and these observations will be validated by the experiments in Section~\ref{sec8-1}.
\begin{table}[t]
\small
\centering
\caption{Non-causal feature selection: objective functions  and causal interpretations}
\begin{tabular}{|l|l|l|}
\hline
Objective function                                                                                                                                                                                                                                                                             & Representative algorithm & Causal interpretation \\ \hline
\begin{tabular}[c]{@{}l@{}}Eq.(\ref{eq5-02}):\\ $X^*=\mathop{\arg\max}_{X\in F\setminus S}I(X;C)$\end{tabular}                                                                                                                                                          & MIM                      & \begin{tabular}[c]{@{}l@{}}prefer $X^*$ which is a direct\\  cause or direct effect of $C$\end{tabular}        \\ \hline
\begin{tabular}[c]{@{}l@{}}Eq.(\ref{eq5-04}): \\ $X^*=\mathop{\arg\max}_{X\in F\setminus S}\{I(X;C)$\\ $-\sum_{i=1}^{|S|}I(F_i;X)\}$\end{tabular}                                                                                                               & MIFS, mRMR, FCBF           & \begin{tabular}[c]{@{}l@{}}prefer  $X^*$  which is a direct \\cause or direct effect of $C$\end{tabular}        \\ \hline
\begin{tabular}[c]{@{}l@{}}Eq.(\ref{eq5-09}):\\ $X^*=\mathop{\arg\max}_{X\in F\setminus S}\{I(X;C)$\\ $-\Sigma_{F_i\in S}\{I(X;F_i)-I(X;F_i|C)\}$\end{tabular}                                                                                                    & JMI, CIFE, CMIM            & \begin{tabular}[c]{@{}l@{}}prefer   $X^*$ which is\\ a direct cause, direct effect,\\ or spouse of $C$\end{tabular}        \\ \hline
\begin{tabular}[c]{@{}l@{}}Eq.(\ref{eq5-12}):\\ $X^*=\mathop{\arg\max}_{X\in F\setminus S}\{I(X;C)$\\ $+\Sigma_{F_i\in S}I(X;F_i|C)$\\ $-\frac{1}{|S|}\Sigma_{F_i\in S}\{I(X;F_i)$\\ $+\Sigma_{F_j\in S, i\neq j}I(X;F_j|F_i)\}\}$\end{tabular} & RelaxMRMR                & \begin{tabular}[c]{@{}l@{}}prefer   $X^*$ which is \\ a direct cause, direct effect,\\ or spouse of $C$\end{tabular}      \\ \hline
\end{tabular}
\label{tb-yu1}
\end{table}

\begin{itemize}

\item If $S$ is empty, $\forall X\in PC(C)$, i.e. $X$ is a direct cause or effect of $C$, for any of its ancestors or descendants $F_i\in F\setminus PC(C)$, $I(X;C)>I(F_i;C)$ holds by Property~\ref{prop4-4}. Thus, Eq.(\ref{eq5-02}), Eq.(\ref{eq5-04}), Eq.(\ref{eq5-09}), and Eq.(\ref{eq5-12}) will will add $C$'s direct causes and effects first to $S$.

\item With Properties~\ref{prop4-1} to~\ref{prop4-4}, the term $I(X;C)-I(X; F_i)$ in Eq.(\ref{eq5-02}), Eq.(\ref{eq5-04}), Eq.(\ref{eq5-09}), and Eq.(\ref{eq5-12}) prefers direct causes and direct effects of $C$ (i.e. $PC(C)$), while the term $I(X;F_i|C)$ in Eq.(\ref{eq5-09}) and Eq.(\ref{eq5-12}) prefers spouses of $C$. Specifically, MIFS, mRMR, FCBF that are based on or that employ Eq.(\ref{eq5-04}) prefer the features $PC(C)$ to be added to $S$ and do not attempt to identify spouses of $C$, since Properties~\ref{prop4-1} to~\ref{prop4-2} state that only when both $X$ and $F_i$ belong to $PC(C)$, $I(X;C)>I(X;F_i)$ holds. Eq.(\ref{eq5-09}) and Eq.(\ref{eq5-12}) attempt to discover not only $PC(C)$, but also spouses of $C$, since if $X$ is a spouse of $C$, there exists a feature $F_i$, i.e. the common child of $C$ and $X$, to make $I(X;C)-I(X;F_i)+I(X;F_i|C)>0$.

\item Assuming that currently $S=\{F_i\}$. If $F_i\in ch(C)$, i.e. $F_i$ is a direct effect or a child of $C$. For two candidate features, $X\in PC(C)$ and $W$ which is a descendant of $C$  and $W\notin ch(C)$, by Property~\ref{prop4-4}, Eq.(\ref{eq5-02}), Eq.(\ref{eq5-04}), Eq.(\ref{eq5-09}), and Eq.(\ref{eq5-12}) would prefer $X$ to $W$. For example, assume that $X\rightarrow C\rightarrow F_i\rightarrow W$,  then $I(X;C)-I(X;F_i)+I(X;F_i|C)>0$ by Property~\ref{prop4-1} and $I(W;C)-I(W;F_i)+I(W;F_i|C)=0$. For MIFS and mRMR, $I(X;C)-I(X;F_i)>0$ while $I(W;C)-I(W;F_i)<0$, and for FCBF, $I(X;C)>I(X;F_i)$ while $I(F_i;C)>I(W;C)$ and $I(W;F_i)>I(W;C)$. Thus, MIFS, mRMR, FCBF prefer $X$ to $W$.
If $F_i\in pa(C)$, $X\in PC(C)$, $W$ is a ancestor of $C$ and $W\notin pa(C)$ (for example, $W\rightarrow X\rightarrow C\rightarrow F_i$), for $X$ and $W$, with a similar analysis above, Eq.(\ref{eq5-02}), Eq.(\ref{eq5-04}), Eq.(\ref{eq5-09}), and Eq.(\ref{eq5-12}) would add $X$ to $S$. 

\end{itemize}

\subsection{Causal feature selection:  assumptions and approximations}\label{sec5-2}
As discussed at the beginning of Section~\ref{sec5} and in the previous sections, non-causal feature selection methods make assumptions on the dependency among features and the class attribute under the naive Bayesian network assumptions.  Causal feature selection methods do  not have such restrictions on the structure of the (causal) Bayesian network representing the dependence relationships of all the variables, including the class attribute and all features. However, in order to learn a (causal) Bayesian network or the local network structure around the class variable, causal feature selection methods employ the Markov condition (assumption) (Definition~\ref{def2-2} in Section~\ref{sec3-1}),  faithfulness assumption (Definition~\ref{def2-4} in Section~\ref{sec3-1}) and causal sufficiency (Definition~\ref{def2-yk1} in Section~\ref{sec3-1}) for the correctness and causal meaning of the features selected.


Assuming $S$ is the feature set currently selected, $ch(C)$ is the children of $C$, $Des(C)$ is the descendants of $C$, and $ND(C)$ is the ancestors of $C$, by the Markov condition, we can get the following properties.
\begin{property}
For an unselected feature $X\in F\setminus S$, if $X\in \{ND(C)\setminus pa(C)\})$ and $pa(C)\subseteq S$, $X$ is conditionally independent of $C$ given $S$, that is, $I(X;C|S)=0$.
\label{prop5-41}
\end{property}

\begin{property}
 For an unselected feature $X\in F\setminus S$, if $X\in \{Des(C)\setminus ch(C)\})$ and $pa(X)\subseteq S$, $X$ is conditionally independent of $C$ given $S$, that is, $I(X;C|S)=0$.
\label{prop5-42}
\end{property}

With the properties, most existing causal feature selection are designed to solve  Eq.(\ref{eq5-012}) (i.e. maximizing $I(X;C|S))$ with a forward-backward strategy based on the below lemmas.
\begin{lemma}
$\forall F_i\in PC(C)$ and $\forall S\subseteq F\setminus F_i$, $I(C;F_i|S)>0$.

\textit{Proof:} By  Proposition~\ref{pro2-1}, $\forall F_i\in PC(C)$ and $\forall S\subseteq F\setminus F_i$, $F_i\nindep C|S$ holds. By Lemma~\ref{lem3-3}, the lemma holds.
\boxend
\label{lem4-1}
\end{lemma}
\begin{lemma}
If $F_i$ is a spouse of $C$ via $F_j\in ch(C)$ (i.e. $F_j$ is a common child of $F_i$ and $C$), $\exists S\subseteq F\setminus\{F_i,F_j\}$ such that  $I(C;F_i|S)=0$ and $I(C;F_i|F_j\cup S)>0$.

\textit{Proof:}  Since $C$ and $F_i$ are not directly connected by an edge, by Proposition~\ref{pro2-1}, there must exist a subset $S$ such that $C$ and $F_i$ are independent given $S$, that is, $I(C;F_i|S)=0$. By Proposition~\ref{pro2-2}, $C$ and $F_i$ are conditionally dependent given any subset containing $F_j$, i.e. the common child of $F_i$ and $C$, thus, the lemma is proven.
\boxend
\label{lem4-2}
\end{lemma}

In this section, we will analyze the search strategies taken by the existing causal feature selection methods for solving Eq.({\ref{eq5-012}). All theorems and lemmas are discussed with the assumption that all  independence tests (mutual information calculation) are reliable.


\subsubsection{A simultaneous MB discovery strategy by conditioning on the entire $S$ for calculating $I(X;C|S)$ in Eq.({\ref{eq5-012}})}

The simultaneous MB discovery strategy aims to find PC (parents and children) and spouses of $C$ simultaneously without distinguishing PC from spouses during the MB discovery. This approach adopts the forward and backward steps to greedily discover $MB(C)$ for maximizing Eq.(\ref{eq5-012}), i.e.  sequentially maximizing $I(X;C|S)$($X\in F\setminus S$) at the forward step (max-relevance) and minimizing $I(C;Y|S\setminus Y)$($Y\in S$) at the backward step (min-redundancy) by conditioning on the entire $S$ currently selected. 
This simultaneous discovery strategy has been employed by two representative algorithms, IAMB and Inter-IAMB~\citep{tsamardinos2003algorithms}.  The assumptions and search strategies of IAMB and inter-IAMB are discussed are follows.

\textbf{IAMB.} The forward and backward steps of IAMB for the sequential optimization of Eq.(\ref{eq5-012}) are as follows.  
\begin{itemize}
\item \textbf{Forward step.} At each iteration, $S$ is the set of features currently selected, and  for each candidate feature within $F\setminus S$,  the one satisfying $\arg\max_{X\in F\setminus S}I(X;C|S)$ and $I(X;C|S)>0$ is added to $S$. The forward step is terminated until $\forall X\in F\setminus S$, $I(C;X|S)=0$.

\item \textbf{Backward step.} IAMB sequentially removes from $S$  the false positive $Y\in S$ satisfying $I(C;Y|S\setminus Y)=0$ until $\forall Y\in S$, $I(C;Y|S\setminus Y)>0$.
\end{itemize}

The forward step will add  all features in the true $MB(C)$  to $S$.
Due to the greedily strategy, some false positives may enter $S$ at the forward step. For example, assuming $X\notin MB(C)$ and $\exists Y\in MB(C)$ such that $I(X;C|S\cup Y)=0$. However, when checking $I(X;C|S)$ and at this time $Y\notin S$, $I(C,X|S)>0$ holds and $X$ will be added to $S$. 
Thus,  the backward step will remove all the false positives in $S$ by Properties~\ref{prop5-41} and~\ref{prop5-42}.

\begin{theorem}\label{the4-3}
The output of IAMB is the optimal set $S^*$ in  Eq.(\ref{eq3_2}).

\textit{Proof}:
Assuming $\overline{S}$ denotes the set $F\setminus S$. At the forward step, at each iteration, $X\in F\setminus S$ is selected that satisfies Eq.(\ref{eq4_1}) below.
\begin{equation}\label{eq4_1}
  X^*=\arg\max_{X\in F\setminus S}\{I(S;C)+I(C;X|S)\}
\end{equation}
At each iteration, for all $X\in F\setminus S$, $I(S;C)$ in Eq.(\ref{eq4_1}) is the same. For the  IAMB algorithm, by Eq.(\ref{eq4_1}), at each iteration, maximizing $I(C;X|S)$ is equivalent to maximizing $I((S,X);C)$. By $I(C;F)=I(C;S)+I(C;\overline{S}|S)$, when $I(C;\overline{S}|S)=0$, then $I(C;S)$ is maximized.  At the forward step, IAMB greedily maximizes $I(C;X|S)$ until for $\forall X\in F\setminus S$, $I(C;X|S)=0$. Then by Lemma~\ref{lem4-1}, all parents and children of $C$ ($PC(C))$ will be gradually added to $S$, while by Properties~\ref{prop5-41} and~\ref{prop5-42}, the ancestors and descendants of $C$ may not be added to $S$. Let the set $SP(C)$ include all spouses of $C$, when all parents and children of $C$ are added to $S$, by Lemma~\ref{lem4-2}, $\forall X\in SP(C)$, $I(X;C|S)>0$, and thus all spouses of $C$ will be added to $S$ initially during the forward step. In any case, at the end of the forward step, all features in the true $MB(C)$ will have been added to $S$.

At the backward step,  $\exists Y^*\in S$ to be removed from $S$  satisfies
\begin{equation}\label{eq4_2}
 Y^*=\arg\min_{Y\in S}I(Y;C|S\setminus Y)
\end{equation}

By  Eq.(\ref{eq4_2}), at each iteration,  if $I(C;Y|S\setminus Y)=0$, IAMB will remove  $Y$ from $S$ until given any feature $Y\in F\setminus S$, $I(Y;C|S\setminus Y)>0$. Then all false positives in $S$ are removed, and thus $S=MB(C)$. By Theorem \ref{the3-1}, the theorem is proved.
\boxend
\end{theorem}

\textbf{Inter-IAMB.} IAMB surfers from the problem of the addition of false positives to $S$ at the forward step, then makes the size of  $S$  possibly become high-dimensional.  The Inter-IAMB strategy mitigates the problem by interleaving the forward and backward steps of IAMB to keep $S$ as small as possible, then maximizes $I(C;X|S)$ for $X\in F\setminus S$ and minimizes $I(C;Y|S\setminus Y)$ for $Y\in S$ simultaneously.

\begin{theorem}
 The output of Inter-IAMB is the optimal set $S^*$ in  Eq.(\ref{eq3_2}).

\textit{Proof}: At each iteration, by Eq.(\ref{eq4_1}), the forward step adds a new feature $X\in F\setminus S$ that maximizes $I(C;X|S)$ to $S$.  Once the new feature $X$ is added to $S$, the backward step is triggered immediately and removes  features in $S$ (false positives) that minimize Eq.(\ref{eq4_2}).  By maximizing $I(C;X|S)$ and minimizing $I(Y;C|S\setminus Y)$ simultaneously, the strategy will convergence that for $\forall X\in F\setminus S$, $I(X;C|S)=0$ and  $\forall Y\in S$, $I(Y;C|\{S\setminus Y\})>0$. After the backward step, $S=MB(C)$. Then by Theorem \ref{the3-1}, the theorem is proved.
\boxend
\label{the4-4}
\end{theorem}

The time complexity of IAMB and Inter-IAMB above is measured in the number of conditional independence tests (association computations) executed. For IAMB and Inter-IAMB, the average time complexity is  $O(n|S|)$ and the worst time complexity is $O(n^2)$ where $n$ is the total number of features and in the worst case with $|S|=n$.

Compare to non-causal feature selection, IAMB and Inter-IAMB both use the entire set of $S$ as the conditioning set for the calculation of $I(X;C|S)$ at each iteration. By Eq.(\ref{eq7-01}) in Section~\ref{sec5-05}, assuming $S_{max}$ is the largest conditioning set during MB search,  thus the minimum number of data samples $N$ required by IAMB and Inter-IAMB is $r_{X_{max}}\times r_C\times  r_{S_{max}}$. Then the number of data instances required by IAMB and Inter-IAMB will increase exponentially in the size of $S$. To mitigate this drawback, in the next section, we will discuss a divide-and-conquer strategy.

\subsubsection{A divide-and-conquer strategy by conditioning on all subsets of $S$ for calculating $I(X;C|S)$ in Eq.({\ref{eq5-012}})}

The main idea behind a divide-and-conquer strategy is that: (1) finding $PC(C)$ and $SP(C)$ separately, and (2) using a feature-subset enumeration strategy to explore subsets of $S$ for discovering $PC(C)$ instead of conditioning on the entire set of $S$. That is, to calculate $I(C;X|S)$, the divide-and-conquer strategy performs a search for a subset, $S'\subseteq S$ such that if $X$ and $C$ are conditional independent given $S'$, i.e. $I(C;X|S')=0$, $X$ will not be added to $S$ and will never be considered  as a candidate feature again.
Then, the minimum number of data samples $N$ required by the divide-and-conquer  strategy is $r_{X_{max}}\times r_C\times  r_{S'}$ where $0\leq|S'|\leq |S_{max}|$. Accordingly, on average, the divide-and-conquer strategy requires much smaller number of data samples than IAMB and Inter-IAMB. Specifically, the divide-and-conquer strategy mainly consists of  the following two steps for solving Eq.(\ref{eq5-012}).
\begin{itemize}
\item  
Discovering $PC(C)$.  At each iteration, assuming $S$ is the set of features currently selected, for each candidate feature $X\in F\setminus S$, if  $\exists S'\subseteq S$ such that $X$ and $C$ conditional independent given $S'$, i.e. $I(X;C|S')=0$, $X$ is discarded and will never be considered  as a candidate parent or child of $C$ again, otherwise $X$ is added to $S$. By Lemma~\ref{lem4-1}, after this step, all parents and children will be added to $S$.


\item Discovering $SP(C)$.  By  Lemmas~\ref{lem4-1} and~\ref{lem4-2}, $\forall X\in SP(C)$, there must exist a subset in $F\setminus \{X\}$ such that $X$ and $C$ are conditional independent given this subset. Therefore, all spouses of $C$ cannot be added to $S$ at the PC discovery step. To find $SP(C)$, by  Lemma~\ref{lem4-2}, $\forall X\in S$, the step employs the PC discovery step to find $PC(X)$, then for each feature $Y\in PC(X)$, if $\exists S'\subseteq F\setminus\{Y, X\}$ such that $I(C;Y|S')=0$ and $I(C;Y|S'\cup X)>0$, $Y\in SP(C)$. 
\end{itemize}

There are four representative approaches to instantiate the divide-and-conquer strategy, i.e. max-min heuristic, simple max-heuristic, backward heuristic, and k-greedy heuristic. The representative algorithms include MMMB~\citep{tsamardinos2003time}, HITON-MB~\citep{aliferis2003hiton}, IPC-MB~\citep{fu2008fast}, and STMB~\citep{gao2017efficient}.

\textbf{1. The max-min heuristic.}
The representative algorithm using the strategy is the MMMB algorithm, which includes the following two steps.

(1) \textbf{Discovering $PC(C)$ step.} This step includes a forward step and a backward step to find $PC(C)$. To select the feature $X^*\in F\setminus S$ to maximize $I(C;X|S)$, the forward and backward steps are implemented as follows.

\begin{itemize}

\item Forward step. The max-min heuristic selects the feature that maximizes the minimum correlation with $C$ conditioned on the subsets of $S$. Specifically, initially $S$ is an empty set,  $\forall X\in F\setminus S$, the minimum correlation, denoted as $corr(C;X|S)$, between $C$ and $X$ conditioned on all possible subsets of $S$, is calculated as Eq.(\ref{eq_23}) below.
\begin{equation}\label{eq_23}
corr(C;X|S)=\min_{S'\subseteq S}I(C;X|S')
\end{equation}
$X^*\in F\setminus S$ will be added to $S$ if $corr(C;X^*|S)>0$ and Eq.(\ref{eq_24}) below hold.
\begin{equation}\label{eq_24}
\begin{array}{lll}
 X^*&=&\arg\max_{X\in \{F\setminus S\}}corr(C;X|S)\\
\end{array}
\end{equation}
The forward step stops until $\forall X\in F\setminus S$, $corr(C;X|S)=0$.

\item Backward step. Each feature in  $S$ selected at the forward step will be checked. If $\exists Y\in S$ satisfies Eq.(\ref{eq_25}) below, it will be removed from $S$ and never considered again.
\begin{equation}\label{eq_25}
\exists S'\subseteq S\setminus Y, I(C;Y|S')=0\\
\end{equation}

\end{itemize}

(2) \textbf{Discovering $SP(C)$ step.} At the step, the max-min heuristic firstly finds the set of parents and children for each feature in $S$ found at the forward step. Assuming $X\in PC(C)$ and $Y\in PC(X)$, if $Y\notin PC(C)$ and $\exists S'\subset F\setminus \{X, Y\}$ to  make Eq.(\ref{eq_26}) below hold, then $Y$ is a spouse of $C$.
\begin{equation}\label{eq_26}
I(C;Y|S')=0\ and\ I(C; Y|S'\cup X)>0
\end{equation}

\textbf{2. Interleaving max-heuristic.} The main difference between the max-heuristic and the  interleaving max-heuristic is that in the discovering $PC(C)$ step, the interleaving max-heuristic interleaves the forward and backward steps to keep the size of $S$ as small as possible. 
The representative algorithm using the strategy is the HITON-MB algorithm.

\textbf{(1) Discovering $PC(C)$ step.} In the step, the interleaving max-heuristic uses a simpler forward strategy than the max-min heuristic. Before interleaving forward and backward steps, $\forall X\in F$,  the interleaving max-heuristic computes $I(C;X)$ and adds the features that satisfy $I(C;X)>0$ to the candidate $PC(C)$ set, called $SPC(C)$, in descending order according to the value of $I(C;X)$.  If $I(C;X)=0$, $X$ will be discarded and never considered as a candidate parent or child again. Then, initially $S$ is an empty set, and for each feature in $SPC(C)$, this strategy  interleaves  Eq.(\ref{eq_27}) and Eq.(\ref{eq_28}) as follows, until $SPC(C)$ is empty.

\begin{itemize}

\item Forward step. $\forall X\in SPC(C)$, if $X$ satisfies Eq.(\ref{eq_27}) below, it will be added to $S$.

\begin{equation}\label{eq_27}
 X^*=\arg\max_{X\in SPC(C)}I(C;X)
\end{equation}

\item Backward step. Once $X$ is added to $S$ at the forward step, the backward step is triggered. Specifically, $SPC(C)=SPC(C)\setminus X$, and $\forall Y\in S$, if $\exists S'\subseteq S\setminus Y$ satisfies Eq.(\ref{eq_28}) below, $Y$ will be removed from $S$ and never considered again.
\begin{equation}\label{eq_28}
 I(C;Y|S')=0
\end{equation}

\end{itemize}

\textbf{(2) Finding spouses.} The step is the same as the max-min heuristic in Eq.(\ref{eq_26}).

Comparing to the simultaneous discovery strategy to discover MBs in Section 5.2.1, the strategies in this section perform an subset search within $S$ instead of conditioning on the entire $S$. Thus, for the max-min heuristic and its interleaving version, the time complexity is $O(M|S|^22^{|S|})$ where $|S|$ denotes the largest size of $S$ during forward and backward steps.

\begin{theorem}
Using the max-min heuristic or its interleaving version, in the discovering $PC(C)$ step, $PC(C)\subseteq S$~\citep{tsamardinos2006max,aliferis2010local1}.



\label{the4-5}
\end{theorem}

\begin{figure}[t]
\centering
\includegraphics[height=1.2in,width=1.2in]{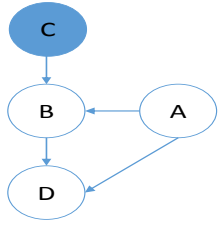}
\caption{An example of  the false positive $D$ being added to $S$ in the discovering $PC(C)$ step using the max-min heuristic or its interleaving version }
\label{fig4-new1}
\end{figure}

Theorem 5 states that in addition to $PC(C)$, the output of the discovering $PC(C)$ step, i.e. $S$, may include some false positives. For example, in Figure~\ref{fig4-new1}, assuming $C$ is the target feature, $B$, $A$, and $D$ is a child, spouse, and descendant of $C$, respectively, $D$ will enter and remain in $S$ in the discovering $PC(C)$ step~\citep{aliferis2010local1}. The explanation is  as follows. $C$ and $D$ are dependent conditioning on the empty set, since the path $C\rightarrow B\rightarrow D$ d-connects $C$ and $D$. By conditioning on $\{B\}$, the path $C\rightarrow B\leftarrow A\rightarrow D$ d-connects $C$ and $D$ by Definition~\ref{def2-5}.  

To remove false positives from $S$, such as $D$, the two max-min heuristics employ a symmetry correction. The idea behind the symmetry correction is that in a Bayesian network, if $X\in PC(C)$, then $C\in PC(X)$. With the symmetry correction, in the discovering $PC(C)$ step, the work~\citep{tsamardinos2006max,pena2007towards} proved that $S=PC(C)$. 
And with symmetry corrections, the work~\citep{aliferis2010local1} proved that the output of the two max-min heuristics is $MB(C)$, that is,  $S=MB(C)$, and thus Theorem~\ref{the4-7}  below holds.

\begin{theorem}
The output of the max-min heuristic (and its interleaving version) employed by MMMB (and HITON-MB) is the optimal set $S^*$ in  Eq.(\ref{eq3_2}) with symmetry correction.

\label{the4-7}
\end{theorem}

\textbf{3. The backward strategy.} The IPC-MB~\citep{fu2008fast} and STMB~\citep{gao2017efficient} algorithms only employ a backward step to discover $PC(C)$ instead of using a forward-backward strategy.  Initially, by setting $S=F$,  the backward step removes features from $S$ one by one, instead of greedily adding features to $S$ one by one for maximizing $I(C;S)$.
 Specifically, in the discovering $PC(C)$ step, 
for $\forall Y\in S$, if $\exists S'\subseteq S\setminus Y$ and  $|S'|=0$ (i.e., the size of $S'$ equals to 0) such that  $I(Y;C|S')=0$, $Y$ is removed from $S$.
Otherwise, if $\exists S'\subseteq S\setminus Y$ and $|S'|=1$ such that $I(Y;C|S')=0$, $Y$ is removed from $S$. The backward step continues in this way by performing level by level of the size of $S'$, until the size of the current $S'$ is larger than the size of the current $S$.

This backward strategy employed by IPC-MB also finds a superset of $PC(C)$, that is, $PC(C)\subseteq S$. Thus,  IPC-MB embeds a symmetry correction in the spouse discovery stage to remove false positives in $S$.
To find spouses, IPC-MB adopts the same idea with MMMB and HITON-MB.

 STMB also employs the backward step to discover $PC(C)$. But STMB has two main differences against IPC-MB. Firstly, STMB finds $SP(C)$ in $F\setminus S$, instead of parents and children of each feature in $S$. Secondly,  STMB uses the found spouses to remove false positives in $S$ found in the discovering $PC(C)$ step instead of using a symmetry correction during the $SP(C)$ discovery step. Specially, assuming $S$ found in the discovering $PC(C)$ step and $SP(C)=\emptyset$, the idea of discovering spouses are summarized below.
\begin{itemize}
\item Finding spouses and removing false parents and children  from $S$:  for each feature $X\in F\setminus S$, if $\exists Y\in S$ and $\exists S'\subset F\setminus\{X\cup Y\}$ s.t. $I(C;X|S')=0$ and $I(C;X|S'\cup Y)>0$, then $X$ is  added to $SP(C)$.  
Once $X$ is added to $SP(C)$, for each feature $Y\in S$, if $\exists S'\subseteq \{S\cup X\}\setminus Y$ s.t. $I(C;Y|S')=0$, then $Y$ and $X$ are removed from $S$ and $SP(C)$, respectively. The process terminates until all features in $F\setminus S$ are checked.

\item Removing false positives from $SP(C)$ and $S$: (1) $\forall X\in SP(C)$, if $I(X;C|S\cup SP(C)\setminus X)=0$, $X$ is removed from $SP(C)$; then (2)  $\forall Y\in S$, if $I(Y;C|S\cup SP(C)\setminus Y)=0$, $Y$ is removed from $S$.
\end{itemize}

IPC-MB and STMB have been proved that $\{S\cup SP(C)\}=MB(C)$~\citep{gao2017efficient,fu2008fast}. Thus,  IPC-MB and STMB greedily find the optimal set $S^*$ in  Eq.(\ref{eq3_2}).
The time complexity of IPC-MB includes finding both $PC(C)$  and  $SP(C)$, then the complexity is $O(n2^{|S|}+|S|n2^{|S|})=O(|S|n2^{|S|})$ where $|S|$ is the largest size of conditional set during search. The worst time complexity  of IPC-MB is $O(n2^n+n^22^{| S|})=O(n^22^{|S|})$ when all features are parents and children of $C$. For STMB,  and the average time complexity is $O(n2^{|S|}+|S||F\setminus S| 2^{|S|})=O(|S||F\setminus S|2^{|S|})$, and  the worst time complexity is $O(n2^n+n^22^{|S|})=O(n^22^{|S|})$.

\textbf{4. $\gamma$-greedy heuristic.} In the discovering $PC(C)$ step, as the size of $S$ becomes large,  it will be computationally expensive or prohibitive when we perform an exhaustive enumeration over all subsets of $S$. For example, to check whether $X$ is able to be added to $S$, in the worst case, the total number of subsets checked is up to $2^{|S|}$.
Accordingly, in the discovering $PC(C)$ step, MMMB, HITON-MB, IPC-MB and STMB employ  a $\gamma$-greedy search method to mitigate this problem. The $\gamma$-greedy search checks all subsets of size less than or equal to a user-defined parameter $\gamma\ (0\leq \gamma<|S|)$,  that is, the maximum size of subsets needed to be checked. In the case of using the $\gamma$-greedy heuristic, MMMB, HITON-MB, IPC-MB and STMB  return an approximate $MB(C)$~\citep{aliferis2010local1}.

\subsection{Practical implication}

In Section~\ref{sec5-1} and Section~\ref{sec5-2},  we discussed the Bayesian network structural assumptions and analyzed in detail how the assumptions led to the different levels of approximations employed by causal and non-causal feature selection methods for the calculation of $I(X;C|S)$. With the structural assumptions, we are able to fill in the gap in our understanding of the relation between the two types of  feature selection methods.

Firstly, the feature sets obtained by causal feature selection methods are closer to $MB(C)$ than non-causal feature selection methods. However, our analysis in Sections~\ref{sec5-1} and~\ref{sec5-2} shows that non-causal feature selection methods are much more computationally efficient and have lower sample requirement than causal feature selection methods. The choice of causal or non-causal feature selection methods depends on the size of the dataset under study.  

Secondly, the strongly relevant features are the same as the MB of $C$. This may motivate us to leverage the advantages of both causal and non-causal feature selection methods to develop  more efficient and robust new feature selection methods.

Thirdly, causal and non-causal feature selection methods implicitly reduce a full Bayesian network classifier to a selective Bayesian network classifier by selecting a subset of features $S$ to make the conditional likelihood $P(C|S)$ as close to $P(C|F)$ as possible, as shown in Figure~\ref{fig6-4}.

\begin{figure}[t]
\centering
\includegraphics[height=2in,width=4.7in]{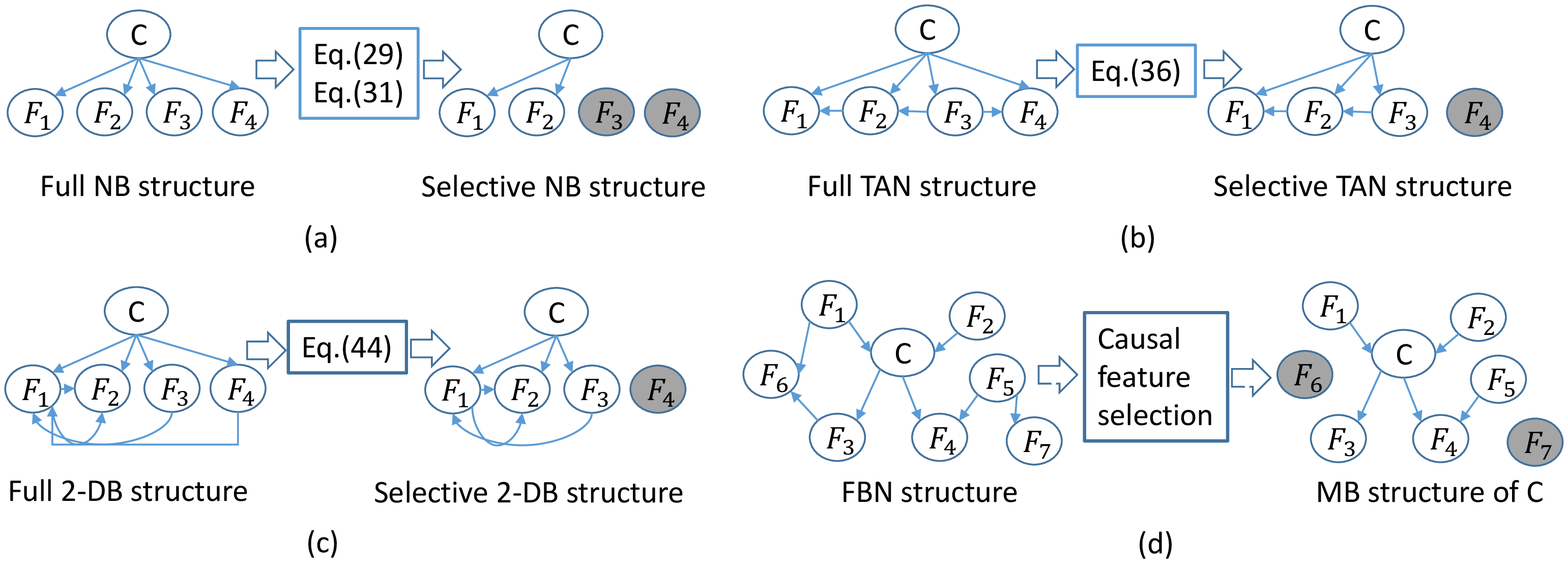}
\caption{(a) A NB to a selective NB  (b) a TAN to a selective TAN  (c) a $2$-DB to selective $2$-DB (d) a general Bayesian network to a MB-based Bayesian  network classifier}
\label{fig6-4}
\end{figure}

}

\section{Error Bounds}\label{sec6}

In the section, we will discuss the error bounds of non-causal and causal feature selection for understanding the impact of assumptions and approximations made by the two types of methods on classification performance.
Since both types of methods are independent of any classifiers, we will analyze the bounds of difference in the information gains between an approximate MB and an exact MB.
In Section 3, Eq.(\ref{eq3_7}) has presented that if a subset $S$ in $D$ maximizing $I(C;S)$, then $S$ also maximizes $L(C|S,D)$ and minimizes $P_{err}$.
Based on Eq.(\ref{eq3_7}), using information gain, in the following, we will discuss the bounds of the difference between an approximate MB and an exact MB.

\subsubsection{Conditioning on the full $S$ and its all subsets (exact MB discovery)}

According to our analysis in Section~\ref{sec5-2}, under certain assumptions, causal feature selection algorithms designed with conditioning on both the full $S$ and all of its subsets can find the exact $MB(C)$ from data. Moreover, the algorithms by conditioning on all subsets of $S$ are also able to find the exact $PC(C)$.  
As $I(C;F\setminus MB(C)|MB(C))=0$,  $I(C;F)=I(C;MB(C))$. Since $H(P_{err})^{-1}\leq P_{ber}\leq 1/2H(C|F)$ (see Eq.(\ref{eq3_4})), Theorem~\ref{the7-1} gives the minimum upper bound of $P_{err}$.
\begin{theorem}
$P_{err}\leq 1/2H(C|MB(C))$.

\textbf{Proof:} By Theorem~\ref{the3-1}, $\forall S\subseteq F$, $I(C;MB(C))\geq I(C;S)$ holds. Since $H(C|MB(C))=H(C)-I(C;MB(C))$ and $H(C|S)=H(C)-I(C;S)$, we get that $\forall S\subseteq F$, $H(C|MB(C))\leq H(C|S)$. By Eq.(\ref{eq3_5}), the theorem is proven.
\boxend
\label{the7-1}
\end{theorem}

By Eq.(\ref{eq3_6}), $\lim_{m\to\infty}{-\ell(C|S,D)}=KL(p(C|S)||q(C|S))+H(C|S)$. As $m\to\infty$, $KL(p(C|S)||q(C|S))$ will approach zero, and thus Eq.(\ref{eq7-11}) presents that $H(C|MB(C))$ minimizes $-\ell(C|MB(C),D)$.
\begin{equation}
\lim_{m\to\infty}{-\ell(C|MB(C),D)}\approx H(C|MB(C))
\label{eq7-11}
\end{equation}

If $S=PC(C)$ holds, by Theorem~\ref{the3-1}, $I(MB(C);C)\geq I(PC(C);C)$ holds. Thus, $H(C|MB(C))\leq H(C|PC(C))$ holds, and  Eq.(\ref{eq7-12}) below gives the Bayes error rates of $PC(C)$, that is, $P_{err}(PC(C))$. Since $H(C|MB(C))\leq H(C|PC(C))$, the upper bound in Eq.(\ref{eq7-12}) is  looser than that in Eq.(\ref{eq7-11}).
\begin{equation}
P_{err}(PC(C))\leq 1/2H(C|PC(C))
\label{eq7-12}
\end{equation}

Since $\lim_{m\to\infty}{-\ell(C|PC(C),D)}\approx H(C|PC(C))$, Eq.(\ref{eq7-13}) below gives the upper bound of the conditional log-likelihood of $PC(C)$  in $D$, that is, $-H(C|MB(C))$.
\begin{equation}
 \lim_{m\to\infty}{\ell(C|PC(C),D)}\leq -H(C|MB(C))
\label{eq7-13}
\end{equation}


\subsubsection{Conditioning on the subsets of $S$ up to size $\gamma$ (causal feature selection).} 

As we discussed in Section~\ref{sec5-2}, the $\gamma$-greedy search employed by causal feature selection methods may return an approximate $MB(C)$.
Let $AMB(C)\subseteq F$ be an approximate MB of $C$, by Theorem~\ref{the3-1}, $H(C|MB(C))\leq H(C|AMB(C))$ holds.
Since Theorem~\ref{the7-1} illustrates that $1/2H(C|MB(C))$ is the minimum upper bound of $P_{err}$, thus, we get
\begin{equation}
P_{err}(AMB(C))\leq 1/2H(C|MB(C)).
\label{eq7-14}
\end{equation}

Since $\lim_{m\to\infty}{-\ell(C|AMB(C),D)}\approx H(C|AMB(C))$ holds, the upper bound of the conditional log-likelihood of any approximate MB of $C$ is 
\begin{equation}
 \lim_{m\to\infty}{\ell(C|AMB(C),D)}\leq -H(C|MB(C)).
\label{eq7-15}
\end{equation}

\subsubsection{Conditioning on the subset of size 0 or 1 (non-causal feature selection).}
As discussed in Section~\ref{sec5-1},  non-causal feature selection algorithms attempt to find $PC(C)$ and some spouses of $C$. With different values of $\psi$ (i.e. the number of selected features), those strategies may return an approximate $MB(C)$, that is, a superset or a subset of $PC(C)$.  In the following, we will focus on  discussing the bounds of the superset or subset of $PC(C)$ found by non-causal feature selection methods.

\begin{corollary}
If $S1\subseteq F\setminus PC(C)$ and $S=PC(C)\cup S1$,

(1) $-H(C|PC(C))\leq\lim_{m\to\infty}{\ell(C|S,D)}\leq -H(C|MB(C))$;

(2) $1/2H(C|MB(C))\leq P_{err}(S)\leq 1/2H(C|PC(C))$.

\textbf{Proof:} Assuming $\overline{PC(C)}=F\setminus PC(C)$ and $\overline{S}=F\setminus S$. Firstly, we prove that $I(C;\overline{PC(C)}|PC(C))\geq I(C;\overline{S}|S)$ holds. By $I(C;F)=I(PC(C);C)+I(C;\overline{PC(C)}|PC(C))$, we get 
\begin{equation}
\begin{array}{rcl}
I(C;F)&=&I((\overline{S},S);C)\\
&=&I(S;C)+I(C;\overline{S}|S)\\
&=&I((PC(C),S1);C)+ I(C;\overline{S}|S)\\
&=&I(PC(C);C)+I(S1;C|PC(C))+ I(C;\overline{S}|S)
\end{array}
\label{eq7-16}
\end{equation}
By the chain rule of mutual information, we can get
\begin{equation}
I(S1;C|PC(C)) =\sum_{j=1}^{|S1|}I(F_j;C|F_{j-1},\cdots, F_1,PC(C)).
\label{eq7-17}
\end{equation}

Since $S1$ only includes spouses, non-descendants and descendants of $C$. By Eq.(\ref{eq7-16}) and Eq.(\ref{eq7-17}), we get the following.

Case 1: if $\exists F_j\in S1$ is a descendant of $C$ and $I(F_j;C|F_{j-1},\cdots, F_1,PC(C))>0$, then  $I(C;\overline{PC(C)}|PC(C))>I(C;\overline{S}|S)$ holds.

Case 2: if $\exists F_{j}\in S1$ and  $F_{j}$ is a spouse of $C$, then $I(F_j;C|F_{j-1},\cdots, F_1,PC(C))>0$. Thus, $I(C;\overline{PC(C)}|PC(C))>I(C;\overline{S}|S)$ holds.

Case 3: if $F_j\in S1$ is a non-descendant of $C$, by the Markov condition,

$I(F_j;C|F_{j-1},\cdots, F_1,PC(C))=0$, then $I(C;\overline{PC(C)}|PC(C))=I(C;\overline{S}|S)$.

By $I(C;S)\leq I(C;MB(C))$, $I(C;\overline{S}|S)\geq I(C;\overline{MB(C)}|MB(C))$ holds.
Then we get
$$I(C;\overline{PC}|PC(C))\geq I(C;\overline{S}|S)\geq I(C;\overline{MB(C)}|MB(C)).$$
Then $I(C;PC(C))\leq I(C;S)\leq I(C;MB(C))$ holds. Thus, we get $H(C|PC(C))\geq H(C|S)\geq H(C|MB(C))$. Thus, (1) and (2) hold.
\boxend
\label{cor7-2}
\end{corollary}

For a subset of $PC(C)$, assuming $S\subset PC(C)$, $\overline{S}=F\setminus S$. If $PC(C)=\{S\cup S'\}$, $I(C;PC(C))=I(C;S)+I(C,S'|S)$. Since
$I(C;S'|S)=\sum_{i=1}^{|S'|}I(F_i;C|F_{i-1},\cdots, F_1,S)$ holds and $S'\subset PC(C))$, then $I(C;S'|S)>0$. Thus, $I(C;PC(C))>I(C;S)$ holds. By $I(C;F)=I(C;PC(C))+I(C;\overline{PC}|PC(C))$ and $I(C;F)=I(C;S)+I(C;\overline{S}|S)$, then $I(C;\overline{PC}|PC(C))<I(C;\overline{S}|S)$.
Accordingly, we can get the bounds between $S\subset PC(C)$ and $PC(C)$ in the following:
\begin{equation}
\lim_{m\to\infty}{\ell(C|S, D)}<-H(C|PC(C))\ and\ P_{err}(S)<1/2H(C|PC(C))
\label{eq7-9}
\end{equation}

By the analysis above, we can see that the errors  of causal and non-causal feature selection methods are bounded by $1/2H(C|MB(C))$ and $1/2H(C|PC(C))$, respectively. This indicates that the error bound of non-causal feature selection is looser than that of causal feature selection. Therefore, referring back to Figure~\ref{fig4-0}, our analysis in this section validates that as causal feature selection methods make no assumption on the structure of the Bayesian network representing dependency of variables, their search strategies are able to find the exact $MB(C)$, while the strong assumptions made by non-causal feature selection methods lead to an approximate $MB(C)$ (referring back to Figure~\ref{fig4-001}).

\section{Experiments}\label{sec8}

In this section, we will conduct extensive experiments to  validate 
our findings of causal and non-causal feature selection, with the following focuses:

\begin{itemize}
\item In Section~\ref{sec8-1}, we validate Theorem~\ref{the3-1} in Section~\ref{sec4-2} (i.e. the MB of $C$ is the optimal set for feature selection), the discussion in Section~\ref{sec5-04} (causal interpretations of non-causal feature selection), and the proposed error bounds in Section~\ref{sec6} using a set of synthetic data sampled from a benchmark Bayesian network.

\item In Section~\ref{sec8-2}, as seen in the experiment results, we investigate the  impact of different levels of approximations made by causal and non-causal feature selection methods on classification performance, the computational and accuracy performance of causal and non-causal feature selection methods, and the impact of data sample sizes on both methods using 25 various types of real-world datasets, including six datasets with large data samples, six datasets with extreme small samples, seven datasets with multiple classes, and six class-imbalanced datasets.
\end{itemize}

To carry out these validations, we have selected the following eight representative feature selection methods:
\begin{itemize}
\item Five representative causal feature selection methods, including three MB discovery algorithms, IAMB, HITON-MB, MMMB and two PC discovery algorithms, HITON-PC and MMPC and we use the implementations of these algorithms obtained from http://www.dsl-lab.org/causal\_explorer;

\item Three representative non-causal feature selection algorithms: mRMR, JMI, and CMIM since these three algorithms provides better
tradeoff in terms of accuracy and scalability than the other non-causal feature selection algorithms (especially with small-sized data samples)~\citep{brown2012conditional}. We use the implementations of mRMR, JMI, and CMIM obtained from https://github.com/Craigacp/FEAST.
\end{itemize}

To evaluate the selected features by each algorithm for classification,  in all experiments, we use Naive Bayes classifier (NBC) and k-Nearest Neighbor (KNN) classifier since SVMs, Random Forests, and Decision trees implicitly embed a feature selection process into themselves while NB and KNN do not. 
All experiments were performed on a Window 7 Dell workstation with an Intel(R) Core(TM) i5-4570, 3.20GHz processor and 8.0GB RAM, and all eight feature selection methods under comparison are implemented in MATLAB, and  NBC and KNN are implemented in MATLAB2014 Statistics Toolbox. In the tables in Section~\ref{sec8}, the notation ``$A\pm B$" denotes that ``A" is the  average performance of an algorithm on a dataset, such as prediction accuracy, while ``B" represents the corresponding standard deviations of the  average performance.

\subsection{Experiments using synthetic datasets}\label{sec8-1}

In this section, we will validate $MB(C)$ is the optimal set for feature selection (Theorem~\ref{the3-1} in Section~\ref{sec4-2}) and the proposed error bounds in Section~\ref{sec6} using a set of synthetic data sampled from the ALARM (A Logical Alarm Reduction Mechanism) network, a benchmark and well-known Bayesian network modelling an alarm message system for patient monitoring~\citep{beinlich1989alarm}.
This network includes 37 variables and the complete structure of the network is shown in Figure~\ref{fig6_1}.  Since the MB of each variable can be read from the network, we are able to evaluate the performance of the feature selection methods against the true MBs. 
\begin{figure}
\centering
\includegraphics[height=3in,width=4in]{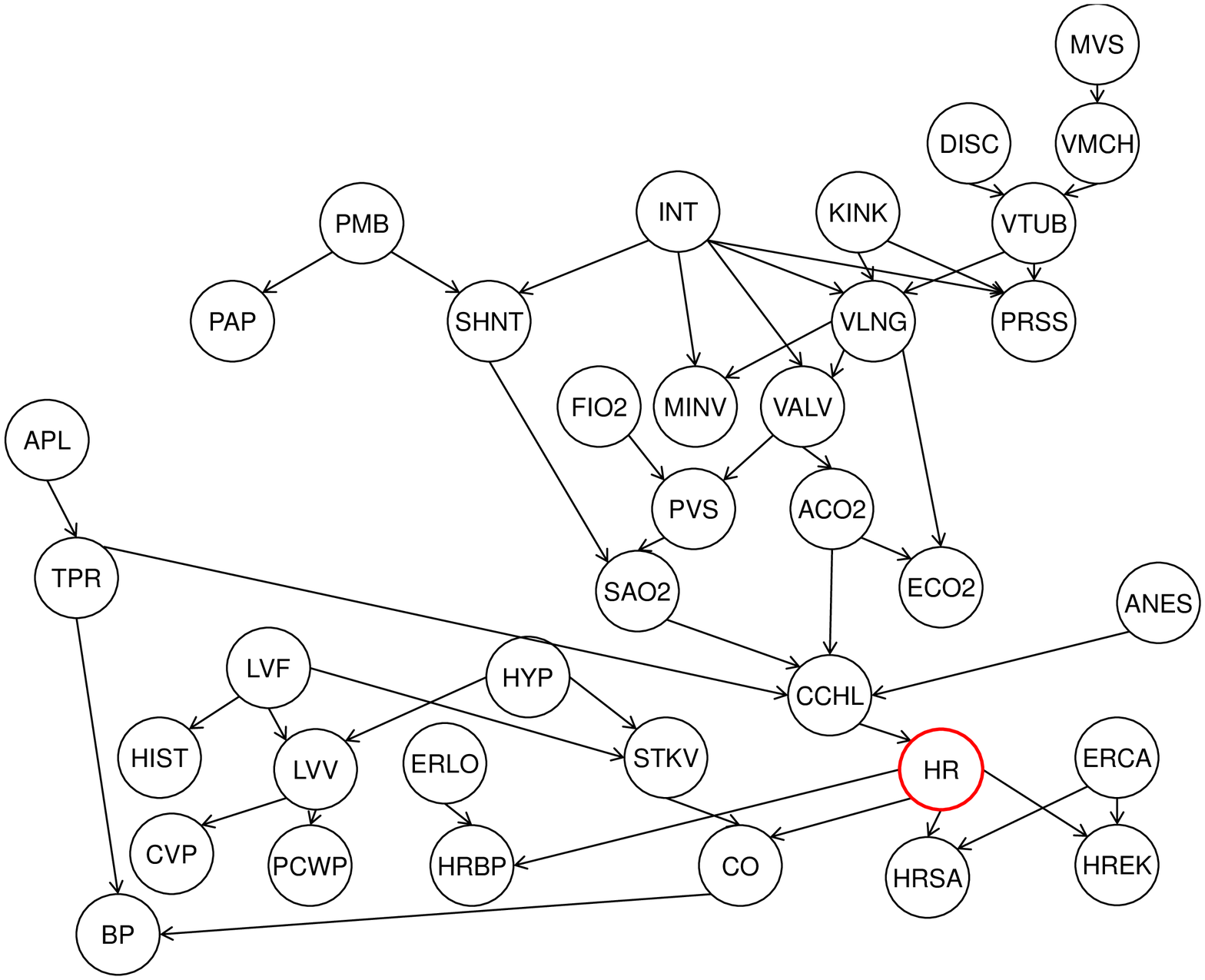}
\caption{The ALARM Bayesian network}
\label{fig6_1}
\end{figure}

In the ALARM network, we choose the ``HR'' (Heart Rate) variable as the class attribute for classification. The variable takes three class labels, ``low'', ``normal'', and ``high'', and has the largest MB among all variables, including one parent, four children, and three spouses.
We randomly sampled 10 training datasets with 5,000 training cases (large-sized data samples) and 50 training cases (small-sized data samples) respectively. For each training dataset, we randomly sampled a testing dataset with 1,000 testing cases. The reported prediction accuracy is the average accuracy of a classifier using the feature sets selected over the 10 runs of a feature selection method on these 10 training datasets. 

In all tables in Section~\ref{sec8-1}, ``TruePC" and ``TrueMB" denote the ground-truths of PC and MB of ``HR'' in the network, respectively. 
For validating the discussion of causal interpretations of non-causal feature selection methods in Section~\ref{sec5-04}, we use the following settings and evaluation metrics.
\begin{itemize}
\item We set the parameter $\psi$, i.e. the numbers of features selected by mRMR, JMI, and CMIM to the size of the true MB (or the true PC) of ``HR'' in the network, which is denoted as ``$\psi$=nMB" (or ``$\psi$=nPC"). 

\item We use the average precision and recall metrics using the feature sets selected over the 10 runs of a feature selection method on the 10 training datasets to observe the percentage of the MB (or the direct causes and direct effects) of  ``HR'' included in the selected features (the output) of each algorithm.
The precision metric is the number of true positives in the output (i.e. the variables in the output belonging to the true MB (or PC) of  ``HR'' in the ALARM network) divided by the number of variables in the output of an algorithm.
The recall metric is the number of true positives in the output divided by the number of true positives (the number of the true MB (or PC) of ``HR'' in the alarm network). 

\item We use the prediction accuracy, the ratio between the number of correct predictions and the total number of testing data samples to validate Theorem~\ref{the3-1} presented in Section~\ref{sec4-2} and the error bounds proposed in Section~\ref{sec6}.
\end{itemize}

\begin{table}[t]
\centering
\scriptsize
\caption{Prediction accuracy of true MB against causal and non-causal algorithms}
\begin{tabular}{|l|l|l|l|l|}
\hline
\multirow{2}{*}{Algorithm} & \multicolumn{2}{c|}{5000 cases}                             & \multicolumn{2}{c|}{50 cases}                               \\ \cline{2-5}
                           & \multicolumn{1}{c|}{KNN}  & \multicolumn{1}{c|}{NBC}   & \multicolumn{1}{c|}{KNN}  & \multicolumn{1}{c|}{NBC}   \\ \hline
TrueMB                     & \textbf{98.99$\pm$0.3542} & \textbf{98.52$\pm$0.3765} & \textbf{98.99$\pm$0.3542} & \textbf{97.27$\pm$0.3889} \\ \hline
IAMB                       & 98.65$\pm$0.3719          & 98.34$\pm$0.5420          & 94.67$\pm$3.8500          & 95.06$\pm$3.8200          \\ \hline
MMMB                       & \textbf{98.99$\pm$0.3542} & \textbf{98.52$\pm$0.3765} & 95.02$\pm$1.3456          & 95.13$\pm$1.3516          \\ \hline
HITON-MB                   & \textbf{98.99$\pm$0.3542} & \textbf{98.52$\pm$0.3765} & 95.02$\pm$1.3456          & 95.13$\pm$1.3516          \\ \hline
mRMR ($\psi$= nMB)              & 98.38$\pm$0.2898          & 98.42$\pm$0.3360 & 95.40$\pm$1.7321          & 95.94$\pm$1.4152          \\ \hline
CMIM ($\psi$= nMB)              & 98.17$\pm$0.3653          & 98.38$\pm$0.4367          & 93.25$\pm$2.5238          & 93.49$\pm$1.8064          \\ \hline
JMI ($\psi$= nMB)               & 98.67$\pm$0.4523          & 98.33$\pm$0.3889          & 94.83$\pm$1.3259          & 95.50$\pm$1.3968          \\ \hline
\end{tabular}
\label{tb6_11}
\end{table}
\begin{table}[t]
\centering
\caption{Precision and Recall of each algorithm for MB discovery}
\scriptsize
\begin{tabular}{|l|l|l|l|l|}
\hline
\multirow{2}{*}{Algorithm} & \multicolumn{2}{c|}{5000 cases} & \multicolumn{2}{c|}{50 cases} \\ \cline{2-5}
   & \multicolumn{1}{c|}{precision}      & \multicolumn{1}{c|}{recall}     & \multicolumn{1}{c|}{precision}      & \multicolumn{1}{c|}{recall}   \\ \hline
IAMB	 &1$\pm$0	  &0.6875$\pm$0.0659	  &1$\pm$0	&0.1250$\pm$0\\ \hline
MMMB	&1$\pm$0	 &1$\pm$0	 &0.6885$\pm$0.1282	 &0.9000$\pm$0.0791\\ \hline
HITON-MB	&1$\pm$0	 &1$\pm$0	&0.6500$\pm$0.0791	&0.6500$\pm$0.0791\\ \hline
MRMR ($\psi$= nMB)	&0.6250$\pm$0	 &0.6250$\pm$0	  &0.6500$\pm$0.0527	&0.6500$\pm$0.0527\\ \hline
CMIM ($\psi$= nMB)	&0.6250$\pm$0	&0.6250$\pm$0	&0.4875$\pm$0.1905	&0.4875$\pm$0.1905\\ \hline
JMI ($\psi$= nMB)	&0.7500$\pm$0	 &0.7500$\pm$0	 &0.6500$\pm$0.0791	&0.6500$\pm$0.0791\\ \hline
\end{tabular}
\label{tb6_12}
\end{table}
\begin{table}[t]
\centering
\scriptsize
\caption{Number of parents and children (PC), spouses (SP), and false positives (FP)}
\begin{tabular}{|l|l|l|l|l|l|l|}
\hline
\multirow{2}{*}{Algorithm} & \multicolumn{3}{c|}{5000 cases}                                                 & \multicolumn{3}{c|}{50 cases}                                                   \\ \cline{2-7}
                           & \multicolumn{1}{c|}{PC} & \multicolumn{1}{c|}{SP} & \multicolumn{1}{c|}{FP} & \multicolumn{1}{c|}{PC} & \multicolumn{1}{c|}{SP} & \multicolumn{1}{c|}{FP} \\ \hline
IAMB                       & 4.3$\pm$0.4830          & 1.2$\pm$0.4216              & 0                       & 1$\pm$0                 & 0                           & 0                       \\ \hline
MMMB                       & 5$\pm$0                 & 3$\pm$0                     & 0                       & 4.9$\pm$0.3162          & 2.3$\pm$0.6749              & 3.5$\pm$1.7159          \\ \hline
HITON-MB                   & 5$\pm$0                 & 3$\pm$0                     & 0                       & 4.9$\pm$0.3162          & 0.3$\pm$0.4830              & 2.8$\pm$0.6325          \\ \hline
mRMR ($\psi$= nMB)              & 5$\pm$0                 & 0                           & 3$\pm$0                 & 4.9$\pm$0.3162          & 0.3$\pm$0.4830              & 2.8$\pm$0.4216          \\ \hline
CMIM ($\psi$= nMB)              & 5$\pm$0                 & 0                           & 3$\pm$0                 & 3.9$\pm$1.5239          & 0                           & 3.1$\pm$1.1972          \\ \hline
JMI ($\psi$= nMB)               & 5$\pm$0                 & 1$\pm$0                     & 2$\pm$0                 & 4.9$\pm$0.3162          & 0.3$\pm$0.4830              & 2.8$\pm$0.6325          \\ \hline
\end{tabular}
\label{tb6_13}
\end{table}
\begin{table}[t]
\centering
\caption{Prediction accuracy of true PC against causal and non-causal algorithms}
\scriptsize
\begin{tabular}{|l|l|l|l|l|}
\hline
\multirow{2}{*}{Algorithm} & \multicolumn{2}{c|}{5000 cases}                       & \multicolumn{2}{c|}{50 cases}                         \\ \cline{2-5}
                & \multicolumn{1}{c|}{KNN}  & \multicolumn{1}{c|}{NBC}   & \multicolumn{1}{c|}{KNN}  & \multicolumn{1}{c|}{NBC}   \\ \hline
TruePC                     & \textbf{98.44$\pm$0.3596} & \textbf{98.57$\pm$0.3199} & \textbf{98.44$\pm$0.3596} & \textbf{97.36$\pm$0.3718} \\ \hline
MMPC                       & \textbf{98.44$\pm$0.3596} & \textbf{98.57$\pm$0.3199} & 96.66$\pm$0.7382          & 97.01$\pm$1.2360          \\ \hline
HITON-PC                   & \textbf{98.44$\pm$0.3596} & \textbf{98.57$\pm$0.3199} & 96.66$\pm$0.7382          & 97.01$\pm$1.2360          \\ \hline
mRMR ($\psi$= nPC)              & \textbf{98.44$\pm$0.3596} & \textbf{98.57$\pm$0.3199} & 96.28$\pm$1.2017          & 96.44$\pm$1.0710          \\ \hline
CMIM ($\psi$= nPC)              & \textbf{98.44$\pm$0.3596} & \textbf{98.57$\pm$0.3199} & 95.77$\pm$1.7601          & 96.31$\pm$0.8850          \\ \hline
JMI ($\psi$= nPC)               & \textbf{98.44$\pm$0.3596} & \textbf{98.57$\pm$0.3199} & 94.08$\pm$3.1435          & 94.59$\pm$1.5975          \\ \hline
\end{tabular}
\label{tb6_14}
\end{table}
\begin{table}[t]
\centering
\caption{Precision and Recall of each algorithm for PC discovery}
\scriptsize
\begin{tabular}{|l|l|l|l|l|}
\hline
\multirow{2}{*}{Algorithm} & \multicolumn{2}{c|}{5000 cases} & \multicolumn{2}{c|}{50 cases} \\ \cline{2-5}
          & \multicolumn{1}{c|}{precision}      & \multicolumn{1}{c|}{recall}     & \multicolumn{1}{c|}{precision}      & \multicolumn{1}{c|}{recall}      \\ \hline
MMPC	    &1$\pm$0	  &1$\pm$0	 &0.9714$\pm$0.0904	 &0.9200$\pm$0.1033\\ \hline
HITON-PC	&1$\pm$0	  &1$\pm$0	 &0.9714$\pm$0.0904	 &0.9200$\pm$ 0.1033\\ \hline
MRMR ($\psi$= nPC)	&1$\pm$0	&1$\pm$0	&0.8600$\pm$0.0966	&0.8600$\pm$0.0966\\ \hline
CMIM ($\psi$= nPC)	&1$\pm$0	&1$\pm$0	&0.5800$\pm$0.1989	&0.5800$\pm$0.1989 \\\hline
JMI ($\psi$= nPC)	&1$\pm$0	&1$\pm$0	&0.8000$\pm$0.0943	&0.8000$\pm$0.0943  \\ \hline
\end{tabular}
\label{tb6_15}
\end{table}

\subsubsection{Validation of Theorem~\ref{the3-1} and the discussion in Section~\ref{sec5-04}}

In this section, we will validate Theorem~\ref{the3-1} (i.e. the MB of $C$ is the optimal set for feature selection), and the discussion of causal interpretations of non-causal feature selection in Section~\ref{sec5-04}. 

\textbf{Validating Theorem~\ref{the3-1}}.
Table~\ref{tb6_11} reports the average prediction accuracies and standard deviations using the datasets containing 5,000 and 50 training cases, respectively.
Table~\ref{tb6_11} states that using both KNN and NBC, the true MB of ``HR" achieves the highest prediction accuracy  than the feature subsets selected by IAMB, HITON-MB, MMMB, mRMR, CMIM, and JMI. Table~\ref{tb6_13} shows the number of parents and children (PC), spouses (SP), and false positives (FP) in the found feature set of each algorithm.
These results indicate that classifiers using $MB(C)$ as the feature set achieve the best classifications results, which validates Theorem~\ref{the3-1} in Section~\ref{sec4-2}.

From Tables~\ref{tb6_12} and~\ref{tb6_13}, we can see that using 5000 cases, both MMMB and HITON-MB find the exact MB of ``HR", and thus get the same prediction accuracy as the true MB of ``HR", while the other four algorithms do not find the exact MB of ``HR".  In addition, from Table~\ref{tb6_13}, we can see that except for IAMB, all feature sets found by the other five algorithms include all variables within the PC set of ``HR". This explains why the IAMB, mRMR, JMI, and CMIM are very competitive on the prediction accuracy. CMIM and mRMR cannot find any spouses using 5000 cases. 

Using 50 cases, in Table~\ref{tb6_11},  mRMR gets the  highest  prediction accuracy among IAMB, MMMB, HITON-MB, CMIM, and JMI using both KNN and NBC, since it finds almost the same PC set  as  the other rivals, but  achieves fewest false positives among all algorithms, as shown in  Table~\ref{tb6_13}.  This shows that non-causal feature selection algorithms deal with small-sized data samples better than causal feature selection algorithms, which is consistent with our discussions of sample requirement in Section~\ref{sec5}. 

\textbf{Validating the discussion presented in Section~\ref{sec5-04}}. Table~\ref{tb6_14} reports the prediction accuracies using MMPC, HITON-PC, mRMR, JMI, and CMIM, while Table~\ref{tb6_15} illustrates the precision and recall of each algorithm for PC discovery. From Tables~\ref{tb6_13} to~\ref{tb6_15}, we can see that the PC set of a target feature plays a key role in predicting the target. From Tables~\ref{tb6_14} to~\ref{tb6_15}, we can see that using 5000 cases (large-sized data samples), the three non-causal feature selection methods find the exact PC set of ``HR", and thus they get the same prediction accuracy as the true PC set. Even using 50 cases (a small-sized data samples), Table~\ref{tb6_15} states that the three non-causal feature selection methods still prefer the features in the PC set of ``HR". Therefore, Tables~\ref{tb6_13} to~\ref{tb6_15} provide strong evidence to support the discussion of causal interpretations of non-causal feature selection in Section~\ref{sec5-04}.
\begin{figure}
\centering
\includegraphics[height=1.45in]{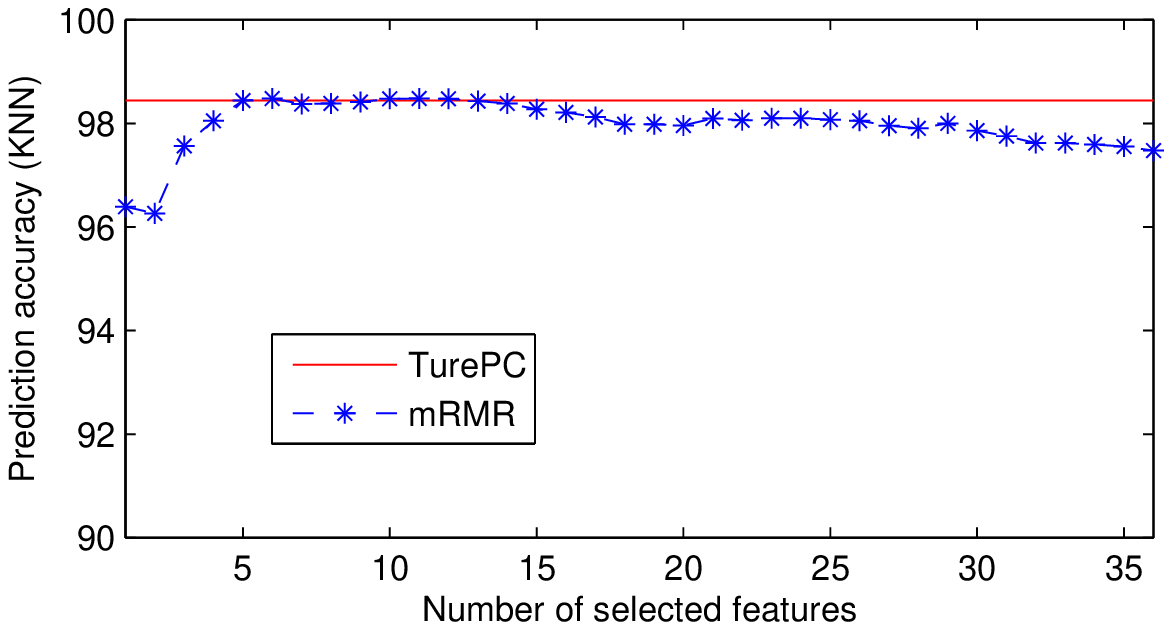}
\includegraphics[height=1.45in]{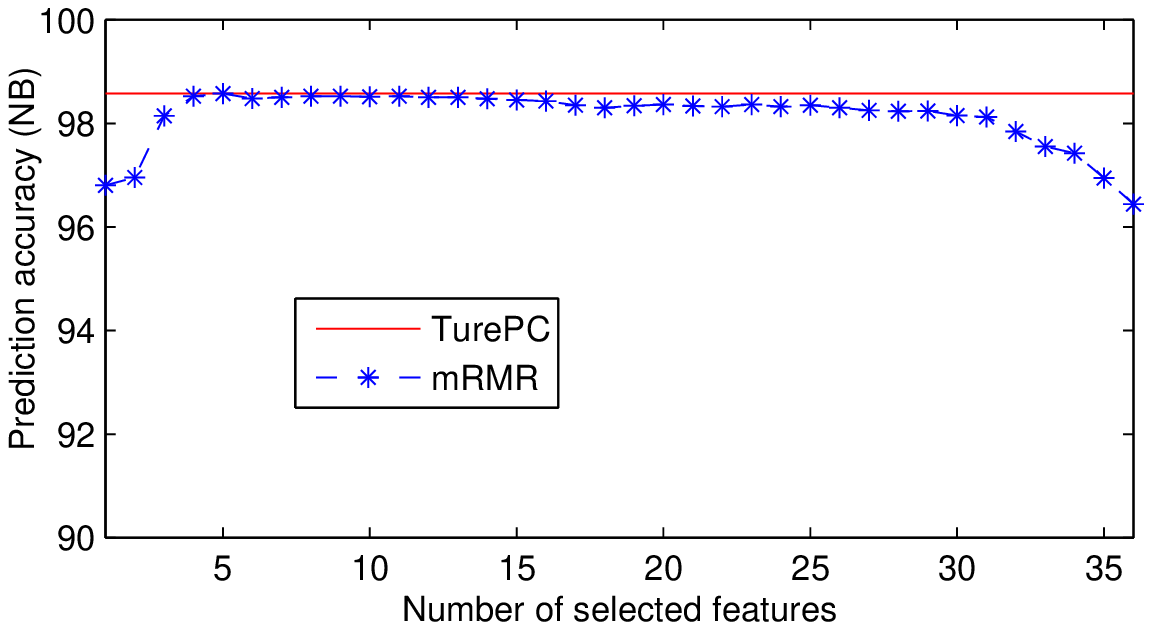}
\caption{mRMR and TurePC}
\label{fig6_11}
\end{figure}
\begin{figure}
\centering
\includegraphics[height=1.45in]{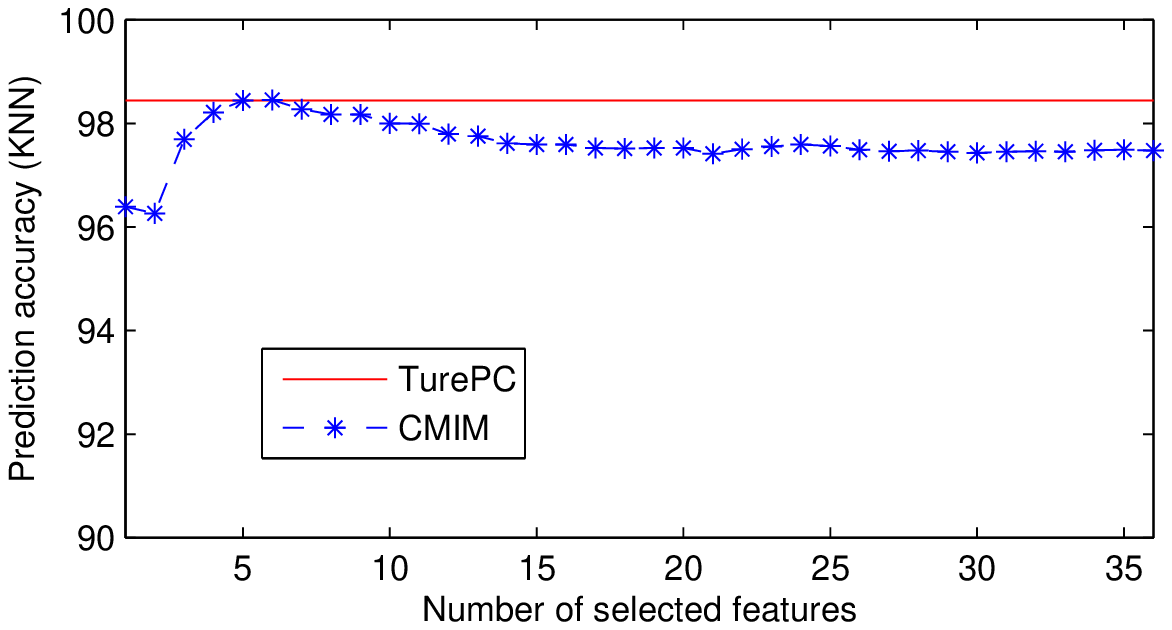}
\includegraphics[height=1.45in]{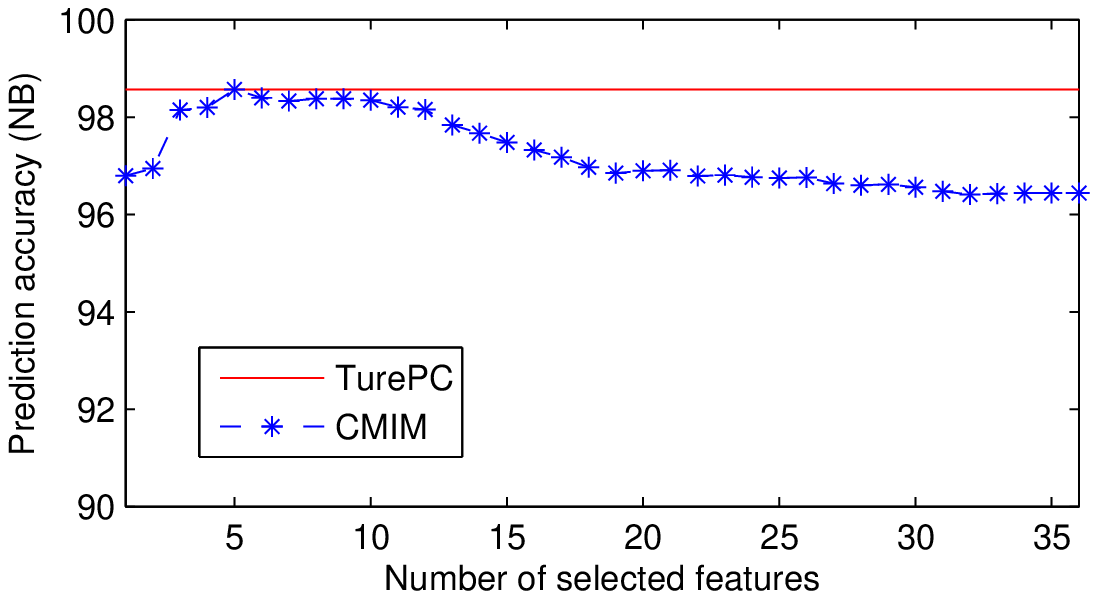}
\caption{CMIM and TurePC}
\label{fig6_12}
\end{figure}
\begin{figure}
\centering
\includegraphics[height=1.45in]{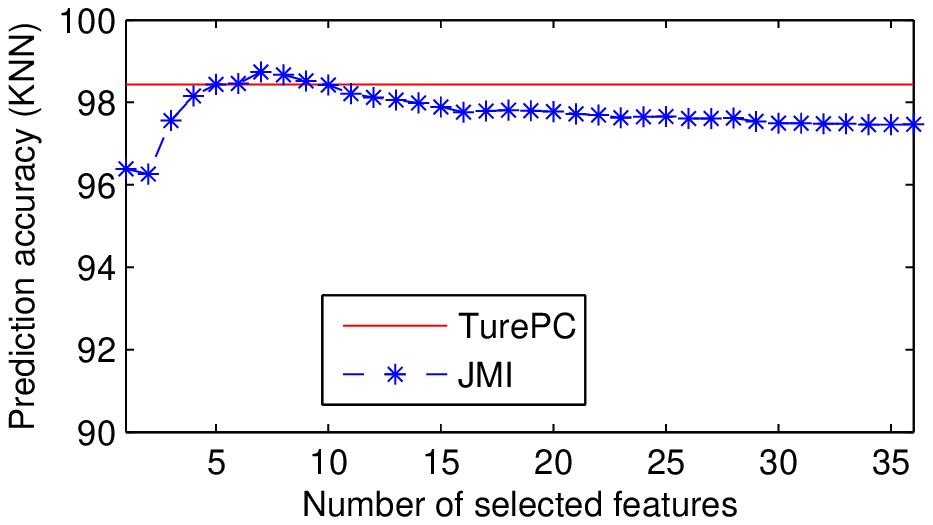}
\includegraphics[height=1.45in]{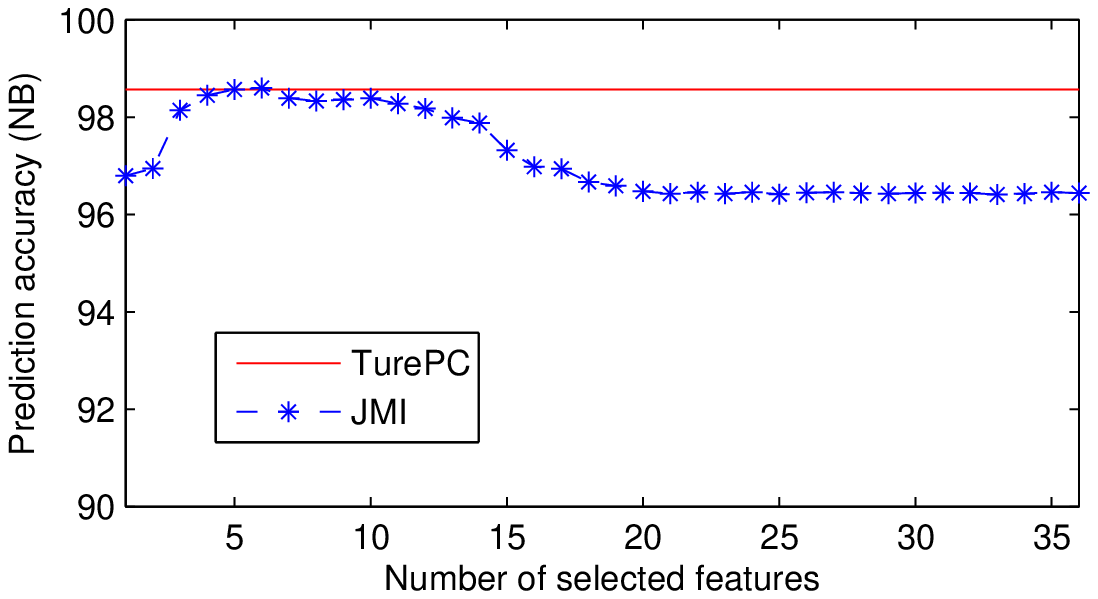}
\caption{JMI and TurePC}
\label{fig6_13}
\end{figure}

\subsubsection{validation of error bounds identified in Section~\ref{sec6}}

In the section, we will examine the proposed error bounds in Section~\ref{sec6}. To achieve the goal, we consider the prediction accuracy of the true PC set of ``HR" as a baseline, since using the PC set of ``HR", the prediction accuracy is almost the same as that using the MB set. Then we check the different prediction accuracies of different feature sets by varying the sizes of the selected feature sets by mMRM, CMIM, and JMI.

From Figures~\ref{fig6_11} to~\ref{fig6_13}, we can see that the prediction accuracies of mRMR, CMIM, and JMI are bounded by the prediction accuracy of the true PC set. The highest accuracy of the three algorithms was achieved with 5 to 8 selected features, where the PC set of ``HR" includes 5 features and the true MB set has 8 features.  Thus, those results further confirm the bounds proposed in Section 6.

\subsection{Evaluation on real-world data}\label{sec8-2}

In this section, we will conduct extensive experiments with 25 real-world datasets to examine the impact of different levels of approximations made by causal and non-causal methods on their performance, the time complexity of both methods,  and the impacts of data sample sizes and different types of datasets on causal and non-causal feature selection algorithms, respectively.
The 25 datasets are divided into four groups: (1) six datasets with large sample sizes and a small feature-to-sample ratio, i.e. ``m$\gg$n"; (2) six datasets with a small sample-to-feature ratio, i.e. ``m$\ll$n"; (3) seven datasets  with multiple classes; (4) six datasets with extremely imbalanced class distributions.

In addition to prediction accuracy used in the previous section, we employ the following metrics to validate all the eight methods:
 \begin{itemize}
\item Number of selected features;
\item Computational efficiency (running time in seconds);
\item AUC: Area Under the ROC (used for imbalanced datasets in Section~\ref{sec8-24});
\item Kappa statistics.
\end{itemize}

The existing stability measures for feature selection always require that the two feature sets under comparison should contain the same number of features, but the eight feature selection methods used in the evaluation return different feature sets of different sizes. Thus,  instead of comparing the stabilities of the features selected by the feature selection methods, in Section~\ref{sec8-2}, we will use the Kappa statistics to measure the stability of a classifier built using the features selected by a feature selection method as an indication of the method's stability.
The Kappa statistic is a measure of consistency amongst different raters, taking into account the agreement occurring by chance~\citep{Cohen1960}.
The statistic is standardized to lie on a -1 to 1 scale, where 1 is perfect agreement, 0 is exactly what would be expected by chance, and negative values indicate agreement less than chance. The detailed value ranges of the Kappa statistics and their corresponding Kappa agreements are shown in  Table~\ref{tb6_1}~\citep{landis1977measurement}.
\begin{table}
\centering
\footnotesize
\caption{Kappa statistic and its corresponding Kappa agreement}
\begin{tabular}{|p{0.17\textwidth}|p{0.17\textwidth}|p{0.1\textwidth}|p{0.1\textwidth}|p{0.1\textwidth}|p{0.14\textwidth}|}
\hline
Kappa statistic	&$<0	$ &0.01-0.20	 &0.21-0.40	&0.61-0.80	 &0.81-0.99\\\hline
Kappa Agreement	 &less than chance agreement	&slight agreement	&moderate agreement	&substantial agreement	&almost perfect agreement\\
\hline
\end{tabular}
\label{tb6_1}
\end{table}

\subsubsection{Datasets with large sample sizes and a small feature-to-sample ratio}\label{sec8-21}

We select six datasets with of large numbers of samples and relatively small numbers of features from the UCI Machine Learning Repository~\citep{bache2013uci}, as shown in Table \ref{tb6_21}.  In the experiment, for mRMR, CMIM, and JMI, since it is hard to decide in advance a suitable parameter $\psi$, i.e., the number of selected features,  for each of these algorithms, we set the user-defined values for $\psi$ to 5, 10, 15, 20, and 25 respectively for the algorithm and choose the feature subset with the highest prediction accuracy as the final feature set selected by the algorithm. 

Tables~\ref{tb6_22} and~\ref{tb6_23} report the prediction accuracy of each algorithm using NBC and KNN, respectively.
Table~\ref{tb6_22} illustrates that with NBC, the non-causal feature selection methods almost have the same performance as the causal feature selection algorithms. IAMB is a bit better than mRMR, CMIM, and JMI. Meanwhile, MMPC and HITON-PC achieve good performance.
From Table~\ref{tb6_23}, we can see that with KNN, MMMB and HITON-MB get better accuracy than mRMR, CMIM, and JMI.

\begin{table}
\centering
\caption{Datasets with large sample sizes and a small feature-to-sample ratio}
\scriptsize
\begin{tabular}{|l|l|l|l|}
\hline
Dataset	&Number of features  &Number of instances	&Number of classes\\\hline
mushroom	&22	&5,644	&2\\\hline
kr-vs-kp	&36	&3,196	&2\\\hline
madelon	&500	&2,000	&2\\\hline
gisstee	&5000	 &7,000	&2\\\hline
spambase	&57	&4,601	         &2\\\hline
bankrupty	&148	&7,063	        &2\\
\hline
\end{tabular}
\label{tb6_21}
\end{table}
\begin{table}[t]
\centering
\caption{Prediction accuracy using NBC}
\scriptsize
\begin{tabular}{|l|l|l|l|l|l|l|l|l|}
\hline
Dataset	&MMPC  &HITON-PC &MMMB &HITON-MB &IAMB &mRMR &CMI &JMI\\\hline
mushroom    &\begin{tabular}[c]{@{}l@{}}0.9940\\$\pm$0.00\end{tabular}    &\begin{tabular}[c]{@{}l@{}}0.9940\\$\pm$0.00\end{tabular}    &\begin{tabular}[c]{@{}l@{}}0.9736\\$\pm$0.01\end{tabular}   &\begin{tabular}[c]{@{}l@{}}0.9736\\$\pm$0.01\end{tabular}   &\begin{tabular}[c]{@{}l@{}}\textbf{0.9995}\\$\pm$0.00\end{tabular}    &\begin{tabular}[c]{@{}l@{}}0.9971\\$\pm$0.00\end{tabular}    &\begin{tabular}[c]{@{}l@{}}0.9957\\$\pm$0.00\end{tabular}  &\begin{tabular}[c]{@{}l@{}}0.9904\\$\pm$0.00\end{tabular}  \\\hline

kr-vs-kp    &\begin{tabular}[c]{@{}l@{}}0.9277\\$\pm$0.02\end{tabular}     &\begin{tabular}[c]{@{}l@{}}0.9277\\$\pm$0.02\end{tabular}     &\begin{tabular}[c]{@{}l@{}}0.9315\\$\pm$0.02\end{tabular}     &\begin{tabular}[c]{@{}l@{}}0.9286\\$\pm$0.02\end{tabular}     &\begin{tabular}[c]{@{}l@{}}\textbf{0.9408}\\$\pm$0.02\end{tabular}     &\begin{tabular}[c]{@{}l@{}}0.9074\\$\pm$0.02\end{tabular}    &\begin{tabular}[c]{@{}l@{}}0.9289\\$\pm$0.02\end{tabular}   &\begin{tabular}[c]{@{}l@{}}0.9289\\$\pm$0.02\end{tabular}       \\\hline

madelon   &\begin{tabular}[c]{@{}l@{}}0.5880\\$\pm$0.03\end{tabular}    &\begin{tabular}[c]{@{}l@{}}0.6000\\$\pm$0.03\end{tabular}   &\begin{tabular}[c]{@{}l@{}}0.5880\\$\pm$0.02\end{tabular}    &\begin{tabular}[c]{@{}l@{}}0.5790\\$\pm$0.02\end{tabular}    &\begin{tabular}[c]{@{}l@{}}0.6070\\$\pm$0.02\end{tabular}    &\begin{tabular}[c]{@{}l@{}}\textbf{0.6195}\\$\pm$0.03\end{tabular}    &\begin{tabular}[c]{@{}l@{}}0.6085\\$\pm$0.03\end{tabular}  &\begin{tabular}[c]{@{}l@{}}0.6000\\$\pm$0.03\end{tabular}       \\\hline

gisstee   &\begin{tabular}[c]{@{}l@{}}0.8856\\$\pm$0.01\end{tabular}     &\begin{tabular}[c]{@{}l@{}}0.8883\\$\pm$0.01\end{tabular}     &\begin{tabular}[c]{@{}l@{}}0.8600\\$\pm$0.01\end{tabular}     &\begin{tabular}[c]{@{}l@{}}0.8533\\$\pm$0.01\end{tabular}   &\begin{tabular}[c]{@{}l@{}}0.8703\\$\pm$0.02\end{tabular}    &\begin{tabular}[c]{@{}l@{}}0.8878\\$\pm$0.01\end{tabular}     &\begin{tabular}[c]{@{}l@{}}\textbf{0.8923}\\$\pm$0.01\end{tabular}   &\begin{tabular}[c]{@{}l@{}}0.8735\\$\pm$0.02\end{tabular}         \\\hline

spambase    &\begin{tabular}[c]{@{}l@{}}0.9081\\$\pm$ 0.01\end{tabular}    &\begin{tabular}[c]{@{}l@{}}\textbf{0.9083}\\$\pm$0.01\end{tabular}     &\begin{tabular}[c]{@{}l@{}}0.8859\\$\pm$0.01\end{tabular}     &\begin{tabular}[c]{@{}l@{}}0.8828\\$\pm$0.01\end{tabular}     &\begin{tabular}[c]{@{}l@{}}0.8942\\$\pm$0.01\end{tabular}     &\begin{tabular}[c]{@{}l@{}}0.8994\\$\pm$0.01\end{tabular}     &\begin{tabular}[c]{@{}l@{}}0.8869\\$\pm$0.01\end{tabular}   &\begin{tabular}[c]{@{}l@{}}0.8911\\$\pm$0.01\end{tabular}    \\\hline

bankrupty    &\begin{tabular}[c]{@{}l@{}}0.8739\\$\pm$0.02\end{tabular}    &\begin{tabular}[c]{@{}l@{}}0.8775\\$\pm$0.02\end{tabular}    &\begin{tabular}[c]{@{}l@{}}0.8390\\$\pm$0.02\end{tabular}    &\begin{tabular}[c]{@{}l@{}}0.8413\\$\pm$0.02\end{tabular}   &\begin{tabular}[c]{@{}l@{}}\textbf{0.8935}\\$\pm$0.00\end{tabular}    &\begin{tabular}[c]{@{}l@{}}0.8863\\$\pm$0.07\end{tabular}    &\begin{tabular}[c]{@{}l@{}}0.8856\\$\pm$0.01\end{tabular}  &\begin{tabular}[c]{@{}l@{}}0.8856\\$\pm$0.01\end{tabular}     \\\hline
\end{tabular}
\label{tb6_22}
\end{table}

\begin{table}
\centering
\caption{Prediction accuracy using KNN}
\scriptsize
\begin{tabular}{|l|l|l|l|l|l|l|l|l|}
\hline
Dataset	&MMPC  &HITON-PC &MMMB &HITON-MB &IAMB &mRMR &CMI &JMI\\\hline

mushroom    &\begin{tabular}[c]{@{}l@{}}\textbf{1.0000}\\$\pm$0.00\end{tabular}     &\begin{tabular}[c]{@{}l@{}}\textbf{1.0000}\\$\pm$0.00\end{tabular}     &\begin{tabular}[c]{@{}l@{}}\textbf{1.0000}\\$\pm$0.00\end{tabular}    &\begin{tabular}[c]{@{}l@{}}\textbf{1.0000}\\$\pm$0.00\end{tabular}   &\begin{tabular}[c]{@{}l@{}}\textbf{1.000}\\$\pm$0.00\end{tabular}    &\begin{tabular}[c]{@{}l@{}}0.9996\\$\pm$0.00\end{tabular}    &\begin{tabular}[c]{@{}l@{}}\textbf{1.0000}\\$\pm$0.00\end{tabular}  &\begin{tabular}[c]{@{}l@{}}\textbf{1.0000}\\$\pm$0.00\end{tabular}      \\\hline

kr-vs-kp      &\begin{tabular}[c]{@{}l@{}}0.8883\\$\pm$0.02\end{tabular}    &\begin{tabular}[c]{@{}l@{}}0.8883\\$\pm$0.02\end{tabular}    &\begin{tabular}[c]{@{}l@{}}\textbf{0.9684}\\$\pm$0.01\end{tabular}   &\begin{tabular}[c]{@{}l@{}}0.9662\\$\pm$0.01\end{tabular}    &\begin{tabular}[c]{@{}l@{}}0.9108\\$\pm$ 0.03\end{tabular}    &\begin{tabular}[c]{@{}l@{}}0.9565\\$\pm$0.01\end{tabular}    &\begin{tabular}[c]{@{}l@{}}0.9603\\$\pm$0.01\end{tabular}  &\begin{tabular}[c]{@{}l@{}}0.9603\\$\pm$0.02\end{tabular}      \\\hline

madelon      &\begin{tabular}[c]{@{}l@{}}0.5450\\$\pm$0.05\end{tabular}    &\begin{tabular}[c]{@{}l@{}}0.5355\\$\pm$0.03\end{tabular}    &\begin{tabular}[c]{@{}l@{}}0.5750\\$\pm$0.05\end{tabular}    &\begin{tabular}[c]{@{}l@{}}0.5655\\$\pm$0.05\end{tabular}   &\begin{tabular}[c]{@{}l@{}}0.5775\\$\pm$0.03\end{tabular}    &\begin{tabular}[c]{@{}l@{}}0.5690\\$\pm$0.03\end{tabular}    &\begin{tabular}[c]{@{}l@{}}0.5670\\$\pm$0.03\end{tabular}  &\begin{tabular}[c]{@{}l@{}}\textbf{0.6235}\\$\pm$0.03\end{tabular}       \\\hline

gisstee        &\begin{tabular}[c]{@{}l@{}}0.9667\\$\pm$0.01\end{tabular}    &\begin{tabular}[c]{@{}l@{}}\textbf{0.9733}\\$\pm$0.01\end{tabular}   &\begin{tabular}[c]{@{}l@{}}0.9617\\$\pm$0.01\end{tabular}    &\begin{tabular}[c]{@{}l@{}}0.9617\\$\pm$0.01\end{tabular}    &\begin{tabular}[c]{@{}l@{}}0.8637\\$\pm$0.01\end{tabular}    &\begin{tabular}[c]{@{}l@{}}0.9347\\$\pm$0.01\end{tabular}    &\begin{tabular}[c]{@{}l@{}}0.9360\\$\pm$0.01\end{tabular}  &\begin{tabular}[c]{@{}l@{}}0.9360\\$\pm$0.01\end{tabular}         \\\hline

spambase   &\begin{tabular}[c]{@{}l@{}}0.9183\\$\pm$0.01\end{tabular}     &\begin{tabular}[c]{@{}l@{}}0.9181\\$\pm$0.01\end{tabular}     &\begin{tabular}[c]{@{}l@{}}0.9233\\$\pm$0.01\end{tabular}     &\begin{tabular}[c]{@{}l@{}}\textbf{0.9242}\\$\pm$0.01\end{tabular}    &\begin{tabular}[c]{@{}l@{}}0.9042\\$\pm$0.01\end{tabular}     &\begin{tabular}[c]{@{}l@{}}0.9189\\$\pm$0.01\end{tabular}     &\begin{tabular}[c]{@{}l@{}}0.9145\\$\pm$0.01\end{tabular}   &\begin{tabular}[c]{@{}l@{}}0.9104\\$\pm$0.01\end{tabular}          \\\hline

bankrupty   &\begin{tabular}[c]{@{}l@{}}0.8697\\$\pm$0.01\end{tabular}    &\begin{tabular}[c]{@{}l@{}}0.8722\\$\pm$0.01\end{tabular}    &\begin{tabular}[c]{@{}l@{}}0.8806\\$\pm$0.01\end{tabular}    &\begin{tabular}[c]{@{}l@{}}0.8792\\$\pm$0.01\end{tabular}    &\begin{tabular}[c]{@{}l@{}}\textbf{0.8840}\\$\pm$0.02\end{tabular}    &\begin{tabular}[c]{@{}l@{}}0.8736\\$\pm$0.02\end{tabular}    &\begin{tabular}[c]{@{}l@{}}0.8622\\$\pm$ 0.01\end{tabular} &\begin{tabular}[c]{@{}l@{}}0.8709\\$\pm$0.01\end{tabular}      \\\hline
\end{tabular}
\label{tb6_23}
\end{table}
\begin{figure}
\centering
\includegraphics[height=1.7in]{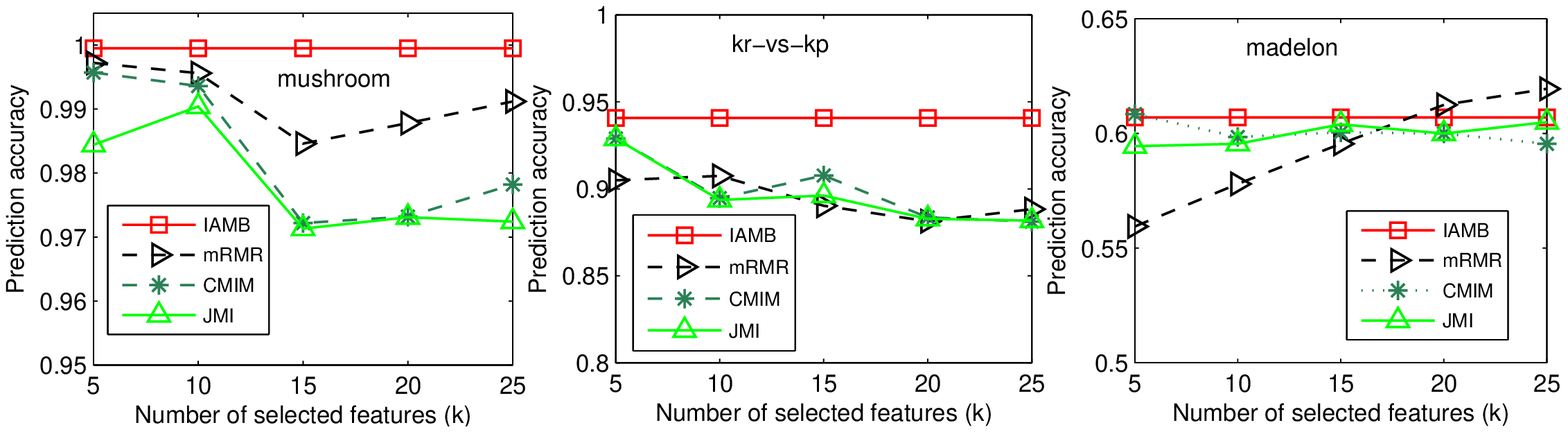}\\
\includegraphics[height=1.7in]{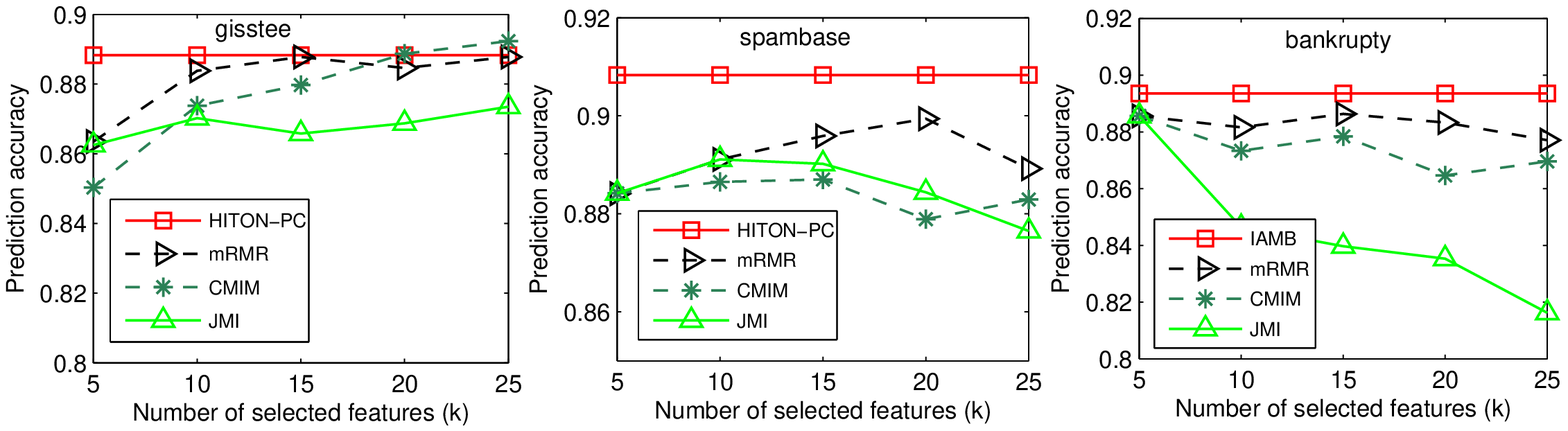}
\caption{Prediction accuracy with different values of $\psi$ using NBC}
\label{fig6_21}
\end{figure}
\begin{figure}
\centering
\includegraphics[height=1.7in]{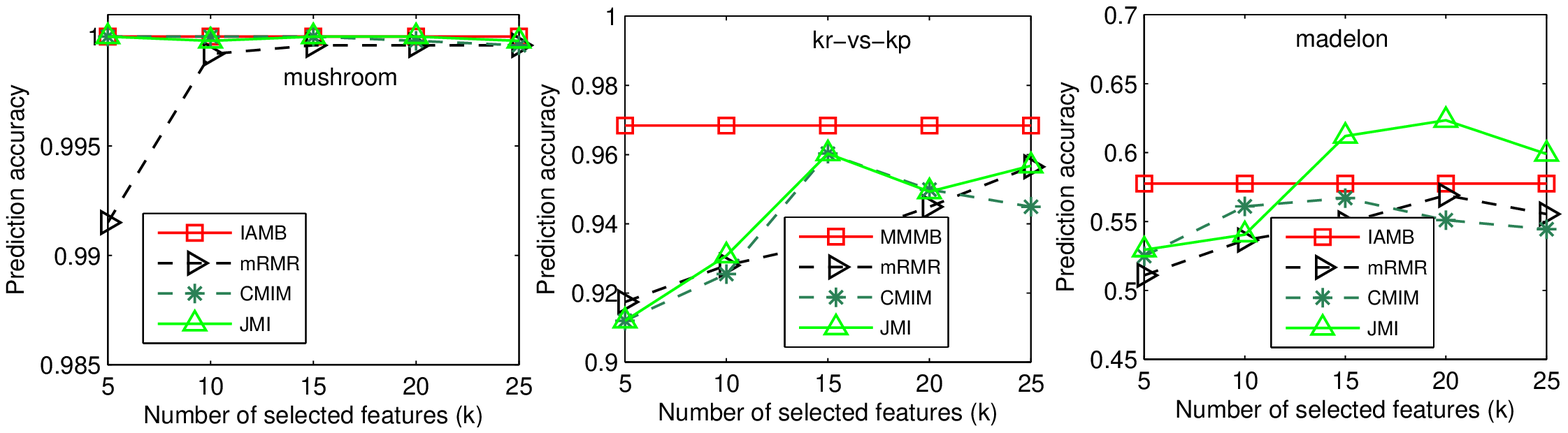}\\
\includegraphics[height=1.7in]{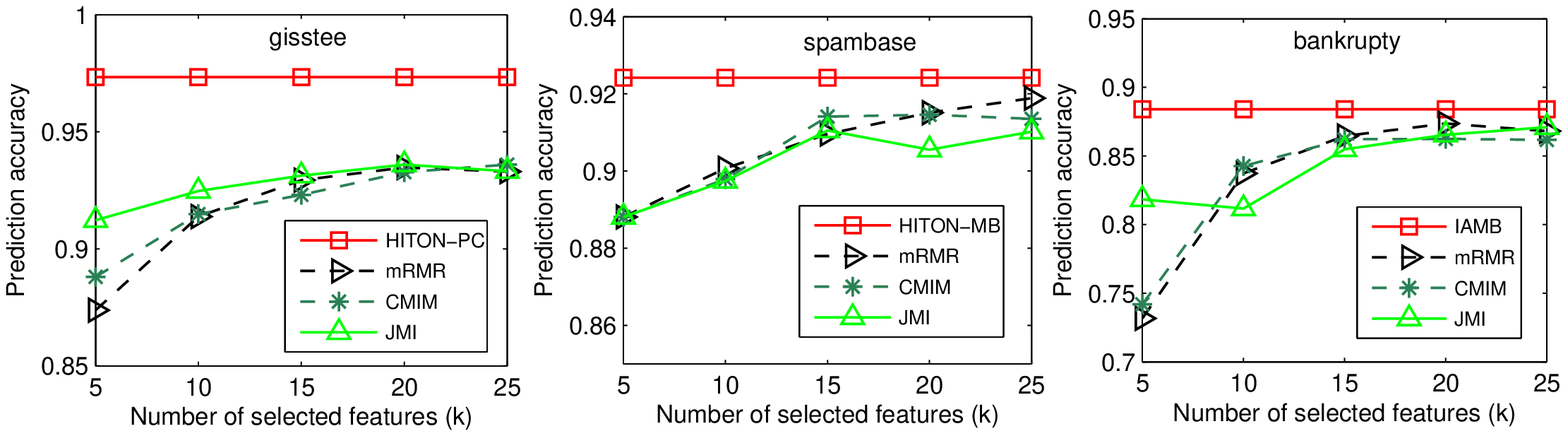}
\caption{Prediction accuracy with different values of $\psi$ using KNN}
\label{fig6_22}
\end{figure}

Tables~\ref{tb6_24} and~\ref{tb6_25} report the Kappa statistic of each algorithm using KNN and NBC, respectively. From Table~\ref{tb6_24}, we can see that  MMPC and HITON-PC achieve better Kappa statistics than the other algorithms except for the madelon dataset. In Table~\ref{tb6_25}, HITON-MB and MMMB are better than the other algorithms, except for the madelon dataset. These results are consistent with those indicated by Tables~\ref{tb6_22} and~\ref{tb6_23}.

Tables~\ref{tb6_27} shows the running time of each algorithm. Clearly, among all causal feature selection algorithms, IAMB is the fastest. However, MMPC, HITON-PC, HITON-MB, and MMMB need to check the subsets of  the feature subset currently selected, therefore they show higher time complexity than IAMB. Moreover, by combining the running time in Tables~\ref{tb6_27}  and the number of features selected in Table~\ref{tb6_26},
we can see that more features are selected,  more expensive the computations of  MMPC, HITON-PC, HITON-MB, and MMMB are. 
Regarding the time complexity of the non-causal feature selection methods, as we discussed at the beginning of Section~\ref{sec5} and in Figure~\ref{fig4-0}, mRMR, CMIM, and JMI use pairwise comparisons, and thus are faster than all causal feature selection algorithms, and this is validated by the result in Table~\ref{tb6_27}.
\begin{table}[t]
\centering
\caption{Kappa statistic using NBC}
\scriptsize
\begin{tabular}{|l|l|l|l|l|l|l|l|l|}
\hline
\multicolumn{1}{|c|}{Dataset} & \multicolumn{1}{c|}{MMPC}                                  & \multicolumn{1}{c|}{HITON-PC}                              & \multicolumn{1}{c|}{MMMB}                                  & \multicolumn{1}{c|}{HITON-MB}                              & \multicolumn{1}{c|}{IAMB}                                  & \multicolumn{1}{c|}{mRMR}                                  & \multicolumn{1}{c|}{CMIM}                                  & \multicolumn{1}{c|}{JMI}                                   \\ \hline
mushroom                      & \begin{tabular}[c]{@{}l@{}}0.9873\\ $\pm$0.01\end{tabular} & \begin{tabular}[c]{@{}l@{}}0.9873\\ $\pm$0.01\end{tabular} & \begin{tabular}[c]{@{}l@{}}0.9434\\ $\pm$0.02\end{tabular} & \begin{tabular}[c]{@{}l@{}}0.9434\\ $\pm$0.02\end{tabular} & \begin{tabular}[c]{@{}l@{}}\textbf{0.9989}\\ $\pm$0.00\end{tabular} & \begin{tabular}[c]{@{}l@{}}0.9940\\ $\pm$0.01\end{tabular} & \begin{tabular}[c]{@{}l@{}}0.9910\\ $\pm$0.01\end{tabular} & \begin{tabular}[c]{@{}l@{}}0.9797\\ $\pm$0.01\end{tabular} \\ \hline
kr-vs-kp                      & \begin{tabular}[c]{@{}l@{}}0.8546\\ $\pm$0.04\end{tabular} & \begin{tabular}[c]{@{}l@{}}0.8546\\ $\pm$0.04\end{tabular} & \begin{tabular}[c]{@{}l@{}}0.8625\\ $\pm$0.03\end{tabular} & \begin{tabular}[c]{@{}l@{}}0.8567\\ $\pm$0.04\end{tabular} & \begin{tabular}[c]{@{}l@{}}\textbf{0.8814}\\ $\pm$0.04\end{tabular} & \begin{tabular}[c]{@{}l@{}}0.8140\\ $\pm$0.04\end{tabular} & \begin{tabular}[c]{@{}l@{}}0.8570\\ $\pm$0.05\end{tabular} & \begin{tabular}[c]{@{}l@{}}0.8570\\ $\pm$0.05\end{tabular} \\ \hline
madelon                       & \begin{tabular}[c]{@{}l@{}}0.1760\\ $\pm$0.05\end{tabular} & \begin{tabular}[c]{@{}l@{}}0.2000\\ $\pm$0.07\end{tabular} & \begin{tabular}[c]{@{}l@{}}0.1760\\ $\pm$0.04\end{tabular} & \begin{tabular}[c]{@{}l@{}}0.1580\\ $\pm$0.04\end{tabular} & \begin{tabular}[c]{@{}l@{}}0.2140\\ $\pm$0.04\end{tabular} & \begin{tabular}[c]{@{}l@{}}\textbf{0.2290}\\ $\pm$0.06\end{tabular} & \begin{tabular}[c]{@{}l@{}}0.2170\\ $\pm$0.07\end{tabular} & \begin{tabular}[c]{@{}l@{}}0.2000\\ $\pm$0.06\end{tabular} \\ \hline
gisstee                       & \begin{tabular}[c]{@{}l@{}}0.8042\\ $\pm$0.01\end{tabular} & \begin{tabular}[c]{@{}l@{}}\textbf{0.8056}\\ $\pm$0.01\end{tabular} & \begin{tabular}[c]{@{}l@{}}0.7762\\ $\pm$0.02\end{tabular} & \begin{tabular}[c]{@{}l@{}}0.7723\\ $\pm$0.03\end{tabular} & \begin{tabular}[c]{@{}l@{}}0.7407\\ $\pm$0.03\end{tabular} & \begin{tabular}[c]{@{}l@{}}0.7760\\ $\pm$0.02\end{tabular} & \begin{tabular}[c]{@{}l@{}}0.7850\\ $\pm$0.02\end{tabular} & \begin{tabular}[c]{@{}l@{}}0.7373\\ $\pm$0.03\end{tabular} \\ \hline
spambase                      & \begin{tabular}[c]{@{}l@{}}0.8055\\ $\pm$0.02\end{tabular} & \begin{tabular}[c]{@{}l@{}}\textbf{0.8059}\\ $\pm$0.02\end{tabular} & \begin{tabular}[c]{@{}l@{}}0.7581\\ $\pm$0.02\end{tabular} & \begin{tabular}[c]{@{}l@{}}0.7515\\ $\pm$0.02\end{tabular} & \begin{tabular}[c]{@{}l@{}}0.7769\\ $\pm$0.02\end{tabular} & \begin{tabular}[c]{@{}l@{}}0.7865\\ $\pm$0.01\end{tabular} & \begin{tabular}[c]{@{}l@{}}0.7590\\ $\pm$0.02\end{tabular} & \begin{tabular}[c]{@{}l@{}}0.7683\\ $\pm$0.03\end{tabular} \\ \hline
bankrupty                     & \begin{tabular}[c]{@{}l@{}}\textbf{0.4505}\\ $\pm$0.04\end{tabular} & \begin{tabular}[c]{@{}l@{}}0.4502\\ $\pm$0.05\end{tabular} & \begin{tabular}[c]{@{}l@{}}0.4167\\ $\pm$0.05\end{tabular} & \begin{tabular}[c]{@{}l@{}}0.4161\\ $\pm$0.06\end{tabular} & \begin{tabular}[c]{@{}l@{}}0.1551\\ $\pm$0.05\end{tabular} & \begin{tabular}[c]{@{}l@{}}0.3000\\ $\pm$0.05\end{tabular} & \begin{tabular}[c]{@{}l@{}}0.1423\\ $\pm$0.04\end{tabular} & \begin{tabular}[c]{@{}l@{}}0.1398\\ $\pm$0.05\end{tabular} \\ \hline
\end{tabular}
\label{tb6_24}
\end{table}
\begin{table}
\centering
\caption{Kappa statistic using KNN}
\scriptsize
\begin{tabular}{|l|l|l|l|l|l|l|l|l|}
\hline
Dataset   & \multicolumn{1}{c|}{MMPC}                                  & \multicolumn{1}{c|}{HITON-PC}                              & \multicolumn{1}{c|}{MMMB}                                  & \multicolumn{1}{c|}{HITON-MB}                              & \multicolumn{1}{c|}{IAMB}                                  & \multicolumn{1}{c|}{mRMR}                                  & \multicolumn{1}{c|}{CMIM}                                  & \multicolumn{1}{c|}{JMI}                                   \\ \hline
mushroom  & \begin{tabular}[c]{@{}l@{}}\textbf{1.00}\\ $\pm$0.00\end{tabular}   & \begin{tabular}[c]{@{}l@{}}\textbf{1.00}\\ $\pm$0.00\end{tabular}   & \begin{tabular}[c]{@{}l@{}}\textbf{1.00}\\ $\pm$0.00\end{tabular}   & \begin{tabular}[c]{@{}l@{}}\textbf{1.00}\\ $\pm$0.00\end{tabular}   & \begin{tabular}[c]{@{}l@{}}\textbf{1.00}\\ $\pm$0.00\end{tabular}   & \begin{tabular}[c]{@{}l@{}}0.9992\\ $\pm$0.00\end{tabular} & \begin{tabular}[c]{@{}l@{}}\textbf{1.00}\\ $\pm$0.00\end{tabular}   & \begin{tabular}[c]{@{}l@{}}\textbf{1.00}\\ $\pm$0.00\end{tabular}   \\ \hline
kr-vs-kp  & \begin{tabular}[c]{@{}l@{}}0.7742\\ $\pm$0.04\end{tabular} & \begin{tabular}[c]{@{}l@{}}0.7742\\ $\pm$0.04\end{tabular} & \begin{tabular}[c]{@{}l@{}}\textbf{0.9366}\\ $\pm$0.02\end{tabular} & \begin{tabular}[c]{@{}l@{}}0.9322\\ $\pm$0.03\end{tabular} & \begin{tabular}[c]{@{}l@{}}0.8205\\ $\pm$0.06\end{tabular} & \begin{tabular}[c]{@{}l@{}}0.9129\\ $\pm$0.02\end{tabular} & \begin{tabular}[c]{@{}l@{}}0.9204\\ $\pm$0.02\end{tabular} & \begin{tabular}[c]{@{}l@{}}0.9204\\ $\pm$0.03\end{tabular} \\ \hline
madelon   & \begin{tabular}[c]{@{}l@{}}0.0900\\ $\pm$0.10\end{tabular} & \begin{tabular}[c]{@{}l@{}}0.0710\\ $\pm$0.06\end{tabular} & \begin{tabular}[c]{@{}l@{}}0.1500\\ $\pm$0.07\end{tabular} & \begin{tabular}[c]{@{}l@{}}0.1310\\ $\pm$0.09\end{tabular} & \begin{tabular}[c]{@{}l@{}}0.1550\\ $\pm$0.05\end{tabular} & \begin{tabular}[c]{@{}l@{}}0.1380\\ $\pm$0.06\end{tabular} & \begin{tabular}[c]{@{}l@{}}0.1340\\ $\pm$0.06\end{tabular} & \begin{tabular}[c]{@{}l@{}}\textbf{0.2470}\\ $\pm$0.07\end{tabular} \\ \hline
gisstee   & \begin{tabular}[c]{@{}l@{}}0.9333\\ $\pm$0.02\end{tabular} & \begin{tabular}[c]{@{}l@{}}\textbf{0.9467}\\ $\pm$0.02\end{tabular} & \begin{tabular}[c]{@{}l@{}}0.9341\\ $\pm$0.03\end{tabular} & \begin{tabular}[c]{@{}l@{}}0.9356\\ $\pm$0.02\end{tabular} & \begin{tabular}[c]{@{}l@{}}0.7273\\ $\pm$0.03\end{tabular} & \begin{tabular}[c]{@{}l@{}}0.8693\\ $\pm$0.02\end{tabular} & \begin{tabular}[c]{@{}l@{}}0.8720\\ $\pm$0.02\end{tabular} & \begin{tabular}[c]{@{}l@{}}0.8720\\ $\pm$0.02\end{tabular} \\ \hline
spambase  & \begin{tabular}[c]{@{}l@{}}0.8281\\ $\pm$0.02\end{tabular} & \begin{tabular}[c]{@{}l@{}}0.8277\\ $\pm$0.02\end{tabular} & \begin{tabular}[c]{@{}l@{}}0.8393\\ $\pm$0.02\end{tabular} & \begin{tabular}[c]{@{}l@{}}\textbf{0.8410}\\ $\pm$0.02\end{tabular} & \begin{tabular}[c]{@{}l@{}}0.7965\\ $\pm$0.02\end{tabular} & \begin{tabular}[c]{@{}l@{}}0.8206\\ $\pm$0.02\end{tabular} & \begin{tabular}[c]{@{}l@{}}0.8172\\ $\pm$0.02\end{tabular} & \begin{tabular}[c]{@{}l@{}}0.8101\\ $\pm$0.02\end{tabular} \\ \hline
bankrupty & \begin{tabular}[c]{@{}l@{}}0.3311\\ $\pm$0.03\end{tabular} & \begin{tabular}[c]{@{}l@{}}0.3344\\ $\pm$0.05\end{tabular} & \begin{tabular}[c]{@{}l@{}}0.3445\\ $\pm$0.05\end{tabular} & \begin{tabular}[c]{@{}l@{}}0.3409\\ $\pm$0.06\end{tabular} & \begin{tabular}[c]{@{}l@{}}\textbf{0.3883}\\ $\pm$0.05\end{tabular} & \begin{tabular}[c]{@{}l@{}}0.3820\\ $\pm$0.04\end{tabular} & \begin{tabular}[c]{@{}l@{}}0.3615\\ $\pm$0.04\end{tabular} & \begin{tabular}[c]{@{}l@{}}0.3138\\ $\pm$0.04\end{tabular} \\ \hline
\end{tabular}
\label{tb6_25}
\end{table}

From Table~\ref{tb6_26}, we can see that MMPC and HITON-PC select fewer features than MMMB and HITON-MB, while IAMB selects the fewest features among the eight algorithms.
For the gisstee dataset,  MMPC, HITON-PC, MMMB, and HITON-MB select significantly more features than the other algorithms, since the features in the dataset are highly correlated. IAMB only selects two features, because the conditioning set is large and requires large number of data samples, and thus may lead to unreliable many conditional tests. This also explains why the prediction accuracy of IAMB in Table~\ref{tb6_23} is significantly low than the other algorithms.
\begin{table}[t]
\centering
\caption{Number of selected features (``A/B" denotes that ``A" represents the number of features with the highest accuracy corresponding to an algorithm using NBC and ``B" is the number of features with the highest accuracy corresponding to an algorithm using KNN)}
\scriptsize
\begin{tabular}{|l|l|l|l|l|l|l|l|l|}
\hline
Dataset	&MMPC  &HITON-PC &MMMB &HITON-MB &IAMB &mRMR &CMI &JMI\\\hline
mushroom  &10	&10	&20	&20	&3	&5/15	&5/5	&10/5\\\hline
kr-vs-kp      &8         &8	&19	&19	&7	&10/25	&5/15	&5/15\\\hline
madelon     &5	&5	&6	&5	&6	&25/20	&5/15	&20/20\\\hline
gisstee       &295	&294	&1384	 &1402 &2	&15/20	&25/25	&25/20\\\hline
spambase  &24	&24	&45	&45	&8	&20/25	&15/20	&10/15\\\hline
bankrupty  &29	&28	&60	&56	&9	&15/20	&5/15	&5/25\\\hline
\end{tabular}
\label{tb6_26}
\end{table}
\begin{table}
\centering
\caption{Running time (in seconds)}
\scriptsize
\begin{tabular}{|l|l|l|l|l|l|l|l|l|}
\hline
Dataset	&MMPC  &HITON-PC &MMMB &HITON-MB &IAMB &mRMR &CMI &JMI\\\hline
mushroom	&0.89	&1.19	&42.98	&44.12	&0.16	  &0.03	  &0.01	&0.03\\\hline
kr-vs-kp	&0.31	  &0.33	  &6.87	&6.21	&0.43	  &0.04	&0.03	  &0.07\\\hline
madelon	&0.18	  &0.21  	&0.87	  &0.9448	&3.07	 &0.4	&0.03	  &1.53 \\\hline
gisstee	&32,684	&65,308	&50,929	&107,870	&12.90	&8.38	 &1.32	&52\\\hline
spambase	&35	&37	&200	&203	&0.7648	&0.09	  &0.06	&0.24\\\hline
bankrupty	&112	&95	&296	&239	&2.06	&0.31	&0.11	&1.27\\
\hline
\end{tabular}
\label{tb6_27}
\end{table}

Finally, Figures~\ref{fig6_21} and~\ref{fig6_22} report the predication accuracy of the causal feature selection methods (with the highest prediction accuracy) in comparison with the three non-causal feature selection methods, mRMR, CMIM and JMI when the number of features selected by the three methods are varied. 
From Figures~\ref{fig6_21} and~\ref{fig6_22}, we can see that, either with KNN or NBC, for all datasets except for madelon,  causal feature selection method always outperforms all the three non-causal feature selection methods regardless the number of selected features specified for these non-causal methods. This result has demonstrated that causal feature selection methods, when the dataset contains sufficient large number of samples, the features selected by them would be closer to the optimal feature set, i.e. the MB of the class attribute.


\subsubsection{Dataset with high dimensionality and small number of data samples}\label{sec8-22}

In this section, we will evaluate the eight feature selection methods using the six datasets with high dimensionality and relatively small numbers of samples. Table~\ref{tb6_31}  provides a summary of the datasets.
In the following tables reporting the results,  ``-" denotes that an algorithm fails to obtain any result with a dataset because of excessive running time. We will do the same for the experiments in Sections~\ref{sec8-23},~\ref{sec8-24}, and ~\ref{sec8-25}.
Since mRMR, CMIM, and JMI use a user-defined parameter $\psi$ to control the size of features selected and the datasets in Table~\ref{tb6_31}  are high dimensionality, we set $\psi$ to the top 5, 10, 15, $\cdots$, 35, and 40 respectively, then report the results about the feature subset with the highest prediction accuracy.

In Tables~\ref{tb6_32} and~\ref{tb6_33}, we can see that using KNN and NBC, the non-causal feature selection methods, mRMR, CMIM, and JMI, all outperform the causal feature selection methods,  MMPC, HITON-PC, MMMB, HITON-MB, and IAMB. This illustrates that with datasets of high dimensionality and small sample size,  as the number of data instances is not enough to support causal feature selection algorithms for  reliable conditional independence tests, whereas  mRMR, CMIM, and JMI  can cope with such datasets. This validates our analysis of sample requirement in Section~\ref{sec5}.  

In addition with mRMR, CMIM, and JMI, we can tune the parameter $\psi$ to control the size of the selected feature set for the trade-off between search efficiency and prediction accuracy. Accordingly, from Tables~\ref{tb6_34} and~\ref{tb6_35}, mRMR, CMIM, and JMI can get more stable features, as indicated by the  better Kappa statistic than MMPC, HITON-PC, MMMB, HITON-MB, and IAMB. Meanwhile, Table~\ref{tb6_36} shows that the computational costs of  HITON-PC, MMMB, HITON-MB are very expensive and even prohibitive on some datasets, such as prostate, dorothea, and leukemia. The explanation is that the class attribute in each of the datasets may have a large PC set or MB set, as shown in Table~\ref{tb6_37}, then this leads to that MMPC, HITON-PC, MMMB, and HITON-MB needs to check an exponential number of subsets.

Figures~\ref{fig6_31} and~\ref{fig6_32} report the predication accuracy of the causal feature selection methods (with the highest prediction accuracy) in comparison with the three non-causal feature selection methods, mRMR, CMIM and JMI when the number of features selected by the three methods are varied. 

From Figures~\ref{fig6_31} and~\ref{fig6_32}, we can see that, either with KNN or NBC, for all datasets,  the three non-causal feature selection methods always outperforms most of causal feature selection methods. This result has demonstrated that causal feature selection methods, when the dataset contains high dimensionality and relatively small number of data samples, the  causal feature selection methods is worse than non-causal feature selection methods. 

The results in Figures~\ref{fig6_21},~\ref{fig6_22},~\ref{fig6_31}, and~\ref{fig6_32} validate that on the one hand, with a large dataset with sufficient number of samples, causal feature selection methods tend to find an exact MB; on the other hand, non-causal feature selection methods can deal with datasets with a small number of data samples and high-dimensionality better.

\begin{table}
\centering
\caption{Dataset with high dimensionality and small data sample sizes}
\scriptsize
\begin{tabular}{|l|l|l|l|}
\hline
Dataset	&Number of features  &Number of instances	&Number of classes\\\hline
prostate	&6,033	&102	&2\\\hline

dexter	&20,000	&300	&2\\\hline

arcene	&10,000	&100	&2\\\hline

dorothea	&100,000	&800	&2\\\hline


leukemia       &7,070	&72	&2\\\hline

breast-cancer	&17817	&286	&2\\

\hline
\end{tabular}
\label{tb6_31}
\end{table}

 \begin{table}
\centering
\caption{Prediction accuracy using NBC}
\scriptsize
\begin{tabular}{|l|l|l|l|l|l|l|l|l|}
\hline
Dataset	&MMPC  &HITON-PC &MMMB &HITON-MB &IAMB &mRMR &CMI &JMI\\\hline

prostate	&\begin{tabular}[c]{@{}l@{}}0.9100\\$\pm$0.10\end{tabular}	&\begin{tabular}[c]{@{}l@{}}0.9300\\$\pm$0.10\end{tabular}	&\begin{tabular}[c]{@{}l@{}}0.9013\\$\pm$0.10\end{tabular}	 &\begin{tabular}[c]{@{}l@{}}0.8667\\$\pm$0.10\end{tabular}	&\begin{tabular}[c]{@{}l@{}}0.9100\\$\pm$0.12\end{tabular}	&\begin{tabular}[c]{@{}l@{}}\textbf{0.9500}\\$\pm$0.09\end{tabular}	 &\begin{tabular}[c]{@{}l@{}}0.9400\\$\pm$0.08\end{tabular}	&\begin{tabular}[c]{@{}l@{}}\textbf{0.9500}\\$\pm$0.09\end{tabular}      \\\hline

dexter	&\begin{tabular}[c]{@{}l@{}}0.8567\\$\pm$0.06\end{tabular}	&\begin{tabular}[c]{@{}l@{}}0.8567\\$\pm$0.06\end{tabular}	&\begin{tabular}[c]{@{}l@{}}0.8533\\$\pm$0.06\end{tabular}	 &\begin{tabular}[c]{@{}l@{}}0.8667\\$\pm$0.06\end{tabular}	&\begin{tabular}[c]{@{}l@{}}0.7900\\$\pm$0.04\end{tabular}	&\begin{tabular}[c]{@{}l@{}}0.9167\\$\pm$0.04\end{tabular}	 &\begin{tabular}[c]{@{}l@{}}\textbf{0.9200}\\$\pm$0.05\end{tabular}	&\begin{tabular}[c]{@{}l@{}}0.8933\\$\pm$0.06\end{tabular}       \\\hline

arcene	&\begin{tabular}[c]{@{}l@{}}0.7136\\$\pm$0.17\end{tabular}	&\begin{tabular}[c]{@{}l@{}}0.7527\\$\pm$0.16\end{tabular}	&\begin{tabular}[c]{@{}l@{}}0.7036\\$\pm$0.16\end{tabular}	 &\begin{tabular}[c]{@{}l@{}}0.7005\\$\pm$0.11\end{tabular}	&\begin{tabular}[c]{@{}l@{}}0.7305\\$\pm$0.16\end{tabular}	&\begin{tabular}[c]{@{}l@{}}0.7627\\$\pm$0.12\end{tabular}	 &\begin{tabular}[c]{@{}l@{}}\textbf{0.7718}\\$\pm$0.13\end{tabular}	&\begin{tabular}[c]{@{}l@{}}0.6896\\$\pm$ 0.10\end{tabular}      \\\hline

dorothea	&\begin{tabular}[c]{@{}l@{}}0.9287\\$\pm$0.02\end{tabular}	&\begin{tabular}[c]{@{}l@{}}0.9325\\$\pm$0.02\end{tabular}	&-	                                &-	                                &\begin{tabular}[c]{@{}l@{}}0.9352\\$\pm$0.03\end{tabular}	&\begin{tabular}[c]{@{}l@{}}0.9426\\$\pm$0.02\end{tabular}	&\begin{tabular}[c]{@{}l@{}}\textbf{0.9463}\\$\pm$0.02\end{tabular}	 &\begin{tabular}[c]{@{}l@{}}0.9376\\$\pm$0.02\end{tabular} \\\hline


leukemia      & \begin{tabular}[c]{@{}l@{}}0.9304\\$\pm$0.07\end{tabular}	&-	         &-	          &-	        &\begin{tabular}[c]{@{}l@{}}0.9446\\$\pm$0.10\end{tabular}                        	 &\begin{tabular}[c]{@{}l@{}}0.9714\\$\pm$0.09\end{tabular}	  &\begin{tabular}[c]{@{}l@{}}0.9714\\$\pm$0.06\end{tabular}	    &\begin{tabular}[c]{@{}l@{}}\textbf{0.9857}\\$\pm$0.05\end{tabular}        \\\hline

breast-cancer &\begin{tabular}[c]{@{}l@{}}0.8360\\$\pm$0.05\end{tabular}	&\begin{tabular}[c]{@{}l@{}}0.7969\\$\pm$0.10\end{tabular}	&\begin{tabular}[c]{@{}l@{}}0.8395\\$\pm$0.05\end{tabular}	 &\begin{tabular}[c]{@{}l@{}}0.7900\\$\pm$0.10\end{tabular}	&\begin{tabular}[c]{@{}l@{}}0.8209\\$\pm$0.07\end{tabular}	&\begin{tabular}[c]{@{}l@{}}\textbf{0.8778}\\$\pm$0.03\end{tabular}	 &\begin{tabular}[c]{@{}l@{}}\textbf{0.8778}\\$\pm$0.04\end{tabular}	&\begin{tabular}[c]{@{}l@{}}0.8740\\$\pm$0.04 \end{tabular}       \\
\hline
\end{tabular}
\label{tb6_32}
\end{table}

 \begin{table}
\centering
\caption{Prediction accuracy using KNN}
\scriptsize
\begin{tabular}{|l|l|l|l|l|l|l|l|l|}
\hline
Dataset	&MMPC  &HITON-PC &MMMB &HITON-MB &IAMB &mRMR &CMI &JMI\\\hline

prostate	          &\begin{tabular}[c]{@{}l@{}}0.8909\\$\pm$0.10\end{tabular}	&\begin{tabular}[c]{@{}l@{}}0.9000\\$\pm$0.10\end{tabular}	&\begin{tabular}[c]{@{}l@{}}0.8710\\$\pm$0.10\end{tabular}	 &\begin{tabular}[c]{@{}l@{}}0.9030\\$\pm$0.10\end{tabular}	&\begin{tabular}[c]{@{}l@{}}0.9400\\$\pm$0.11\end{tabular}	&\begin{tabular}[c]{@{}l@{}}\textbf{0.9500}\\$\pm$0.09\end{tabular}	 &\begin{tabular}[c]{@{}l@{}}\textbf{0.9500}\\$\pm$0.07\end{tabular}	&\begin{tabular}[c]{@{}l@{}}\textbf{0.9500}\\$\pm$0.09\end{tabular}        \\\hline

dexter	          &\begin{tabular}[c]{@{}l@{}}0.7900\\$\pm$0.06\end{tabular}	&\begin{tabular}[c]{@{}l@{}}0.7733\\$\pm$0.08\end{tabular}		&\begin{tabular}[c]{@{}l@{}}0.7900\\$\pm$0.06\end{tabular}	 &\begin{tabular}[c]{@{}l@{}}0.7767\\$\pm$0.08\end{tabular}	&\begin{tabular}[c]{@{}l@{}}0.7367\\$\pm$0.08\end{tabular}	&\begin{tabular}[c]{@{}l@{}}0.8833\\$\pm$0.06\end{tabular}	 &\begin{tabular}[c]{@{}l@{}}0.8833\\$\pm$0.06\end{tabular}	&\begin{tabular}[c]{@{}l@{}}\textbf{0.8900}\\$\pm$0.05\end{tabular}        \\\hline

arcene	          &\begin{tabular}[c]{@{}l@{}}0.6623\\$\pm$0.17\end{tabular}	&\begin{tabular}[c]{@{}l@{}}0.6410\\$\pm$0.15\end{tabular}	&\begin{tabular}[c]{@{}l@{}}0.6523\\$\pm$0.18\end{tabular}	 &\begin{tabular}[c]{@{}l@{}}0.6410\\$\pm$0.15\end{tabular}	&\begin{tabular}[c]{@{}l@{}}0.7094\\$\pm$0.16\end{tabular}	&\begin{tabular}[c]{@{}l@{}}0.7616\\$\pm$0.09\end{tabular}	 &\begin{tabular}[c]{@{}l@{}}\textbf{0.8017}\\$\pm$0.09\end{tabular}	&\begin{tabular}[c]{@{}l@{}}0.7796\\$\pm$0.08\end{tabular}       \\\hline

dorothea	          &\begin{tabular}[c]{@{}l@{}}0.9099\\$\pm$0.03\end{tabular}	&\begin{tabular}[c]{@{}l@{}}0.9212\\$\pm$0.03\end{tabular}	&-	           &-	          &\begin{tabular}[c]{@{}l@{}}0.9176\\$\pm$0.04\end{tabular}	&\begin{tabular}[c]{@{}l@{}}0.9313\\$\pm$0.02\end{tabular}	&\begin{tabular}[c]{@{}l@{}}0.9263\\$\pm$0.01\end{tabular}	 &\begin{tabular}[c]{@{}l@{}}\textbf{0.9363}\\$\pm$0.02\end{tabular}       \\\hline


leukemia                  &\begin{tabular}[c]{@{}l@{}}\textbf{0.9714}\\$\pm$0.06\end{tabular}	&-	          &-	           &-	          &\begin{tabular}[c]{@{}l@{}}0.9446\\$\pm$0.10\end{tabular}	 &\begin{tabular}[c]{@{}l@{}}\textbf{0.9714}\\$\pm$0.06\end{tabular}	&\begin{tabular}[c]{@{}l@{}}\textbf{0.9714}\\$\pm$0.06\end{tabular}	&\begin{tabular}[c]{@{}l@{}}\textbf{0.9714}\\$\pm$0.06\end{tabular} \\\hline

breast-cancer        &\begin{tabular}[c]{@{}l@{}}0.8217\\$\pm$0.09\end{tabular}	&\begin{tabular}[c]{@{}l@{}}0.8005\\$\pm$0.06\end{tabular}	&\begin{tabular}[c]{@{}l@{}}0.8184\\$\pm$0.06\end{tabular}	 &\begin{tabular}[c]{@{}l@{}}0.7901\\$\pm$0.07\end{tabular}       &\begin{tabular}[c]{@{}l@{}}0.8070\\$\pm$0.09\end{tabular} 	&\begin{tabular}[c]{@{}l@{}}0.8739\\$\pm$0.05\end{tabular}	 &\begin{tabular}[c]{@{}l@{}}0.8671\\$\pm$0.04\end{tabular}	&\begin{tabular}[c]{@{}l@{}}\textbf{0.8776}\\$\pm$0.05\end{tabular}       \\
\hline
\end{tabular}
\label{tb6_33}
\end{table}

\begin{figure}
\centering
\includegraphics[height=1.7in]{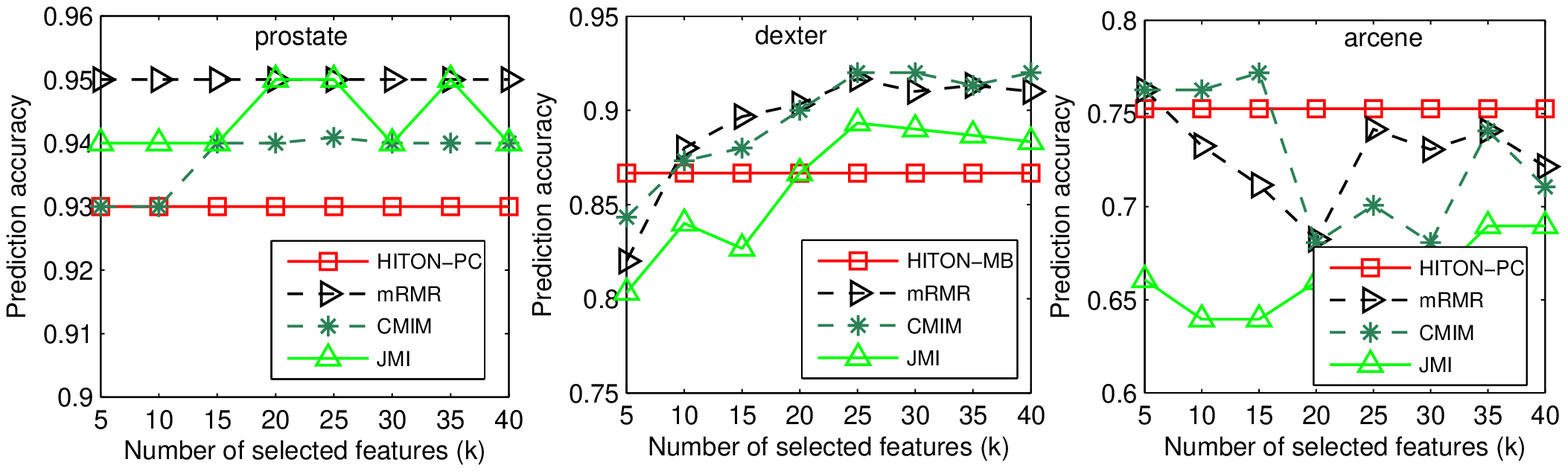}\\
\includegraphics[height=1.7in,width=5.85in]{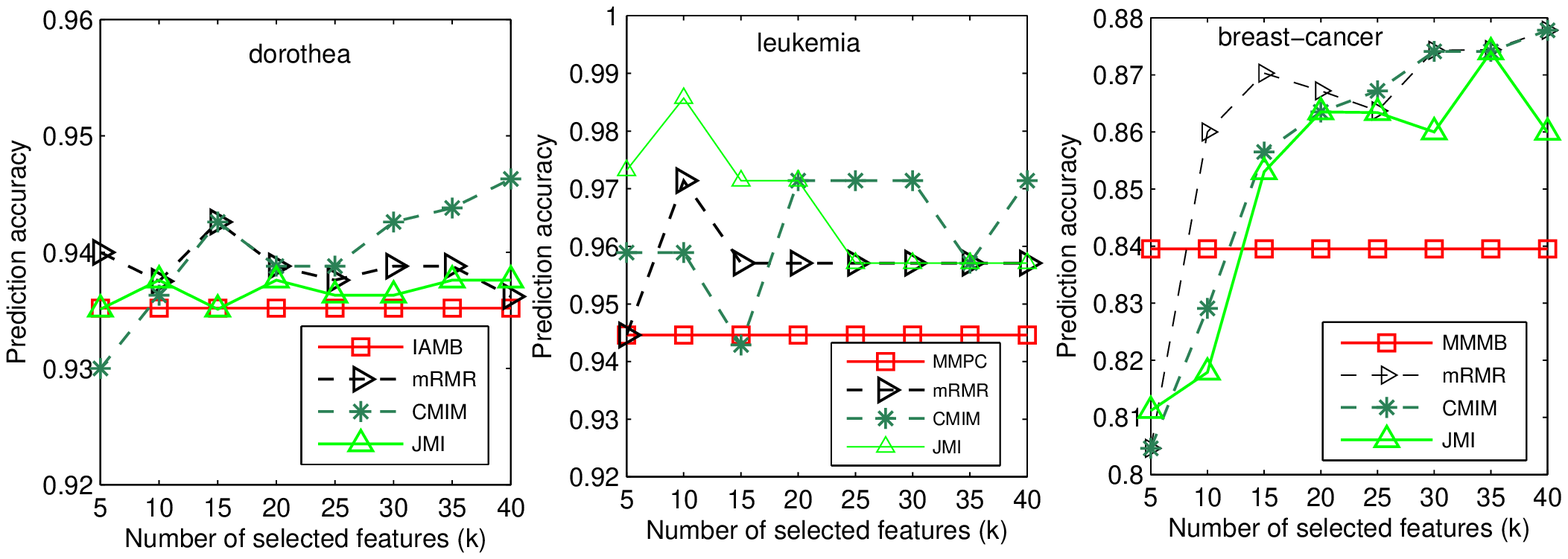}
\caption{Prediction accuracy with different values of$\psi$ using NBC}
\label{fig6_31}
\end{figure}
\begin{figure}[t]
\centering
\includegraphics[height=1.7in,width=5.85in]{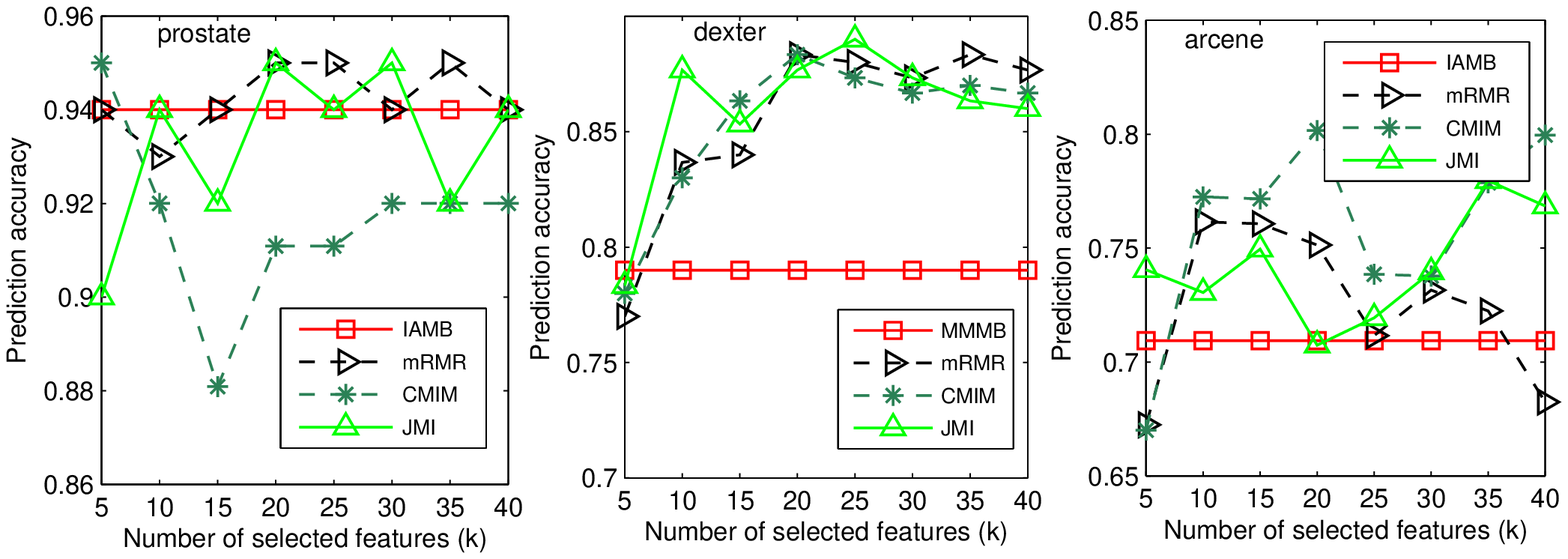}\\
\includegraphics[height=1.7in,width=5.85in]{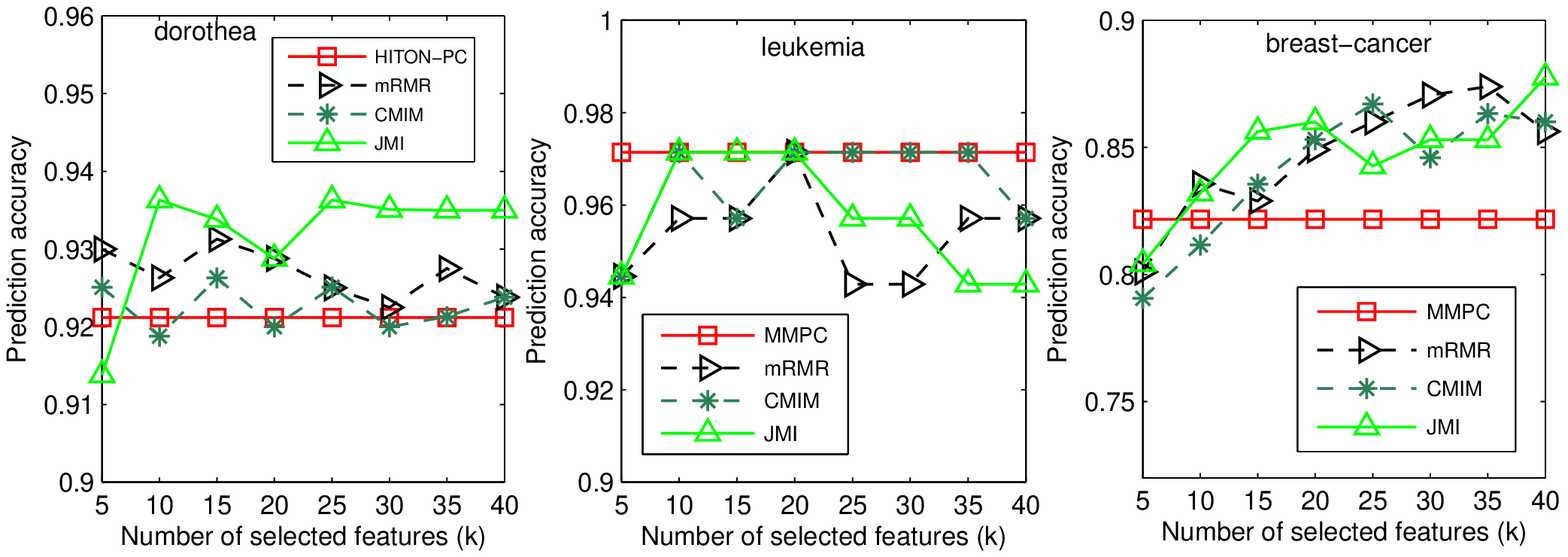}
\caption{Prediction accuracy with different values of $\psi$ using KNN}
\label{fig6_32}
\end{figure}

\begin{table}
\centering
\caption{Kappa statistic using NBC}
\scriptsize
\begin{tabular}{|l|l|l|l|l|l|l|l|l|}
\hline
\multicolumn{1}{|c|}{Dataset} & \multicolumn{1}{c|}{MMPC}                                  & \multicolumn{1}{c|}{HITON-PC}                              & \multicolumn{1}{c|}{MMMB}                                  & \multicolumn{1}{c|}{HITON-MB}                              & \multicolumn{1}{c|}{IAMB}                                  & \multicolumn{1}{c|}{mRMR}                                  & \multicolumn{1}{c|}{CMIM}                                  & \multicolumn{1}{c|}{JMI}                                   \\ \hline
prostate                      & \begin{tabular}[c]{@{}l@{}}0.8200\\ $\pm$0.20\end{tabular} & \begin{tabular}[c]{@{}l@{}}0.8600\\ $\pm$0.19\end{tabular} & \begin{tabular}[c]{@{}l@{}}0.7920\\ $\pm$0.25\end{tabular} & \begin{tabular}[c]{@{}l@{}}0.7805\\ $\pm$0.26\end{tabular} & \begin{tabular}[c]{@{}l@{}}0.8200\\ $\pm$0.24\end{tabular} & \begin{tabular}[c]{@{}l@{}}0.9000\\ $\pm$0.17\end{tabular} & \begin{tabular}[c]{@{}l@{}}0.8800\\ $\pm$0.17\end{tabular} & \begin{tabular}[c]{@{}l@{}}0.9000\\ $\pm$0.17\end{tabular} \\ \hline
dexter                        & \begin{tabular}[c]{@{}l@{}}0.7133\\ $\pm$0.11\end{tabular} & \begin{tabular}[c]{@{}l@{}}0.7133\\ $\pm$0.11\end{tabular} & \begin{tabular}[c]{@{}l@{}}0.7122\\ $\pm$0.12\end{tabular} & \begin{tabular}[c]{@{}l@{}}0.7333\\ $\pm$0.13\end{tabular} & \begin{tabular}[c]{@{}l@{}}0.5800\\ $\pm$0.09\end{tabular} & \begin{tabular}[c]{@{}l@{}}0.8333\\ $\pm$0.07\end{tabular} & \begin{tabular}[c]{@{}l@{}}0.8400\\ $\pm$0.10\end{tabular} & \begin{tabular}[c]{@{}l@{}}0.7867\\ $\pm$0.13\end{tabular} \\ \hline
arcene                        & \begin{tabular}[c]{@{}l@{}}0.4128\\ $\pm$0.36\end{tabular} & \begin{tabular}[c]{@{}l@{}}0.4880\\ $\pm$0.34\end{tabular} & \begin{tabular}[c]{@{}l@{}}0.4104\\ $\pm$0.31\end{tabular} & \begin{tabular}[c]{@{}l@{}}0.3929\\ $\pm$0.24\end{tabular} & \begin{tabular}[c]{@{}l@{}}0.4562\\ $\pm$0.36\end{tabular} & \begin{tabular}[c]{@{}l@{}}0.5204\\ $\pm$0.23\end{tabular} & \begin{tabular}[c]{@{}l@{}}0.5356\\ $\pm$0.29\end{tabular} & \begin{tabular}[c]{@{}l@{}}0.3806\\ $\pm$0.20\end{tabular} \\ \hline
dorothea                      & \begin{tabular}[c]{@{}l@{}}0.4809\\ $\pm$0.21\end{tabular} & \begin{tabular}[c]{@{}l@{}}0.5094\\ $\pm$0.22\end{tabular} & -                                                          & -                                                          & \begin{tabular}[c]{@{}l@{}}0.6092\\ $\pm$0.16\end{tabular} & \begin{tabular}[c]{@{}l@{}}0.6466\\ $\pm$0.15\end{tabular} & \begin{tabular}[c]{@{}l@{}}0.6721\\ $\pm$0.13\end{tabular} & \begin{tabular}[c]{@{}l@{}}0.6279\\ $\pm$0.14\end{tabular} \\ \hline
leukemia                      & \begin{tabular}[c]{@{}l@{}}0.9696\\ $\pm$0.10\end{tabular} & -                                                          & -                                                          & -                                                          & \begin{tabular}[c]{@{}l@{}}0.8907\\ $\pm$0.19\end{tabular} & \begin{tabular}[c]{@{}l@{}}0.9417\\ $\pm$0.18\end{tabular} & \begin{tabular}[c]{@{}l@{}}0.9416\\ $\pm$0.12\end{tabular} & \begin{tabular}[c]{@{}l@{}}0.9720\\ $\pm$0.09\end{tabular} \\ \hline
breast-cancer                 & \begin{tabular}[c]{@{}l@{}}0.5754\\ $\pm$0.12\end{tabular} & \begin{tabular}[c]{@{}l@{}}0.4854\\ $\pm$0.28\end{tabular} & \begin{tabular}[c]{@{}l@{}}0.5869\\ $\pm$0.12\end{tabular} & \begin{tabular}[c]{@{}l@{}}0.4649\\ $\pm$0.28\end{tabular} & \begin{tabular}[c]{@{}l@{}}0.5440\\ $\pm$0.18\end{tabular} & \begin{tabular}[c]{@{}l@{}}0.6985\\ $\pm$0.07\end{tabular} & \begin{tabular}[c]{@{}l@{}}0.6993\\ $\pm$0.09\end{tabular} & \begin{tabular}[c]{@{}l@{}}0.6832\\ $\pm$0.12\end{tabular} \\ \hline
\end{tabular}
\label{tb6_34}
\end{table}

 \begin{table}
\centering
\caption{Kappa statistic using KNN}
\scriptsize
\begin{tabular}{|l|l|l|l|l|l|l|l|l|}
\hline
Dataset       & \multicolumn{1}{c|}{MMPC}                                  & \multicolumn{1}{c|}{HITON-PC}                              & \multicolumn{1}{c|}{MMMB}                                  & \multicolumn{1}{c|}{HITON-MB}                              & \multicolumn{1}{c|}{IAMB}                                  & \multicolumn{1}{c|}{mRMR}                                  & \multicolumn{1}{c|}{CMIM}                                  & \multicolumn{1}{c|}{JMI}                                   \\ \hline
prostate      & \begin{tabular}[c]{@{}l@{}}0.7814\\ $\pm$0.20\end{tabular} & \begin{tabular}[c]{@{}l@{}}0.8000\\ $\pm$0.19\end{tabular} & \begin{tabular}[c]{@{}l@{}}0.7814\\ $\pm$0.28\end{tabular} & \begin{tabular}[c]{@{}l@{}}0.8045\\ $\pm$0.27\end{tabular} & \begin{tabular}[c]{@{}l@{}}0.8800\\ $\pm$0.22\end{tabular} & \begin{tabular}[c]{@{}l@{}}0.9000\\ $\pm$0.17\end{tabular} & \begin{tabular}[c]{@{}l@{}}0.9000\\ $\pm$0.14\end{tabular} & \begin{tabular}[c]{@{}l@{}}0.9000\\ $\pm$0.17\end{tabular} \\ \hline
dexter        & \begin{tabular}[c]{@{}l@{}}0.5800\\ $\pm$0.13\end{tabular} & \begin{tabular}[c]{@{}l@{}}0.5467\\ $\pm$0.16\end{tabular} & \begin{tabular}[c]{@{}l@{}}0.5800\\ $\pm$0.13\end{tabular} & \begin{tabular}[c]{@{}l@{}}0.5533\\ $\pm$0.15\end{tabular} & \begin{tabular}[c]{@{}l@{}}0.4733\\ $\pm$0.16\end{tabular} & \begin{tabular}[c]{@{}l@{}}0.7667\\ $\pm$0.13\end{tabular} & \begin{tabular}[c]{@{}l@{}}0.7667\\ $\pm$0.13\end{tabular} & \begin{tabular}[c]{@{}l@{}}0.7800\\ $\pm$0.10\end{tabular} \\ \hline
arcene        & \begin{tabular}[c]{@{}l@{}}0.2986\\ $\pm$0.36\end{tabular} & \begin{tabular}[c]{@{}l@{}}0.2392\\ $\pm$0.32\end{tabular} & \begin{tabular}[c]{@{}l@{}}0.2814\\ $\pm$0.40\end{tabular} & \begin{tabular}[c]{@{}l@{}}0.2392\\ $\pm$0.32\end{tabular} & \begin{tabular}[c]{@{}l@{}}0.4102\\ $\pm$0.36\end{tabular} & \begin{tabular}[c]{@{}l@{}}0.5061\\ $\pm$0.20\end{tabular} & \begin{tabular}[c]{@{}l@{}}0.6026\\ $\pm$0.18\end{tabular} & \begin{tabular}[c]{@{}l@{}}0.5473\\ $\pm$0.19\end{tabular} \\ \hline
dorothea      & \begin{tabular}[c]{@{}l@{}}0.4104\\ $\pm$0.25\end{tabular} & \begin{tabular}[c]{@{}l@{}}0.5149\\ $\pm$0.20\end{tabular} & -                                                          & -                                                          & \begin{tabular}[c]{@{}l@{}}0.4761\\ $\pm$0.26\end{tabular} & \begin{tabular}[c]{@{}l@{}}0.6466\\ $\pm$0.15\end{tabular} & \begin{tabular}[c]{@{}l@{}}0.5070\\ $\pm$0.14\end{tabular} & \begin{tabular}[c]{@{}l@{}}0.5775\\ $\pm$0.14\end{tabular} \\ \hline
leukemia      & \begin{tabular}[c]{@{}l@{}}0.9284\\ $\pm$0.15\end{tabular} & -                                                          & -                                                          & -                                                          & \begin{tabular}[c]{@{}l@{}}0.8907\\ $\pm$0.19\end{tabular} & \begin{tabular}[c]{@{}l@{}}0.9416\\ $\pm$0.12\end{tabular} & \begin{tabular}[c]{@{}l@{}}0.9416\\ $\pm$0.12\end{tabular} & \begin{tabular}[c]{@{}l@{}}0.9416\\ $\pm$0.12\end{tabular} \\ \hline
breast-cancer & \begin{tabular}[c]{@{}l@{}}0.5125\\ $\pm$0.22\end{tabular} & \begin{tabular}[c]{@{}l@{}}0.4853\\ $\pm$0.15\end{tabular} & \begin{tabular}[c]{@{}l@{}}0.5176\\ $\pm$0.16\end{tabular} & \begin{tabular}[c]{@{}l@{}}0.4512\\ $\pm$0.21\end{tabular} & \begin{tabular}[c]{@{}l@{}}0.5151\\ $\pm$0.20\end{tabular} & \begin{tabular}[c]{@{}l@{}}0.6718\\ $\pm$0.14\end{tabular} & \begin{tabular}[c]{@{}l@{}}0.6603\\ $\pm$0.10\end{tabular} & \begin{tabular}[c]{@{}l@{}}0.6770\\ $\pm$0.15\end{tabular} \\ \hline
\end{tabular}
\label{tb6_35}
\end{table}

\begin{table}
\centering
\caption{Running time (in seconds)}
\scriptsize
\begin{tabular}{|l|l|l|l|l|l|l|l|l|}
\hline
Dataset	&MMPC  &HITON-PC &MMMB &HITON-MB &IAMB &mRMR &CMI &JMI\\\hline

prostate	&2	&2	&41,568	&54,315	&6.37	 &2.4	&0.09	  &4\\\hline
dexter	&4	&3	&31	&19	&54	&16	&1	&29\\\hline
arcene	&3	&3	&16	&15	&20	&1	&0.3	&19\\\hline
dorothea	&59	&705	&-	&-	&594	&4	&4	&59\\\hline
leukemia	&10,033	&-	&-	&-	&5	 &2	&0.3	&3\\\hline
breast-cancer	&9	&11	&45	&43	&43.23	&17	&0.7	&31\\
\hline
\end{tabular}
\label{tb6_36}
\end{table}

\begin{table}
\centering
\caption{Number of selected features}
\scriptsize
\begin{tabular}{|l|l|l|l|l|l|l|l|l|}
\hline
Dataset	&MMPC  &HITON-PC &MMMB &HITON-MB &IAMB &mRMR &CMI &JMI\\\hline

prostate	&9	&8	&175	&98	&2	&5/20	            &15/5	     &20/20     \\\hline
dexter	&8	&8	&11	&10	&4	&25/20	&25/20	&25/25\\\hline
arcene	&4	&4	&5	&6	&3	&5/10	          &15/20	&35/35\\\hline
dorothea	&24	&28	&-	&-	&6	&15/15	&40/15	&10/10\\\hline
leukemia	&1014	 &-	&-	&-	&1	&10/20	&20/10	&10/10\\\hline
breast-cancer	&8	&6	&10	&7	&4	&40/35	&40/25	&35/40\\
\hline
\end{tabular}
\label{tb6_37}
\end{table}

\subsubsection{Dataset with multiple classes}\label{sec8-23}

In this section, we will evaluate the eight feature selection methods using the seven datasets with multiple classes. Table~\ref{tb6_31}  provides a summary of the datasets.
As for mRMR, CMIM, and JMI,  we set $\psi$ to the top 5, 10, 15, 20, and 25, respectively, then report the results about the feature subset with the highest prediction accuracy.

From Tables~\ref{tb6_42} and~\ref{tb6_43}, we can see that given a dataset with a small number of features and a large number of data instances, even if the dataset with a large number of classes, MMPC, HITON-PC, MMMB, and HITOM-MB have almost the same prediction accuracy as three non-causal feature selection, and even better than them on some datasets, such as  the landsat dataset with six classes.
Meanwhile, Tables~\ref{tb6_44} and~\ref{tb6_45} shows that MMPC, HITON-PC, MMMB, HITOM-MB, and IAMB achieve better Kappa statitic on the connect-4, splice, waveform, and landsat datasets, using both KNN and NBC.

However, given a dataset with a very small number of data instances and a larger number of classes, MMPC, HITON-PC, MMMB, HITOM-MB, and IAMB fail to select any features due to data inefficiency, while  mRMR, CMIM, and JMI seem to work well, especially CMIM.
Tables~\ref{tb6_46} and~~\ref{tb6_47} report the number of selected features and running time of each algorithm. We can see that as expected, mRMR, CMIM, and JMI are faster than  MMPC, HITON-PC, MMMB, HITOM-MB, and IAMB.

Figures~\ref{fig6_41} and~\ref{fig6_42} report the predication accuracy of the causal feature selection methods (with the highest prediction accuracy) in comparison with the three non-causal feature selection methods, mRMR, CMIM and JMI when the number of features selected by the three methods are varied. 
From~\ref{fig6_41} and~\ref{fig6_42}, we can see that, either with KNN or NBC, given a dataset with multiple classes, if the dataset has a large number of data samples, causal feature selection can work well, and they perform better than non-causal feature selection on most of the datasets. However, if the dataset has not enough data samples, causal feature selection fails, while non-causal feature selection can work well on the dataset. 

\begin{table}
\centering
\caption{Dataset with multiple classes}
\scriptsize
\begin{tabular}{|l|l|l|l|}
\hline
Dataset	&Number of features  &Number of instances	&Number of classes\\\hline

connect- 4	  &42	&67,557	&3\\\hline
splice	&60	&3,175	                &3\\\hline
waveform	&40	&5,000  &3\\\hline
landsat 	&36	&6,435	  &6\\\hline
lung	&325	&73	&7\\\hline
lymph	&4,026  &96	&9\\\hline
NCI9	&9,712	  &60	&9\\
\hline
\end{tabular}
\label{tb6_41}
\end{table}

 \begin{table}
\centering
\caption{Prediction accuracy using NBC}
\scriptsize
\begin{tabular}{|l|l|l|l|l|l|l|l|l|}
\hline
Dataset	&MMPC  &HITON-PC &MMMB &HITON-MB &IAMB &mRMR &CMI &JMI\\\hline

connect-4    &\begin{tabular}[c]{@{}l@{}}0.7225\\$\pm$0.00\end{tabular}    &\begin{tabular}[c]{@{}l@{}}0.7225\\$\pm$0.00\end{tabular}    &\begin{tabular}[c]{@{}l@{}}0.7116\\$\pm$ 0.00\end{tabular}  &\begin{tabular}[c]{@{}l@{}}0.7116\\$\pm$0.00\end{tabular}   &\begin{tabular}[c]{@{}l@{}}0.6829\\$\pm$0.01\end{tabular}   &\begin{tabular}[c]{@{}l@{}}0.7059\\$\pm$0.00\end{tabular}    &\begin{tabular}[c]{@{}l@{}}\textbf{0.7233}\\$\pm$0.00\end{tabular}    &\begin{tabular}[c]{@{}l@{}}0.7208\\$\pm$0.00\end{tabular}
\\\hline

splice            &\begin{tabular}[c]{@{}l@{}}0.9619\\$\pm$0.01\end{tabular}   &\begin{tabular}[c]{@{}l@{}}0.9619\\$\pm$0.01\end{tabular}   &\begin{tabular}[c]{@{}l@{}}0.9597\\$\pm$0.01\end{tabular}   &\begin{tabular}[c]{@{}l@{}}0.9594\\$\pm$0.01\end{tabular}    &\begin{tabular}[c]{@{}l@{}}0.8003\\$\pm$0.02\end{tabular}   &\begin{tabular}[c]{@{}l@{}}\textbf{0.9628}\\$\pm$0.01\end{tabular}   &\begin{tabular}[c]{@{}l@{}}0.9613\\$\pm$0.01\end{tabular}    &\begin{tabular}[c]{@{}l@{}}0.9631\\$\pm$0.01\end{tabular}
\\\hline

waveform    &\begin{tabular}[c]{@{}l@{}}0.7910\\$\pm$0.02\end{tabular}   &\begin{tabular}[c]{@{}l@{}}0.7910\\$\pm$0.02\end{tabular}    &\begin{tabular}[c]{@{}l@{}}0.7910\\$\pm$0.02\end{tabular}    &\begin{tabular}[c]{@{}l@{}}0.7910\\$\pm$0.02\end{tabular}    &\begin{tabular}[c]{@{}l@{}}0.7170\\$\pm$0.02\end{tabular}  &\begin{tabular}[c]{@{}l@{}}\textbf{0.8008}\\$\pm$0.02\end{tabular}   &\begin{tabular}[c]{@{}l@{}}0.8004\\$\pm$0.02\end{tabular}    &\begin{tabular}[c]{@{}l@{}}0.7934\\$\pm$0.02\end{tabular}
\\\hline

landsat        &\begin{tabular}[c]{@{}l@{}}\textbf{0.7953}\\$\pm$0.01\end{tabular}   &\begin{tabular}[c]{@{}l@{}}\textbf{0.7953}\\$\pm$0.01\end{tabular}    &\begin{tabular}[c]{@{}l@{}}\textbf{0.7953}\\$\pm$0.01\end{tabular}   &\begin{tabular}[c]{@{}l@{}}\textbf{0.7953}\\$\pm$0.01\end{tabular}    &\begin{tabular}[c]{@{}l@{}}0.7829\\$\pm$0.01\end{tabular}   &\begin{tabular}[c]{@{}l@{}}0.7841\\$\pm$0.01\end{tabular}    &\begin{tabular}[c]{@{}l@{}}0.7885\\$\pm$0.02\end{tabular}   &\begin{tabular}[c]{@{}l@{}}0.7918\\$\pm$0.02\end{tabular}
\\\hline


lung	                      &-                   & -	            &-	    &-	       & -	             &\begin{tabular}[c]{@{}l@{}}0.8363\\$\pm$0.17\end{tabular}	 &\begin{tabular}[c]{@{}l@{}}\textbf{0.8488}\\$\pm$0.15\end{tabular}	&\begin{tabular}[c]{@{}l@{}}\textbf{0.8488}\\$\pm$0.11\end{tabular}
\\\hline

lymph	                    & -	          & -                   &  -	  & -	       & -	               &\begin{tabular}[c]{@{}l@{}}0.8466\\$\pm$0.12	\end{tabular}  &\begin{tabular}[c]{@{}l@{}}\textbf{0.9094}\\$\pm$0.12\end{tabular}	&\begin{tabular}[c]{@{}l@{}}0.8488\\$\pm$0.12\end{tabular}
\\\hline

NCI9	                    & -	           &-	           & -	   &-	       & -               & \begin{tabular}[c]{@{}l@{}}\textbf{0.6708}\\$\pm$0.20\end{tabular}	  &\begin{tabular}[c]{@{}l@{}}0.6494\\$\pm$0.30\end{tabular}	&\begin{tabular}[c]{@{}l@{}}0.5940\\$\pm$0.19\end{tabular}
\\\hline
\end{tabular}
\label{tb6_42}
\end{table}

\begin{table}[t]
\centering
\caption{Prediction accuracy using KNN}
\scriptsize
\begin{tabular}{|l|l|l|l|l|l|l|l|l|}
\hline
Dataset	&MMPC  &HITON-PC &MMMB &HITON-MB &IAMB &mRMR &CMI &JMI\\\hline

connect-4         &\begin{tabular}[c]{@{}l@{}}0.6673\\$\pm$0.01\end{tabular}    &\begin{tabular}[c]{@{}l@{}}0.6673\\$\pm$0.01\end{tabular}    &\begin{tabular}[c]{@{}l@{}}0.6620\\$\pm$0.01\end{tabular}   &\begin{tabular}[c]{@{}l@{}}0.6592\\$\pm$0.01\end{tabular}    &\begin{tabular}[c]{@{}l@{}}0.6397\\$\pm$0.01\end{tabular}    &\begin{tabular}[c]{@{}l@{}}0.6601\\$\pm$0.01\end{tabular}    &\begin{tabular}[c]{@{}l@{}}\textbf{0.7217}\\$\pm$0.00\end{tabular} &\begin{tabular}[c]{@{}l@{}}0.7139\\$\pm$0.00\end{tabular}
\\\hline
splice                &\begin{tabular}[c]{@{}l@{}}0.7338\\$\pm$0.02\end{tabular}    &\begin{tabular}[c]{@{}l@{}}0.7338\\$\pm$0.02\end{tabular}     &\begin{tabular}[c]{@{}l@{}}0.7181\\$\pm$0.02\end{tabular}     &\begin{tabular}[c]{@{}l@{}}0.7190\\$\pm$0.02\end{tabular}     &\begin{tabular}[c]{@{}l@{}}0.7770\\$\pm$0.01\end{tabular}    &\begin{tabular}[c]{@{}l@{}}\textbf{0.8771}\\$\pm$0.02\end{tabular}     &\begin{tabular}[c]{@{}l@{}}\textbf{0.8771}\\$\pm$0.02\end{tabular}     &\begin{tabular}[c]{@{}l@{}}0.8771\\$\pm$0.02\end{tabular}
\\\hline
waveform        &\begin{tabular}[c]{@{}l@{}}0.8022\\$\pm$0.02\end{tabular}     &\begin{tabular}[c]{@{}l@{}}0.8022\\$\pm$0.02\end{tabular}     &\begin{tabular}[c]{@{}l@{}}0.8022\\$\pm$0.02\end{tabular}     &\begin{tabular}[c]{@{}l@{}}0.8022\\$\pm$0.02\end{tabular}     &\begin{tabular}[c]{@{}l@{}}0.7028\\$\pm$0.02\end{tabular}     &\begin{tabular}[c]{@{}l@{}}\textbf{0.8032}\\$\pm$0.02\end{tabular}    &\begin{tabular}[c]{@{}l@{}}\textbf{0.8032}\\$\pm$0.02\end{tabular}     &\begin{tabular}[c]{@{}l@{}}\textbf{0.8032}\\$\pm$0.02\end{tabular}
\\\hline
landsat            &\begin{tabular}[c]{@{}l@{}}\textbf{0.8696}\\$\pm$0.01\end{tabular}    &\begin{tabular}[c]{@{}l@{}}\textbf{0.8696}\\$\pm$0.01\end{tabular}    &\begin{tabular}[c]{@{}l@{}}\textbf{0.8696}\\$\pm$0.01\end{tabular}    &\begin{tabular}[c]{@{}l@{}}\textbf{0.8696}\\$\pm$0.01\end{tabular}    &\begin{tabular}[c]{@{}l@{}}0.7433\\$\pm$0.01\end{tabular}    &\begin{tabular}[c]{@{}l@{}}0.8605\\$\pm$0.01\end{tabular}    &\begin{tabular}[c]{@{}l@{}}0.8589\\$\pm$0.01\end{tabular}    &\begin{tabular}[c]{@{}l@{}}0.8614\\$\pm$0.01\end{tabular}
\\\hline
lung	                 & -	        &  -	             &-	     &-	         & -	   &\begin{tabular}[c]{@{}l@{}}0.8524\\$\pm$0.12\end{tabular}	 &\begin{tabular}[c]{@{}l@{}}\textbf{0.8649}\\$\pm$0.08\end{tabular}	     &\begin{tabular}[c]{@{}l@{}}0.8464\\$\pm$0.11\end{tabular}
\\\hline
lymph	                  &-	          &-	             &-	    &-	         &-	   &\begin{tabular}[c]{@{}l@{}}\textbf{0.9269}\\$\pm$0.07\end{tabular}	&\begin{tabular}[c]{@{}l@{}}0.9176\\$\pm$0.07\end{tabular}	     &\begin{tabular}[c]{@{}l@{}}0.9053\\$\pm$0.06\end{tabular}
\\\hline
NCI9	                  &-	          &-	             &-	    &-	          &-	   &\begin{tabular}[c]{@{}l@{}}\textbf{0.5875}\\$\pm$0.22\end{tabular}	 &\begin{tabular}[c]{@{}l@{}}0.5494\\$\pm$0.18\end{tabular}	    &\begin{tabular}[c]{@{}l@{}}0.5684\\$\pm$0.15\end{tabular}
\\\hline
\end{tabular}
\label{tb6_43}
\end{table}

\begin{figure}
\centering
\includegraphics[height=1.5in]{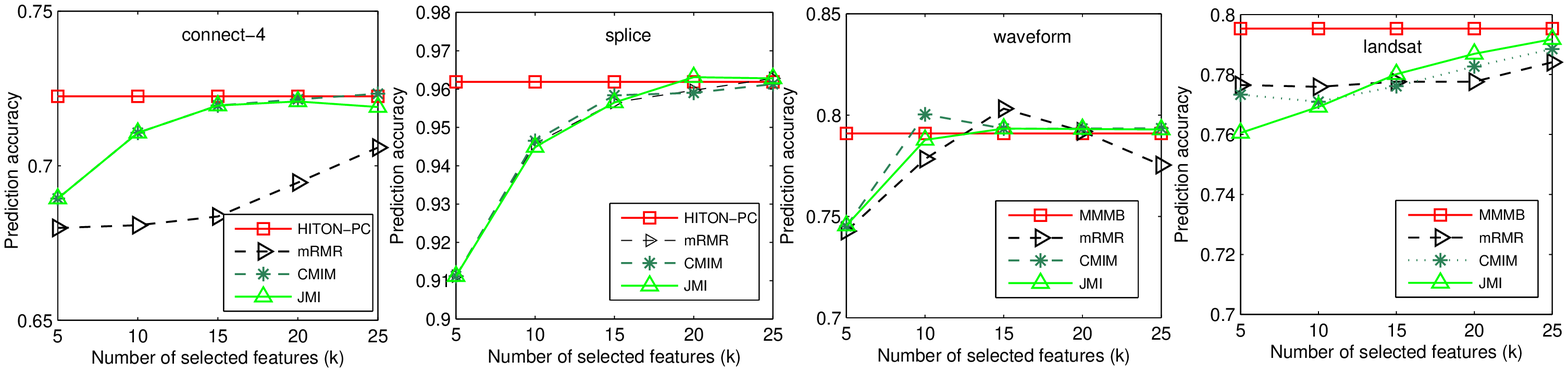}
\caption{AUC with different values of $\psi$ using NBC}
\label{fig6_41}
\end{figure}
\begin{figure}
\centering
\includegraphics[height=1.7in,width=5.85in]{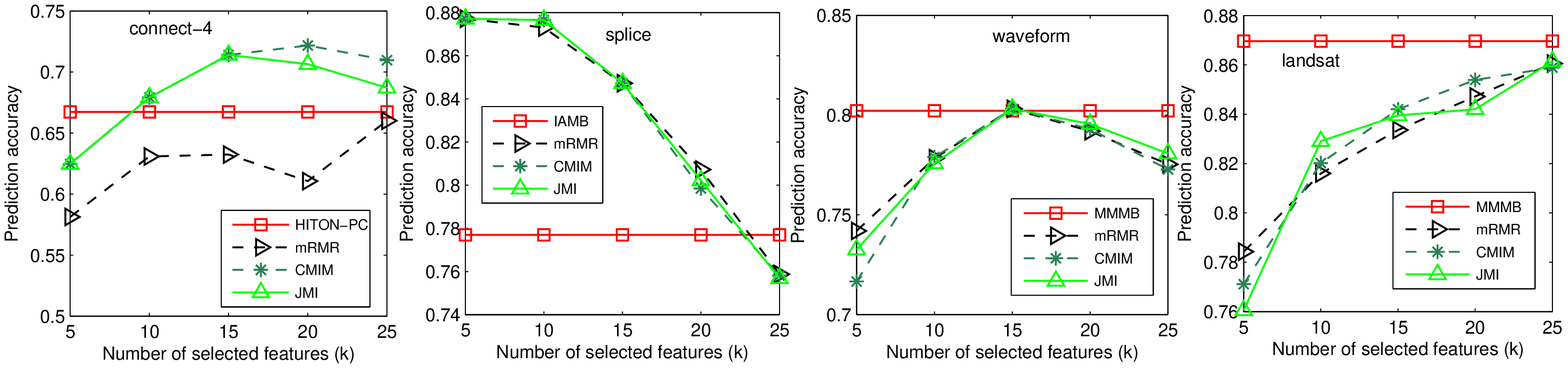}
\caption{AUC with different values of $\psi$ using KNN}
\label{fig6_42}
\end{figure}

\begin{table}
\centering
\caption{Kappa statistic using NBC}
\scriptsize
\begin{tabular}{|l|l|l|l|l|l|l|l|l|}
\hline
\multicolumn{1}{|c|}{Dataset} & \multicolumn{1}{c|}{MMPC}                                  & \multicolumn{1}{c|}{HITON-PC}                              & \multicolumn{1}{c|}{MMMB}                                  & \multicolumn{1}{c|}{HITON-MB}                              & \multicolumn{1}{c|}{IAMB}                                  & \multicolumn{1}{c|}{mRMR}                                  & \multicolumn{1}{c|}{CMIM}                                  & \multicolumn{1}{c|}{JMI}                                   \\ \hline
connect-4                     & \begin{tabular}[c]{@{}l@{}}\textbf{0.3299}\\ $\pm$0.01\end{tabular} & \begin{tabular}[c]{@{}l@{}}\textbf{0.3299}\\ $\pm$0.01\end{tabular} & \begin{tabular}[c]{@{}l@{}}0.3125\\ $\pm$0.02\end{tabular} & \begin{tabular}[c]{@{}l@{}}0.3125\\ $\pm$0.02\end{tabular} & \begin{tabular}[c]{@{}l@{}}0.1804\\ $\pm$0.02\end{tabular} & \begin{tabular}[c]{@{}l@{}}0.2701\\ $\pm$0.01\end{tabular} & \begin{tabular}[c]{@{}l@{}}0.3176\\ $\pm$0.00\end{tabular} & \begin{tabular}[c]{@{}l@{}}0.3082\\ $\pm$0.01\end{tabular} \\ \hline
splice                        & \begin{tabular}[c]{@{}l@{}}0.9382\\ $\pm$0.01\end{tabular} & \begin{tabular}[c]{@{}l@{}}0.9382\\ $\pm$0.01\end{tabular} & \begin{tabular}[c]{@{}l@{}}0.9346\\ $\pm$0.02\end{tabular} & \begin{tabular}[c]{@{}l@{}}0.9341\\ $\pm$0.02\end{tabular} & \begin{tabular}[c]{@{}l@{}}0.6841\\ $\pm$0.03\end{tabular} & \begin{tabular}[c]{@{}l@{}}0.9398\\ $\pm$0.01\end{tabular} & \begin{tabular}[c]{@{}l@{}}0.9372\\ $\pm$0.01\end{tabular} & \begin{tabular}[c]{@{}l@{}}\textbf{0.9403}\\ $\pm$0.01\end{tabular} \\ \hline

waveform  & \begin{tabular}[c]{@{}l@{}}0.6870\\ $\pm$0.02\end{tabular} & \begin{tabular}[c]{@{}l@{}}0.6870\\ $\pm$0.02\end{tabular} & \begin{tabular}[c]{@{}l@{}}0.6870\\ $\pm$0.02\end{tabular} & \begin{tabular}[c]{@{}l@{}}0.6870\\ $\pm$0.02\end{tabular} & \begin{tabular}[c]{@{}l@{}}0.5758\\ $\pm$0.03\end{tabular} & \begin{tabular}[c]{@{}l@{}}0.7015\\ $\pm$0.02\end{tabular} & \begin{tabular}[c]{@{}l@{}}\textbf{0.7009}\\ $\pm$0.02\end{tabular} & \begin{tabular}[c]{@{}l@{}}0.6906\\ $\pm$0.02\end{tabular} \\ \hline

landsat                       & \begin{tabular}[c]{@{}l@{}}\textbf{0.7494}\\ $\pm$0.01\end{tabular} & \begin{tabular}[c]{@{}l@{}}\textbf{0.7494}\\ $\pm$0.01\end{tabular} & \begin{tabular}[c]{@{}l@{}}\textbf{0.7494}\\ $\pm$0.01\end{tabular} & \begin{tabular}[c]{@{}l@{}}\textbf{0.7494}\\ $\pm$0.01\end{tabular} & \begin{tabular}[c]{@{}l@{}}0.7298\\ $\pm$0.02\end{tabular} & \begin{tabular}[c]{@{}l@{}}0.7349\\ $\pm$0.02\end{tabular} & \begin{tabular}[c]{@{}l@{}}0.7402\\ $\pm$0.02\end{tabular} & \begin{tabular}[c]{@{}l@{}}0.7444\\ $\pm$0.02\end{tabular} \\ \hline
lung                          & -                                                          & -                                                          & -                                                          & -                                                          & -                                                          & \begin{tabular}[c]{@{}l@{}}0.7950\\ $\pm$0.21\end{tabular} & \begin{tabular}[c]{@{}l@{}}0.8082\\ $\pm$0.19\end{tabular} & \begin{tabular}[c]{@{}l@{}}\textbf{0.8086}\\ $\pm$0.14\end{tabular} \\ \hline
lymph                         & -                                                          & -                                                          & -                                                          & -                                                          & -                                                          & \begin{tabular}[c]{@{}l@{}}0.7885\\ $\pm$0.16\end{tabular} & \begin{tabular}[c]{@{}l@{}}\textbf{0.8790}\\ $\pm$0.15\end{tabular} & \begin{tabular}[c]{@{}l@{}}0.7969\\ $\pm$0.16\end{tabular} \\ \hline
NCI9                          & -                                                          & -                                                          & -                                                          & -                                                          & -   & \begin{tabular}[c]{@{}l@{}}\textbf{0.6223}\\ $\pm$0.20\end{tabular} & \begin{tabular}[c]{@{}l@{}}0.6058\\ $\pm$0.28\end{tabular} & \begin{tabular}[c]{@{}l@{}}0.5337\\ $\pm$0.22\end{tabular} \\ \hline
\end{tabular}
\label{tb6_44}
\end{table}

\begin{table}
\centering
\caption{Kappa statistic uisng KNN}
\scriptsize
\begin{tabular}{|l|l|l|l|l|l|l|l|l|}
\hline
Dataset   & \multicolumn{1}{c|}{MMPC}                                  & \multicolumn{1}{c|}{HITON-PC}                              & \multicolumn{1}{c|}{MMMB}                                  & \multicolumn{1}{c|}{HITON-MB}                              & \multicolumn{1}{c|}{IAMB}                                  & \multicolumn{1}{c|}{mRMR}                                  & \multicolumn{1}{c|}{CMIM}                                  & \multicolumn{1}{c|}{JMI}                                   \\ \hline
connect-4 & \begin{tabular}[c]{@{}l@{}}0.3058\\ $\pm$0.01\end{tabular} & \begin{tabular}[c]{@{}l@{}}0.3058\\ $\pm$0.01\end{tabular} & \begin{tabular}[c]{@{}l@{}}0.2863\\ $\pm$0.01\end{tabular} & \begin{tabular}[c]{@{}l@{}}0.2863\\ $\pm$0.01\end{tabular} & \begin{tabular}[c]{@{}l@{}}0.2751\\ $\pm$0.04\end{tabular} & \begin{tabular}[c]{@{}l@{}}0.2560\\ $\pm$0.01\end{tabular} & \begin{tabular}[c]{@{}l@{}}\textbf{0.4052}\\ $\pm$0.01\end{tabular} & \begin{tabular}[c]{@{}l@{}}0.3943\\ $\pm$0.01\end{tabular} \\ \hline

splice    & \begin{tabular}[c]{@{}l@{}}0.5954\\ $\pm$0.02\end{tabular} & \begin{tabular}[c]{@{}l@{}}0.5954\\ $\pm$0.02\end{tabular} & \begin{tabular}[c]{@{}l@{}}0.5674\\ $\pm$0.04\end{tabular} & \begin{tabular}[c]{@{}l@{}}0.5703\\ $\pm$0.04\end{tabular} & \begin{tabular}[c]{@{}l@{}}0.6456\\ $\pm$0.02\end{tabular} & \begin{tabular}[c]{@{}l@{}}\textbf{0.8027}\\ $\pm$0.03\end{tabular} & \begin{tabular}[c]{@{}l@{}}\textbf{0.8027}\\ $\pm$0.03\end{tabular} & \begin{tabular}[c]{@{}l@{}}\textbf{0.8027}\\ $\pm$0.03\end{tabular} \\ \hline

waveform  & \begin{tabular}[c]{@{}l@{}}0.7033\\ $\pm$0.03\end{tabular} & \begin{tabular}[c]{@{}l@{}}0.7033\\ $\pm$0.03\end{tabular} & \begin{tabular}[c]{@{}l@{}}0.7033\\ $\pm$0.03\end{tabular} & \begin{tabular}[c]{@{}l@{}}0.7033\\ $\pm$0.03\end{tabular} & \begin{tabular}[c]{@{}l@{}}0.5541\\ $\pm$0.03\end{tabular} & \begin{tabular}[c]{@{}l@{}}\textbf{0.7048}\\ $\pm$0.02\end{tabular} & \begin{tabular}[c]{@{}l@{}}\textbf{0.7048}\\ $\pm$0.02\end{tabular} & \begin{tabular}[c]{@{}l@{}}\textbf{0.7048}\\ $\pm$0.02\end{tabular} \\ \hline

landsat   & \begin{tabular}[c]{@{}l@{}}\textbf{0.8387}\\ $\pm$0.01\end{tabular} & \begin{tabular}[c]{@{}l@{}}\textbf{0.8387}\\ $\pm$0.01\end{tabular} & \begin{tabular}[c]{@{}l@{}}\textbf{0.8387}\\ $\pm$0.01\end{tabular} & \begin{tabular}[c]{@{}l@{}}\textbf{0.8387}\\ $\pm$0.01\end{tabular} & \begin{tabular}[c]{@{}l@{}}0.6772\\ $\pm$0.03\end{tabular} & \begin{tabular}[c]{@{}l@{}}0.8270\\ $\pm$0.01\end{tabular} & \begin{tabular}[c]{@{}l@{}}0.8250\\ $\pm$0.01\end{tabular} & \begin{tabular}[c]{@{}l@{}}0.8282\\ $\pm$0.02\end{tabular} \\ \hline

lung      & -                                                          & -                                                          & -                                                          & -                                                          & -  & \begin{tabular}[c]{@{}l@{}}0.8123\\ $\pm$0.15\end{tabular} & \begin{tabular}[c]{@{}l@{}}\textbf{0.8273}\\ $\pm$0.11\end{tabular} & \begin{tabular}[c]{@{}l@{}}0.8017\\ $\pm$0.14\end{tabular} \\ \hline

lymph     & -                                                          & -                                                          & -                                                          & -                                                          & -    & \begin{tabular}[c]{@{}l@{}}\textbf{0.8972}\\ $\pm$0.10\end{tabular} & \begin{tabular}[c]{@{}l@{}}0.8853\\ $\pm$0.10\end{tabular} & \begin{tabular}[c]{@{}l@{}}0.8662\\ $\pm$0.09\end{tabular} \\ \hline

NCI9      & -                                                          & -                                                          & -                                                          & -                                                          & -   & \begin{tabular}[c]{@{}l@{}}0.5190\\ $\pm$0.25\end{tabular} & \begin{tabular}[c]{@{}l@{}}0.4856\\ $\pm$0.18\end{tabular} & \begin{tabular}[c]{@{}l@{}}\textbf{0.5087}\\ $\pm$0.14\end{tabular} \\ \hline

\end{tabular}
\label{tb6_45}
\end{table}

\begin{table}
\centering
\caption{Number of selected features}
\scriptsize
\begin{tabular}{|l|l|l|l|l|l|l|l|l|}
\hline
Dataset	&MMPC  &HITON-PC &MMMB &HITON-MB &IAMB &mRMR &CMI &JMI\\\hline

connect-4	&37	  &37	&42	      &42	         &7	 &25/25	  &25/20	  &20/15
\\\hline
splice	      &28	  &28	&50	      &49	        &3	       &25/5	  &25/5	  &20/5
\\\hline
waveform     &17	  &17	&17	      &17	        &3	       &10/15	  &10/15	  &15/15
\\\hline
landsat 	      &36	  &36	&36	      &36	        &3	       &25/25	  &25/25       &25/25
\\\hline
lung	          &-	  &-	      &-	       &-	       &- 	      &40/35        &40/30	&40/35
\\\hline
lymph	     &-	  &-	      &-	       &-	       &- 	      &35/40	       &25/35	&20/40
\\\hline
NCI9	    &-	  &-	      &-	       &-	       &- 	     &20/30	      &10/25     &40/40
\\\hline
\end{tabular}
\label{tb6_46}
\end{table}

\begin{table}
\centering
\caption{Running time (in seconds)}
\scriptsize
\begin{tabular}{|l|l|l|l|l|l|l|l|l|}
\hline
Dataset	&MMPC  &HITON-PC &MMMB &HITON-MB &IAMB &mRMR &CMI &JMI\\\hline

connect-4	&2679	&2949	&23564	   &23795	&7.85	     &0.76	&1.4	&2.33
\\\hline
splice	       &12	&14	      &34	         &34	      &0.30	     &0.08	&0.08	&0.25
\\\hline
waveform	&4	      &4	     &16	        &16	     &0.2917	     &0.05	&0.03	&0.17
\\\hline
landsat 	      &32	      &43	     &798	       &1019	     &0.2727	     &0.08	&0.14	&0.26
\\\hline
lung          &-	  &-	      &-	       &-	       &-	    &0.5      &0.06	  &0.7
\\\hline
lymph	   &-	  &-	      &-	       &-	       &-	    &6	   &00.3	  &10
\\\hline
NCI9	   &-	  &-	      &-	       &-	       &-	    &10	   &0.4	  &22
\\\hline
\end{tabular}
\label{tb6_47}
\end{table}

\subsubsection{Dataset with imbalanced classes}\label{sec8-24}

In this section, we use six class-imbalanced datasets in Table~\ref{tb6_51} to examine the performance of causal and non-causal feature selection methods. 
For mRMR, CMIM, and JMI, we set $\psi$ to the top 5, 10, 15, $\cdots$, 25, and 30, then select the feature subset with the highest prediction accuracy as the reporting result.

From Tables~\ref{tb6_52} and~\ref{tb6_53}, we can see that  all the eight algorithms get good prediction accuracy, but each of them achieves a very low AUC, as seen from Tables~\ref{tb6_54} and~\ref{tb6_55}. In addition, on both prediction accuracy and AUC, the five causal feature selection methods and the three non-causal feature selection algorithms achieve almost the same performance.
At the same time,  as see from  Tables~\ref{tb6_58} and~\ref{tb6_59},  all the eight algorithms do not achieve better Kappa statistic  regardless of using both KNN and NBC.

Tables~\ref{tb6_56} and~\ref{tb6_57} report the number of selected features and running time of each algorithm. We can see that MMMB and HITON-MB are the slowest algorithm among the nine algorithms under comparison.
Thus,  we can conclude that both the causal feature selection methods and  non-causal feature selection algorithms cannot deal with class-imbalanced datasets well.

From Figures~\ref{fig6_51} and~\ref{fig6_52}, when the number of selected features by mRMR, CMIM and JMI are varied, we can see that, either with KNN or NBC, given a class-imbalanced dataset,  both non-causal feature selection and causal feature selection are not able to deal with the dataset well.

\begin{table}
\centering
\caption{Class-imbalanced datasets}
\scriptsize
\begin{tabular}{|l|l|l|l|l|}
\hline
Dataset	&Number of features  &Number of instances	&Number of classes &ratio\\\hline

hiva	&1,617	&4,229	   &2	&3.52\%\\\hline
ohsumed	&14,373	&5,000	   &2	&5.56\%\\\hline
acpj	&28,228	&15,779	&2	&1.3\%\\\hline
sido0	&4,932	           &12,678	&2	&3.54\%\\\hline
thrombin	&13,9351	&2,543	  &2	&7.55\%\\\hline
infant	&86	&5,339	  &2	&6.31\%\\
\hline
\end{tabular}
\label{tb6_51}
\end{table}

 \begin{table}
\centering
\caption{Prediction accuracy using NBC}
\scriptsize
\begin{tabular}{|l|l|l|l|l|l|l|l|l|}
\hline
Dataset	&MMPC  &HITON-PC &MMMB &HITON-MB &IAMB &mRMR &CMI &JMI\\\hline

hiva                &\begin{tabular}[c]{@{}l@{}}0.9660\\$\pm$0.01\end{tabular}    &\begin{tabular}[c]{@{}l@{}}\textbf{0.9674}\\$\pm$0.01\end{tabular}    &\begin{tabular}[c]{@{}l@{}}0.9660\\$\pm$0.01\end{tabular}    &\begin{tabular}[c]{@{}l@{}}\textbf{0.9674}\\$\pm$0.01\end{tabular}    &\begin{tabular}[c]{@{}l@{}}0.9657\\$\pm$0.01\end{tabular}    &\begin{tabular}[c]{@{}l@{}}0.9565\\$\pm$0.01\end{tabular}    &\begin{tabular}[c]{@{}l@{}}0.9409\\$\pm$0.01\end{tabular}  &\begin{tabular}[c]{@{}l@{}}0.9470\\$\pm$0.01\end{tabular}
\\\hline

ohsumed        &\begin{tabular}[c]{@{}l@{}}0.9506\\$\pm$0.01\end{tabular}    &\begin{tabular}[c]{@{}l@{}}0.9514\\$\pm$0.01\end{tabular}    &\begin{tabular}[c]{@{}l@{}}0.9486\\$\pm$0.01\end{tabular}    &\begin{tabular}[c]{@{}l@{}}0.9486\\$\pm$0.01\end{tabular}    &\begin{tabular}[c]{@{}l@{}}0.9480\\$\pm$0.00\end{tabular}    &\begin{tabular}[c]{@{}l@{}}\textbf{0.9510}\\$\pm$0.01\end{tabular}    &\begin{tabular}[c]{@{}l@{}}0.9490\\$\pm$0.01\end{tabular}    &\begin{tabular}[c]{@{}l@{}}0.9490\\$\pm$0.01\end{tabular}
\\\hline

acpj               &\begin{tabular}[c]{@{}l@{}}0.9527\\$\pm$0.01\end{tabular}    &\begin{tabular}[c]{@{}l@{}}0.9528\\$\pm$0.01\end{tabular}     & -                          & -                         &\begin{tabular}[c]{@{}l@{}}0.9612\\$\pm$0.00\end{tabular}     &\begin{tabular}[c]{@{}l@{}}\textbf{0.9625}\\$\pm$0.00\end{tabular}    &\begin{tabular}[c]{@{}l@{}}0.9364\\$\pm$0.00\end{tabular}    &\begin{tabular}[c]{@{}l@{}}0.9271\\$\pm$0.00\end{tabular}
\\\hline

sido0             &\begin{tabular}[c]{@{}l@{}}0.9535\\$\pm$0.01\end{tabular}    &\begin{tabular}[c]{@{}l@{}}0.9257\\$\pm$0.02\end{tabular}     &\begin{tabular}[c]{@{}l@{}}0.9391\\$\pm$0.01\end{tabular}    &\begin{tabular}[c]{@{}l@{}}0.9076\\$\pm$0.01\end{tabular}   &\begin{tabular}[c]{@{}l@{}}\textbf{0.9584}\\$\pm$0.01\end{tabular}     &\begin{tabular}[c]{@{}l@{}}0.9144\\$\pm$0.01\end{tabular}    &\begin{tabular}[c]{@{}l@{}}0.9106\\$\pm$0.01\end{tabular}    &\begin{tabular}[c]{@{}l@{}}0.9018\\$\pm$0.01\end{tabular}
\\\hline

thrombin       &\begin{tabular}[c]{@{}l@{}}0.9218\\$\pm$0.02\end{tabular}    &\begin{tabular}[c]{@{}l@{}}0.9029\\$\pm$0.03\end{tabular}    &\begin{tabular}[c]{@{}l@{}}0.8746\\$\pm$0.02\end{tabular}    &\begin{tabular}[c]{@{}l@{}}0.8482\\$\pm$0.02\end{tabular}    &\begin{tabular}[c]{@{}l@{}}0.9265\\$\pm$0.02\end{tabular}    &\begin{tabular}[c]{@{}l@{}}\textbf{0.9426}\\$\pm$0.01\end{tabular}    &\begin{tabular}[c]{@{}l@{}}0.8883\\$\pm$0.02\end{tabular}     &\begin{tabular}[c]{@{}l@{}}0.9080\\$\pm$0.02\end{tabular}
\\\hline

infant           &\begin{tabular}[c]{@{}l@{}}0.9522\\$\pm$0.01\end{tabular}    &\begin{tabular}[c]{@{}l@{}}0.9518\\$\pm$0.01\end{tabular}    &\begin{tabular}[c]{@{}l@{}}0.9507\\$\pm$0.01\end{tabular}    &\begin{tabular}[c]{@{}l@{}}0.9518\\$\pm$0.01\end{tabular}    &\begin{tabular}[c]{@{}l@{}}\textbf{0.9530}\\$\pm$0.01\end{tabular}    &\begin{tabular}[c]{@{}l@{}}0.9417\\$\pm$0.01\end{tabular}    &\begin{tabular}[c]{@{}l@{}}0.9427\\$\pm$0.01\end{tabular}    &\begin{tabular}[c]{@{}l@{}}0.9378\\$\pm$0.01\end{tabular}
\\\hline
\end{tabular}
\label{tb6_52}
\end{table}

\section{Conclusion}\label{sec9}

In this paper, we have proposed a unified view to fill in the gap in the research of the relation between causal and non-causal feature selection methods. With this view, we have analyzed the mechanisms of both types of feature selection methods and have shown that both major approaches to feature selection use different strategies to discover the MB of a class attribute under different  Bayesian network structural assumptions. In theory, the feature sets obtained by causal feature selection methods are closer to the MB of the class attribute than non-causal feature selection methods, while non-causal methods are more computationally efficient and need fewer data samples than causal methods.
With this view, we have provided causal interpretations to the output of non-causal feature selection methods and analyzed the error bounds of causal and non-causal methods.  In addition, we have conducted extensive experiments to validate our findings in the paper. 

From the theoretical and experimental analysis in the paper, we can find that both types of feature selection still face many changes as listed below and we hope this paper can stimulate the interest of researchers in machine learning to develop new methods to address these challenges.
\begin{itemize}

\item Small sample size. Causal feature selection cannot deal with a dataset with high dimensionality and small sample size. Then how can we leverage non-causal feature selection to help causal feature selection to improve the computational performance and accuracy of causal feature selection methods for large dimensional problems and small sample sizes?

\item Imbalanced classes. The majority of existing causal and non-causal feature selection methods cannot deal with datasets with imbalanced classes, which exist in many real-world applications. It is important to develop new feature selection methods to address this problem.

\item Large-sized MBs. A large  MB makes causal feature selection methods suffer from the data-inefficient or time-inefficient problem, 
Thus, it is essential for big data analysts to develop efficient causal feature selection methods for dealing with large MB containing hundreds of features.

\item Selection of proper parameter values. It is a hard problem for non-causal feature selection to determine a suitable value of $\psi$. How do both types of feature selection methods benefit each other to solve the problem?

\item Efficiency. Most local-to-global Bayesian network learning methods employ causal feature selection methods to learn MBs for constructing a causal structure skeleton. Can we leverage non-casual feature selection methods to improve the computational performance of  local-to-global learning methods with theoretical guarantees? 
\end{itemize}

\begin{table}
\centering
\caption{Prediction accuracy using KNN}
\scriptsize
\begin{tabular}{|l|l|l|l|l|l|l|l|l|}
\hline
Dataset	&MMPC  &HITON-PC &MMMB &HITON-MB &IAMB &mRMR &CMI &JMI\\\hline

hiva                   &\begin{tabular}[c]{@{}l@{}}0.9650\\$\pm$0.01\end{tabular}   &\begin{tabular}[c]{@{}l@{}}0.9655\\$\pm$0.01\end{tabular}    &\begin{tabular}[c]{@{}l@{}}0.9652\\$\pm$0.01\end{tabular}    &\begin{tabular}[c]{@{}l@{}}0.9655\\$\pm$0.01\end{tabular}    &\begin{tabular}[c]{@{}l@{}}\textbf{0.9664}\\$\pm$0.00\end{tabular}    &\begin{tabular}[c]{@{}l@{}}0.9643\\$\pm$0.01\end{tabular}    &\begin{tabular}[c]{@{}l@{}}0.9631\\$\pm$0.01\end{tabular}    &\begin{tabular}[c]{@{}l@{}}0.9650\\$\pm$0.00\end{tabular}

\\\hline
ohsumed           &\begin{tabular}[c]{@{}l@{}}0.9450\\$\pm$0.01\end{tabular}    &\begin{tabular}[c]{@{}l@{}}0.9442\\$\pm$0.01\end{tabular}    &\begin{tabular}[c]{@{}l@{}}\textbf{0.9456}\\$\pm$0.00\end{tabular}    &\begin{tabular}[c]{@{}l@{}}0.9452\\$\pm$0.00\end{tabular}    &\begin{tabular}[c]{@{}l@{}}0.9450\\$\pm$0.00\end{tabular}    &\begin{tabular}[c]{@{}l@{}}0.9400\\$\pm$0.01\end{tabular}   &\begin{tabular}[c]{@{}l@{}}0.9388\\$\pm$0.01\end{tabular}    &\begin{tabular}[c]{@{}l@{}}0.9388\\$\pm$0.01\end{tabular}

\\\hline
acpj                   &\begin{tabular}[c]{@{}l@{}}0.9845\\$\pm$0.00\end{tabular}   &\begin{tabular}[c]{@{}l@{}}0.9850\\$\pm$0.00\end{tabular}     &-              &-              &\begin{tabular}[c]{@{}l@{}}0.9839\\$\pm$0.00\end{tabular}    &\begin{tabular}[c]{@{}l@{}}\textbf{0.9867}\\$\pm$0.02\end{tabular}    &\begin{tabular}[c]{@{}l@{}}0.9860\\$\pm$0.02\end{tabular}    &\begin{tabular}[c]{@{}l@{}}0.9856\\$\pm$0.00\end{tabular}

\\\hline
sido0                 &\begin{tabular}[c]{@{}l@{}}0.9683\\$\pm$0.00\end{tabular}    &\begin{tabular}[c]{@{}l@{}}0.9643\\$\pm$0.01\end{tabular}    &\begin{tabular}[c]{@{}l@{}}\textbf{0.9696}\\$\pm$0.00\end{tabular}    &\begin{tabular}[c]{@{}l@{}}0.9682\\$\pm$0.00\end{tabular}    &\begin{tabular}[c]{@{}l@{}}0.9653\\$\pm$0.00\end{tabular}    &\begin{tabular}[c]{@{}l@{}}0.9669\\$\pm$0.00\end{tabular}    &\begin{tabular}[c]{@{}l@{}}0.9677\\$\pm$0.00\end{tabular}    &\begin{tabular}[c]{@{}l@{}}0.9549\\$\pm$0.02\end{tabular}
\\\hline

thrombin           &\begin{tabular}[c]{@{}l@{}}0.9481\\$\pm$0.01\end{tabular}    &\begin{tabular}[c]{@{}l@{}}0.9465\\$\pm$0.01\end{tabular}    &\begin{tabular}[c]{@{}l@{}}0.9459\\$\pm$0.02\end{tabular}   &\begin{tabular}[c]{@{}l@{}}0.9465\\$\pm$0.01\end{tabular}    &\begin{tabular}[c]{@{}l@{}}0.9406\\$\pm$0.01\end{tabular}    &\begin{tabular}[c]{@{}l@{}}\textbf{0.9505}\\$\pm$0.01\end{tabular}    &\begin{tabular}[c]{@{}l@{}}0.9481\\$\pm$0.01\end{tabular}    &\begin{tabular}[c]{@{}l@{}}0.9489\\$\pm$0.01\end{tabular}

\\\hline
infant                &\begin{tabular}[c]{@{}l@{}}0.9533\\$\pm$0.01\end{tabular}    &\begin{tabular}[c]{@{}l@{}}\textbf{0.9548}\\$\pm$0.01\end{tabular}    &\begin{tabular}[c]{@{}l@{}}\textbf{0.9548}\\$\pm$0.01\end{tabular}    &\begin{tabular}[c]{@{}l@{}}0.9530\\$\pm$0.01\end{tabular}    &\begin{tabular}[c]{@{}l@{}}0.9509\\$\pm$0.01\end{tabular}   &\begin{tabular}[c]{@{}l@{}}0.9545\\$\pm$0.00\end{tabular}    &\begin{tabular}[c]{@{}l@{}}0.9532\\$\pm$0.01\end{tabular}  &\begin{tabular}[c]{@{}l@{}}0.9526\\$\pm$0.00\end{tabular}
\\\hline
\end{tabular}
\label{tb6_53}
\end{table}

\begin{table}
\centering
\caption{ AUC using NBC}
\scriptsize
\begin{tabular}{|l|l|l|l|l|l|l|l|l|}
\hline
Dataset	&MMPC  &HITON-PC &MMMB &HITON-MB &IAMB &mRMR &CMI &JMI\\\hline

hiva	            &\begin{tabular}[c]{@{}l@{}}0.5779\\$\pm$0.05\end{tabular}	&\begin{tabular}[c]{@{}l@{}}0.5754\\$\pm$0.05\end{tabular}	&\begin{tabular}[c]{@{}l@{}}0.5843\\$\pm$0.07\end{tabular}	 &\begin{tabular}[c]{@{}l@{}}0.5754\\$\pm$0.05\end{tabular}	&\begin{tabular}[c]{@{}l@{}}0.5617\\$\pm$0.03\end{tabular}	&\begin{tabular}[c]{@{}l@{}}\textbf{0.6797}\\$\pm$0.05\end{tabular}	 &\begin{tabular}[c]{@{}l@{}}0.6523\\$\pm$0.05\end{tabular}	&\begin{tabular}[c]{@{}l@{}}0.6587\\$\pm$0.07\end{tabular}

\\\hline
ohsumed	 &\begin{tabular}[c]{@{}l@{}}0.6522\\$\pm$0.04\end{tabular}	&\begin{tabular}[c]{@{}l@{}}0.6560\\$\pm$0.05\end{tabular}	&\begin{tabular}[c]{@{}l@{}}0.6629\\$\pm$0.05\end{tabular}	 &\begin{tabular}[c]{@{}l@{}}0.6613\\$\pm$0.05\end{tabular}	&\begin{tabular}[c]{@{}l@{}}0.5830\\$\pm$0.03\end{tabular}	&\begin{tabular}[c]{@{}l@{}}0.7032\\$\pm$0.05\end{tabular}	 &\begin{tabular}[c]{@{}l@{}}0.6904\\$\pm$0.05\end{tabular}	&\begin{tabular}[c]{@{}l@{}}\textbf{0.7105}\\$\pm$0.04\end{tabular}

\\\hline
acpj	           &\begin{tabular}[c]{@{}l@{}}0.7407\\$\pm$0.06\end{tabular}	&\begin{tabular}[c]{@{}l@{}}0.7432\\$\pm$0.05\end{tabular}	&-	           &-	          &\begin{tabular}[c]{@{}l@{}}0.7428\\$\pm$0.06\end{tabular}	&\begin{tabular}[c]{@{}l@{}}0.7480\\$\pm$0.05\end{tabular}	&\begin{tabular}[c]{@{}l@{}}0.8098\\$\pm$0.07\end{tabular}	 &\begin{tabular}[c]{@{}l@{}}\textbf{0.8238}\\$\pm$0.05\end{tabular}

\\\hline
sido0	           &\begin{tabular}[c]{@{}l@{}}0.7417\\$\pm$0.03\end{tabular}	&\begin{tabular}[c]{@{}l@{}}0.8198\\$\pm$0.04\end{tabular}	&\begin{tabular}[c]{@{}l@{}}0.7797\\$\pm$0.04\end{tabular}	 &\begin{tabular}[c]{@{}l@{}}0.8796\\$\pm$0.04\end{tabular}	&\begin{tabular}[c]{@{}l@{}}0.7046\\$\pm$0.03\end{tabular}	&\begin{tabular}[c]{@{}l@{}}0.8607\\$\pm$0.02\end{tabular}	 &\begin{tabular}[c]{@{}l@{}}0.8672\\$\pm$0.03\end{tabular}	&\begin{tabular}[c]{@{}l@{}}\textbf{0.8958}\\$\pm$0.02\end{tabular}

\\\hline
thrombin	&\begin{tabular}[c]{@{}l@{}}0.8067\\$\pm$0.06\end{tabular}	&\begin{tabular}[c]{@{}l@{}}0.8063\\$\pm$0.07\end{tabular}	&\begin{tabular}[c]{@{}l@{}}0.8087\\$\pm$0.06\end{tabular}	 &\begin{tabular}[c]{@{}l@{}}0.7983\\$\pm$0.06\end{tabular}	&\begin{tabular}[c]{@{}l@{}}0.7758\\$\pm$0.08\end{tabular}	&\begin{tabular}[c]{@{}l@{}}0.8102\\$\pm$0.09\end{tabular}	 &\begin{tabular}[c]{@{}l@{}}\textbf{0.8319}\\$\pm$0.07\end{tabular}	&\begin{tabular}[c]{@{}l@{}}0.8160\\$\pm$0.08\end{tabular}

\\\hline
infant	           &\begin{tabular}[c]{@{}l@{}}0.7336\\$\pm$0.05\end{tabular}	&\begin{tabular}[c]{@{}l@{}}0.7334\\$\pm$0.05\end{tabular}	&\begin{tabular}[c]{@{}l@{}}0.7383\\$\pm$0.05\end{tabular}	 &\begin{tabular}[c]{@{}l@{}}0.7334\\$\pm$0.05\end{tabular}	&\begin{tabular}[c]{@{}l@{}}0.7299\\$\pm$0.05\end{tabular}	&\begin{tabular}[c]{@{}l@{}}0.7474\\$\pm$0.04\end{tabular}	 &\begin{tabular}[c]{@{}l@{}}0.7464\\$\pm$0.04\end{tabular}	&\begin{tabular}[c]{@{}l@{}}\textbf{0.7480}\\$\pm$0.04\end{tabular}
\\\hline
\end{tabular}
\label{tb6_54}
\end{table}

\begin{table}
\centering
\caption{ AUC using KNN}
\scriptsize
\begin{tabular}{|l|l|l|l|l|l|l|l|l|}
\hline
Dataset	&MMPC  &HITON-PC &MMMB &HITON-MB &IAMB &mRMR &CMI &JMI\\\hline

hiva	           &\begin{tabular}[c]{@{}l@{}}0.5387\\$\pm$0.04\end{tabular}	&\begin{tabular}[c]{@{}l@{}}0.5453\\$\pm$0.06\end{tabular}	&\begin{tabular}[c]{@{}l@{}}0.5420\\$\pm$0.04\end{tabular}	 &\begin{tabular}[c]{@{}l@{}}0.5453\\$\pm$0.06\end{tabular}	&\begin{tabular}[c]{@{}l@{}}0.5651\\$\pm$0.03\end{tabular}	&\begin{tabular}[c]{@{}l@{}}0.5768\\$\pm$0.04\end{tabular}	 &\begin{tabular}[c]{@{}l@{}}\textbf{0.5895}\\$\pm$0.04\end{tabular}	&\begin{tabular}[c]{@{}l@{}}0.5582\\$\pm$0.04\end{tabular}

\\\hline
ohsumed	&\begin{tabular}[c]{@{}l@{}}0.6055\\$\pm$0.05\end{tabular}	&\begin{tabular}[c]{@{}l@{}}0.6101\\$\pm$0.05\end{tabular}	&\begin{tabular}[c]{@{}l@{}}0.5754\\$\pm$0.03\end{tabular}	 &\begin{tabular}[c]{@{}l@{}}0.5750\\$\pm$0.03\end{tabular}	&\begin{tabular}[c]{@{}l@{}}0.5932\\$\pm$0.05\end{tabular}	&\begin{tabular}[c]{@{}l@{}}0.6218\\$\pm$0.06\end{tabular}	 &\begin{tabular}[c]{@{}l@{}}\textbf{0.6225}\\$\pm$0.05\end{tabular}	&\begin{tabular}[c]{@{}l@{}}\textbf{0.6225}\\$\pm$0.05\end{tabular}

\\\hline
acpj	          &\begin{tabular}[c]{@{}l@{}}0.5298\\$\pm$0.02\end{tabular}	&\begin{tabular}[c]{@{}l@{}}0.5301\\$\pm$0.02\end{tabular}	&-	                      &-	                      &\begin{tabular}[c]{@{}l@{}}0.5490\\$\pm$0.02\end{tabular} 	&\begin{tabular}[c]{@{}l@{}}\textbf{0.5648}\\$\pm$0.06\end{tabular}	&\begin{tabular}[c]{@{}l@{}}0.5596\\$\pm$0.03\end{tabular}	 &\begin{tabular}[c]{@{}l@{}}0.5622\\$\pm$0.04\end{tabular}

\\\hline
sido0	          &\begin{tabular}[c]{@{}l@{}}0.6618\\$\pm$0.03\end{tabular}	&\begin{tabular}[c]{@{}l@{}}0.6534\\$\pm$0.04\end{tabular}	&\begin{tabular}[c]{@{}l@{}}\textbf{0.6879}\\$\pm$0.03\end{tabular}	 &\begin{tabular}[c]{@{}l@{}}0.6873\\$\pm$0.02\end{tabular}	&\begin{tabular}[c]{@{}l@{}}0.6443\\$\pm$0.03\end{tabular}	&\begin{tabular}[c]{@{}l@{}}0.6483\\$\pm$0.03\end{tabular}	 &\begin{tabular}[c]{@{}l@{}}0.6563\\$\pm$0.04\end{tabular}	&\begin{tabular}[c]{@{}l@{}}0.6648\\$\pm$0.12\end{tabular}

\\\hline
thrombin      &\begin{tabular}[c]{@{}l@{}}0.7316\\$\pm$0.08\end{tabular}	&\begin{tabular}[c]{@{}l@{}}0.7289\\$\pm$0.06\end{tabular}	&\begin{tabular}[c]{@{}l@{}}0.7129\\$\pm$0.08\end{tabular}	 &\begin{tabular}[c]{@{}l@{}}0.7481\\$\pm$0.08\end{tabular}	&\begin{tabular}[c]{@{}l@{}}0.7018\\$\pm$0.08\end{tabular}	&\begin{tabular}[c]{@{}l@{}}\textbf{0.7576}\\$\pm$0.05\end{tabular}	 &\begin{tabular}[c]{@{}l@{}}0.7465\\$\pm$0.09\end{tabular}	&\begin{tabular}[c]{@{}l@{}}0.7471\\$\pm$0.07\end{tabular}

\\\hline
infant	         &\begin{tabular}[c]{@{}l@{}}0.6595\\$\pm$0.04\end{tabular}	&\begin{tabular}[c]{@{}l@{}}\textbf{0.6797}\\$\pm$0.04\end{tabular}	&\begin{tabular}[c]{@{}l@{}}\textbf{0.6797}\\$\pm$0.04\end{tabular}	 &\begin{tabular}[c]{@{}l@{}}0.6551\\$\pm$0.04\end{tabular}	&\begin{tabular}[c]{@{}l@{}}0.6360\\$\pm$0.05\end{tabular}	&\begin{tabular}[c]{@{}l@{}}0.6739\\$\pm$0.03\end{tabular}	 &\begin{tabular}[c]{@{}l@{}}0.6734\\$\pm$0.04\end{tabular}	&\begin{tabular}[c]{@{}l@{}}0.6729\\$\pm$0.03\end{tabular}
\\\hline
\end{tabular}
\label{tb6_55}
\end{table}
\begin{figure}
\centering
\includegraphics[height=1.7in]{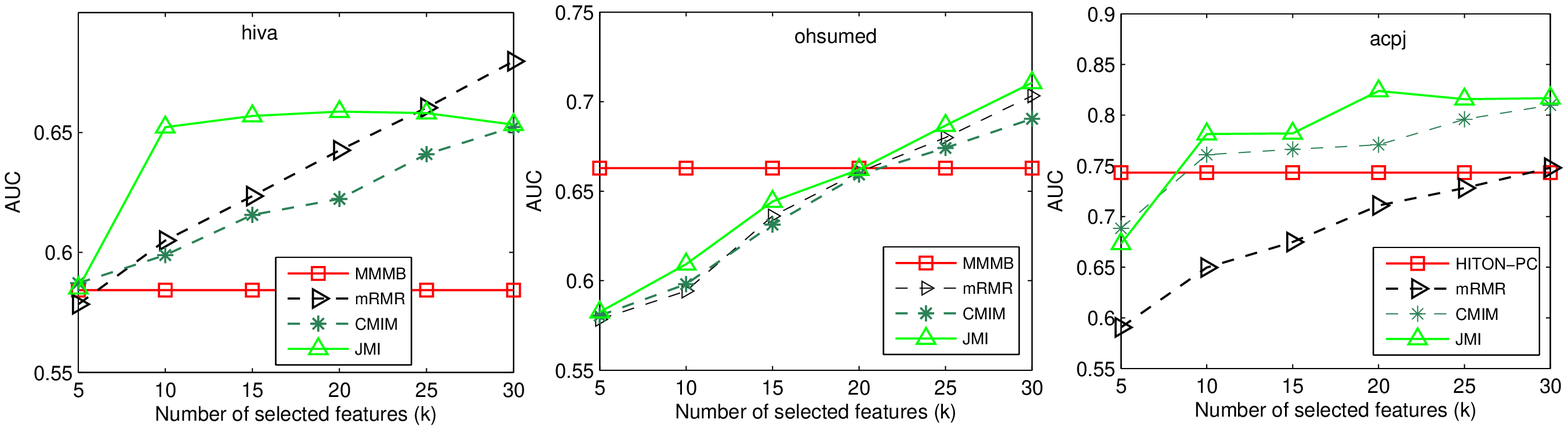}\\
\includegraphics[height=1.7in]{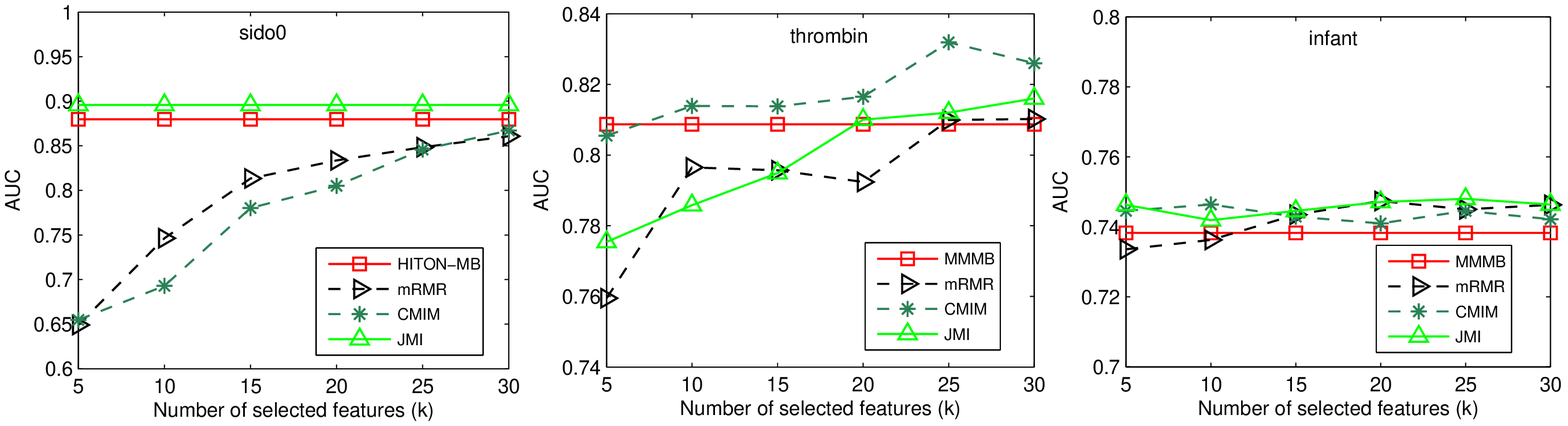}
\caption{Prediction accuracy with different values of $\psi$ using NBC}
\label{fig6_51}
\end{figure}
\begin{figure}
\centering
\includegraphics[height=1.7in]{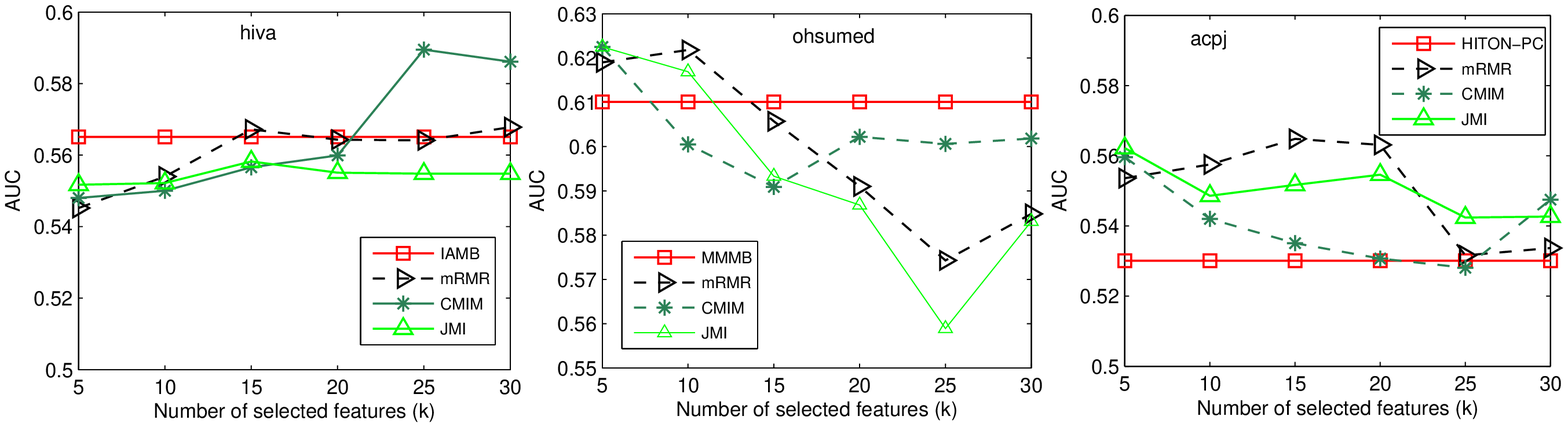}\\
\includegraphics[height=1.7in]{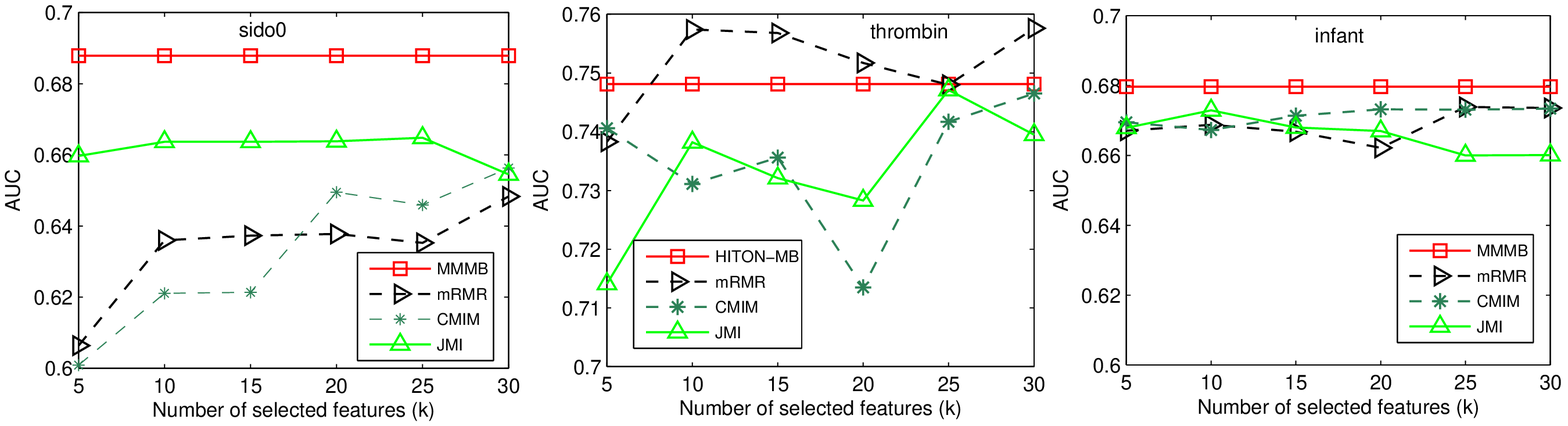}
\caption{Prediction accuracy with different values of $\psi$ using KNN}
\label{fig6_52}
\end{figure}
\begin{table}
\centering
\caption{Kappa statistic using NBC}
\scriptsize
\begin{tabular}{|l|l|l|l|l|l|l|l|l|}
\hline
\multicolumn{1}{|c|}{Dataset} & \multicolumn{1}{c|}{MMPC}                                  & \multicolumn{1}{c|}{HITON-PC}                              & \multicolumn{1}{c|}{MMMB}                                    & \multicolumn{1}{c|}{HITON-MB}                              & \multicolumn{1}{c|}{IAMB}                                  & \multicolumn{1}{c|}{mRMR}                                  & \multicolumn{1}{c|}{CMIM}                                  & \multicolumn{1}{c|}{JMI}                                   \\ \hline
hiva                          & \begin{tabular}[c]{@{}l@{}}0.2311\\ $\pm$0.12\end{tabular} & \begin{tabular}[c]{@{}l@{}}0.2295\\ $\pm$0.12\end{tabular} & \begin{tabular}[c]{@{}l@{}}0.2397\\ $\pm$0.15\end{tabular}   & \begin{tabular}[c]{@{}l@{}}0.2295\\ $\pm$0.12\end{tabular} & \begin{tabular}[c]{@{}l@{}}0.1948\\ $\pm$0.09\end{tabular} & \begin{tabular}[c]{@{}l@{}}\textbf{0.3602}\\ $\pm$0.10\end{tabular} & \begin{tabular}[c]{@{}l@{}}0.2601\\ $\pm$0.08\end{tabular} & \begin{tabular}[c]{@{}l@{}}0.2941\\ $\pm$0.14\end{tabular} \\ \hline

ohsumed                       & \begin{tabular}[c]{@{}l@{}}0.3905\\ $\pm$0.10\end{tabular} & \begin{tabular}[c]{@{}l@{}}0.3992\\ $\pm$0.11\end{tabular} & \begin{tabular}[c]{@{}l@{}}0.3974\\ $\pm$0.10\end{tabular}   & \begin{tabular}[c]{@{}l@{}}0.3936\\ $\pm$0.11\end{tabular} & \begin{tabular}[c]{@{}l@{}}0.2464\\ $\pm$0.08\end{tabular} & \begin{tabular}[c]{@{}l@{}}0.4626\\ $\pm$0.09\end{tabular} & \begin{tabular}[c]{@{}l@{}}0.4370\\ $\pm$0.11\end{tabular} & \begin{tabular}[c]{@{}l@{}}\textbf{0.4642}\\ $\pm$0.09\end{tabular} \\ \hline

acpj                          & \begin{tabular}[c]{@{}l@{}}0.2082\\ $\pm$0.04\end{tabular} & \begin{tabular}[c]{@{}l@{}}0.2105\\ $\pm$0.04\end{tabular} & -                                                            & -                                                          & \begin{tabular}[c]{@{}l@{}}0.2427\\ $\pm$0.06\end{tabular} & \begin{tabular}[c]{@{}l@{}}\textbf{0.2517}\\ $\pm$0.04\end{tabular} & \begin{tabular}[c]{@{}l@{}}0.1995\\ $\pm$0.04\end{tabular} & \begin{tabular}[c]{@{}l@{}}0.1853\\ $\pm$0.02\end{tabular} \\ \hline

sido0                         & \begin{tabular}[c]{@{}l@{}}\textbf{0.4176}\\ $\pm$0.05\end{tabular} & \begin{tabular}[c]{@{}l@{}}0.3763\\ $\pm$0.05\end{tabular} & \begin{tabular}[c]{@{}l@{}}0.3906\\ $\pm$0.05\end{tabular}   & \begin{tabular}[c]{@{}l@{}}0.3630\\ $\pm$0.03\end{tabular} & \begin{tabular}[c]{@{}l@{}}0.4125\\ $\pm$0.05\end{tabular} & \begin{tabular}[c]{@{}l@{}}0.3684\\ $\pm$0.02\end{tabular} & \begin{tabular}[c]{@{}l@{}}0.3636\\ $\pm$0.04\end{tabular} & \begin{tabular}[c]{@{}l@{}}0.3584\\ $\pm$0.03\end{tabular} \\ \hline

thrombin                      & \begin{tabular}[c]{@{}l@{}}0.5256\\ $\pm$0.10\end{tabular} & \begin{tabular}[c]{@{}l@{}}0.4762\\ $\pm$0.12\end{tabular} & \begin{tabular}[c]{@{}l@{}}0.4862\\ $\pm$0.09\end{tabular}   & \begin{tabular}[c]{@{}l@{}}0.3543\\ $\pm$0.07\end{tabular} & \begin{tabular}[c]{@{}l@{}}0.5114\\ $\pm$0.12\end{tabular} & \begin{tabular}[c]{@{}l@{}}\textbf{0.5935}\\ $\pm$0.12\end{tabular} & \begin{tabular}[c]{@{}l@{}}0.4533\\ $\pm$0.08\end{tabular} & \begin{tabular}[c]{@{}l@{}}0.4870\\ $\pm$0.09\end{tabular} \\ \hline

infant                        & \begin{tabular}[c]{@{}l@{}}\textbf{0.5329}\\ $\pm$0.09\end{tabular} & \begin{tabular}[c]{@{}l@{}}0.5303\\ $\pm$0.08\end{tabular} & \begin{tabular}[c]{@{}l@{}}0.5311\\ $\pm$0.0.09\end{tabular} & \begin{tabular}[c]{@{}l@{}}0.5306\\ $\pm$0.08\end{tabular} & \begin{tabular}[c]{@{}l@{}}\textbf{0.5329}\\ $\pm$0.08\end{tabular} & \begin{tabular}[c]{@{}l@{}}0.5027\\ $\pm$0.07\end{tabular} & \begin{tabular}[c]{@{}l@{}}0.5055\\ $\pm$0.09\end{tabular} & \begin{tabular}[c]{@{}l@{}}0.4867\\ $\pm$0.08\end{tabular} \\ \hline
\end{tabular}
\label{tb6_58}
\end{table}
\begin{table}
\centering
\caption{Kappa statistic using KNN}
\scriptsize
\begin{tabular}{|l|l|l|l|l|l|l|l|l|}
\hline
Dataset  & \multicolumn{1}{c|}{MMPC}                                  & \multicolumn{1}{c|}{HITON-PC}                              & \multicolumn{1}{c|}{MMMB}                                  & \multicolumn{1}{c|}{HITON-MB}                              & \multicolumn{1}{c|}{IAMB}                                  & \multicolumn{1}{c|}{mRMR}                                  & \multicolumn{1}{c|}{CMIM}                                  & \multicolumn{1}{c|}{JMI}                                   \\ \hline
hiva     & \begin{tabular}[c]{@{}l@{}}0.1234\\ $\pm$0.12\end{tabular} & \begin{tabular}[c]{@{}l@{}}0.1373\\ $\pm$0.16\end{tabular} & \begin{tabular}[c]{@{}l@{}}0.1308\\ $\pm$0.13\end{tabular} & \begin{tabular}[c]{@{}l@{}}0.1447\\ $\pm$0.16\end{tabular} & \begin{tabular}[c]{@{}l@{}}0.2022\\ $\pm$0.10\end{tabular} & \begin{tabular}[c]{@{}l@{}}0.2236\\ $\pm$0.11\end{tabular} & \begin{tabular}[c]{@{}l@{}}\textbf{0.2448}\\ $\pm$0.10\end{tabular} & \begin{tabular}[c]{@{}l@{}}0.1823\\ $\pm$0.13\end{tabular} \\ \hline

ohsumed  & \begin{tabular}[c]{@{}l@{}}0.2852\\ $\pm$0.13\end{tabular} & \begin{tabular}[c]{@{}l@{}}0.2922\\ $\pm$0.12\end{tabular} & \begin{tabular}[c]{@{}l@{}}0.2227\\ $\pm$0.07\end{tabular} & \begin{tabular}[c]{@{}l@{}}0.2210\\ $\pm$0.06\end{tabular} & \begin{tabular}[c]{@{}l@{}}0.2553\\ $\pm$0.10\end{tabular} & \begin{tabular}[c]{@{}l@{}}0.2881\\ $\pm$0.10\end{tabular} & \begin{tabular}[c]{@{}l@{}}\textbf{0.2892}\\ $\pm$0.10\end{tabular} & \begin{tabular}[c]{@{}l@{}}\textbf{0.2892}\\ $\pm$0.10\end{tabular} \\ \hline

acpj     & \begin{tabular}[c]{@{}l@{}}0.0907\\ $\pm$0.05\end{tabular} & \begin{tabular}[c]{@{}l@{}}0.0928\\ $\pm$0.05\end{tabular} & -                                                          & -                                                          & \begin{tabular}[c]{@{}l@{}}0.1378\\ $\pm$0.07\end{tabular} & \begin{tabular}[c]{@{}l@{}}\textbf{0.1889}\\ $\pm$0.14\end{tabular} & \begin{tabular}[c]{@{}l@{}}0.1770\\ $\pm$0.08\end{tabular} & \begin{tabular}[c]{@{}l@{}}0.1744\\ $\pm$0.09\end{tabular} \\ \hline

sido0    & \begin{tabular}[c]{@{}l@{}}0.4119\\ $\pm$0.06\end{tabular} & \begin{tabular}[c]{@{}l@{}}0.3703\\ $\pm$0.10\end{tabular} & \begin{tabular}[c]{@{}l@{}}\textbf{0.4582}\\ $\pm$0.07\end{tabular} & \begin{tabular}[c]{@{}l@{}}0.4481\\ $\pm$0.06\end{tabular} & \begin{tabular}[c]{@{}l@{}}0.3628\\ $\pm$0.06\end{tabular} & \begin{tabular}[c]{@{}l@{}}0.3807\\ $\pm$0.08\end{tabular} & \begin{tabular}[c]{@{}l@{}}0.3970\\ $\pm$0.08\end{tabular} & \begin{tabular}[c]{@{}l@{}}0.3215\\ $\pm$0.07\end{tabular} \\ \hline

thrombin & \begin{tabular}[c]{@{}l@{}}0.5459\\ $\pm$0.14\end{tabular} & \begin{tabular}[c]{@{}l@{}}0.5395\\ $\pm$0.12\end{tabular} & \begin{tabular}[c]{@{}l@{}}0.5436\\ $\pm$0.13\end{tabular} & \begin{tabular}[c]{@{}l@{}}0.5551\\ $\pm$0.12\end{tabular} & \begin{tabular}[c]{@{}l@{}}0.4721\\ $\pm$0.15\end{tabular} & \begin{tabular}[c]{@{}l@{}}\textbf{0.5896}\\ $\pm$0.09\end{tabular} & \begin{tabular}[c]{@{}l@{}}0.5584\\ $\pm$0.15\end{tabular} & \begin{tabular}[c]{@{}l@{}}0.5674\\ $\pm$0.14\end{tabular} \\ \hline

infant   & \begin{tabular}[c]{@{}l@{}}0.4433\\ $\pm$0.08\end{tabular} & \begin{tabular}[c]{@{}l@{}}0.4344\\ $\pm$0.09\end{tabular} & \begin{tabular}[c]{@{}l@{}}\textbf{0.4824}\\ $\pm$0.09\end{tabular} & \begin{tabular}[c]{@{}l@{}}0.4385\\ $\pm$0.11\end{tabular} & \begin{tabular}[c]{@{}l@{}}0.3889\\ $\pm$0.11\end{tabular} & \begin{tabular}[c]{@{}l@{}}0.4726\\ $\pm$0.07\end{tabular} & \begin{tabular}[c]{@{}l@{}}0.4650\\ $\pm$0.09\end{tabular} & \begin{tabular}[c]{@{}l@{}}0.4621\\ $\pm$0.05\end{tabular} \\ \hline
\end{tabular}
\label{tb6_59}
\end{table}

\begin{table}
\centering
\caption{Number of selected features}
\scriptsize
\begin{tabular}{|l|l|l|l|l|l|l|l|l|}
\hline
Dataset	&MMPC  &HITON-PC &MMMB &HITON-MB &IAMB &mRMR &CMI &JMI\\\hline

hiva	             &7	   &6	&9	   &7	&8	      &30/30	&30/25	&20/15
\\\hline
ohsumed	&26	   &26	&38	   &37	&8	      &30/10	&20/5	 &30/5
\\\hline
acpj	             &16	   &16	&-      &-	&10      &30/15	&30/5	&20/5
\\\hline
sido0	      &16	   &17	&62	  &68	&10	     &30/30	&30/30	&5/25
\\\hline
thrombin      &15	   &11	&38	  &42	&7	     &30	/30  &25/30	 &30/25
\\\hline
infant	      &5	   &5	&7	   &6	&3	     &20/25	&10/30	&25/10
\\
\hline
\end{tabular}
\label{tb6_56}
\end{table}

\begin{table}
\centering
\caption{Running time (in seconds)}
\scriptsize
\begin{tabular}{|l|l|l|l|l|l|l|l|l|}
\hline
Dataset	&MMPC  &HITON-PC &MMMB &HITON-MB &IAMB &mRMR &CMI &JMI\\\hline

hiva	              &2	       &2	       &40	    &15	    &13	       &1	    &1	            &4
\\\hline
ohsumed	 &61	&54	       &563	   &577	    &89	      &32	   &2	            &50
\\\hline
acpj	             &29	       &3         &-	          &-	    &905	      &136   &9	            &175
\\\hline
sido0	       &33	&77	      &8,928	  &8,669	    &206	      &5	    &2	             &13
\\\hline
thrombin	&544	&133	&23,456	  &17,615   &1429	&241   &11    	      &291
\\\hline
infant	       &1	      &1	        &1	         &1	    &0.5         &0.1   &0.01        	&0.1
\\
\hline
\end{tabular}
\label{tb6_57}
\end{table}

\bibliography{cfs.bib}

\end{document}